\documentclass[conference]{IEEEtran}
\IEEEoverridecommandlockouts 
\usepackage{times}
\usepackage{amssymb}
\usepackage{amsmath}
\usepackage{float}
\usepackage{comment}
\usepackage{graphicx}
\usepackage{hyperref}
\usepackage{mathtools}
\usepackage{subcaption}
\usepackage[numbers]{natbib}
\usepackage{multicol}

\usepackage{amsthm, algorithm, algorithmicx}  
\usepackage[noend]{algpseudocode}
\usepackage[english]{babel}

\let\labelindent\relax

\usepackage[inline]{enumitem}
\newtheorem{innercustomgeneric}{\customgenericname}
\providecommand{\customgenericname}{}
\newcommand{\newcustomtheorem}[2]{
  \newenvironment{#1}[1]
  {
   \renewcommand\customgenericname{#2}
   \renewcommand\theinnercustomgeneric{##1}
   \innercustomgeneric
  }
  {\endinnercustomgeneric}
}

\newcustomtheorem{customthm}{Theorem}
\newcustomtheorem{customlemma}{Lemma}
\newcustomtheorem{customprop}{Proposition}

\theoremstyle{plain}  
\newtheorem{theorem}{Theorem}
\newtheorem{lemma}[theorem]{Lemma}
\newtheorem{definition}[theorem]{Definition}
\newtheorem{remark}[theorem]{Remark}
\newtheorem{proposition}[theorem]{Proposition}

\algnewcommand\algorithmicinput{\textbf{Input:}}
\algnewcommand\algorithmicoutput{\textbf{Output:}}
\algnewcommand\Input{\item[\algorithmicinput]}
\algnewcommand\Output{\item[\algorithmicoutput]}

\DeclareMathOperator*{\argmin}{arg\,min}
\DeclareMathOperator*{\argmax}{arg\,max}

\usepackage{xcolor}

\newcommand{\norm}[1]{\left\lVert#1\right\rVert}

\begin{document}

\title{
Decentralized Multi-Robot Line-of-Sight Connectivity Maintenance under Uncertainty
}

\author{Yupeng Yang$^1$, Yiwei Lyu$^2$, Yanze Zhang$^1$, Sha Yi$^3$ and Wenhao Luo$^1$
\thanks{$^*$This work was supported in part by the U.S. National Science Foundation under Grant CMMI-2301749.}
\thanks{$^1$The authors are with the Department of Computer Science, University of North Carolina at Charlotte, Charlotte, NC 28223, USA. Email: {\tt \{yyang52, yzhang94, wenhao.luo\}@charlotte.edu}}
\thanks{$^{2}$The author is with the Department of Electrical and Computer Engineering, Carnegie Mellon University, Pittsburgh, PA 15213, USA. Email: {\tt yiweilyu@andrew.cmu.edu}}
\thanks{$^{3}$The author is with the Robotics Institute, Carnegie Mellon University, Pittsburgh, PA 15213, USA. Email: {\tt shayi@cs.cmu.edu}}}

\maketitle

\begin{abstract}
In this paper, we propose a novel decentralized control method to maintain Line-of-Sight connectivity for multi-robot networks in the presence of Guassian-distributed localization uncertainty. In contrast to most existing work that assumes perfect positional information about robots or enforces overly restrictive rigid formation against uncertainty, our method enables robots to preserve Line-of-Sight connectivity with high probability under unbounded Gaussian-like positional noises while remaining minimally intrusive to the original robots’ tasks. This is achieved by a motion coordination framework that jointly optimizes the set of existing Line-of-Sight edges to preserve and control revisions to the nominal task-related controllers, subject to the safety constraints and the corresponding composition of uncertainty-aware Line-of-Sight control constraints. Such compositional control constraints, expressed by our novel notion of probabilistic Line-of-Sight connectivity barrier certificates (PrLOS-CBC) for pairwise robots using control barrier functions, explicitly characterize the deterministic admissible control space for the two robots. The resulting motion ensures Line-of-Sight connectedness for the robot team with high probability. Furthermore, we propose a fully decentralized algorithm that decomposes the motion coordination framework by interleaving the composite constraint specification and solving for the resulting optimization-based controllers. The optimality of our approach is justified by the theoretical proofs. Simulation and real-world experiments results are given to demonstrate the effectiveness of our method.
\end{abstract}

\section{Introduction}\label{sec:intro}
To facilitate efficient information sharing among teams of robots, connectivity maintenance often considers maintaining the distance-based multi-robot communication graph as one connected component by constraining the coordinated robots' motions, commonly referred to as maintaining \emph{global connectivity} \cite{capelli2020connectivity,ong2021network,sabattini2013decentralized}. 
Local methods seek to maintain connectivity among multiple robots by preserving the initial communication graph topology over time~\cite{ji2007distributed,zavlanos2007flocking},
while in global methods the algebraic connectivity of the communication graph is maintained by a secondary connectivity controller that keeps the second-smallest eigenvalue of the graph Laplacian positive at all times~\cite{capelli2020connectivity,sabattini2013decentralized}. In realistic environments, however, the established multi-robot networks dependent on conventional distance-based communication models may suffer from potential interruptions, e.g., wireless signals between two robots within communication range could still be obstructed by thick metal walls~\cite{tuck2021dec} and 
thus disconnect the whole network.

Hence, Line-of-Sight (LOS) connectivity maintenance is gaining attention for ensuring reliable data exchange among robots~\cite{tuck2021dec,sun2020optimal,yang2023minimally}. 
However, these methods assume perfect positional information, and uncertainties like noisy locational data from real-world sensing can significantly affect its performance. Although connectivity-focused controllers can mitigate uncertainty with conservative motion \cite{shetty2023decentralized}, they may hinder other primary robot objectives. Therefore, addressing uncertainties while enabling flexible robot motion and ensuring LOS communication is vital for multi-robot coordination. 

To solve the probabilistic constraint satisfaction problems due to uncertainties such as noisy observations, recent works \cite{zhu2019chance, castillo2020real} have focused on enforcing the probability through state-related chance constraints in applications such as collision avoidance. However, constraints based on robot states may necessitate enlarged bounding volumes to account for braking distances, potentially overestimating the probability and leading to conservative behaviors. By leveraging Control Barrier Functions (CBFs) the researches in \cite{zhu2023probabilistic, zhang2023occlusion} introduce deterministic control constraints to ensure chance-constrained properties such as safety.
These deterministic constraints, in quadratic form, can raise computational costs when integrated into optimization problems. Authors in \cite{luo2020multi} proposed Probabilistic Safety Control Barrier Certificates (PrSBC), which develop linear control constraints for pairwise robots to guarantee chance-constrained state-related safety under positional and motion noises. The resulting robot motions are less restrictive and the developed control constraints are easy to compose for achieving team-level specifications. This inspires further research on composition of pairwise connectivity constraints for ensuring team-level properties such as communication graph connectivity among robots.

On the other hand, existing connectivity maintenance methods often rely on centralized computation and synchronized communication \cite{yang2023minimally,wang2016multi}, raising scalability issues for multi-robot systems. 
Instead, decentralized frameworks such as C-ADMM \cite{boyd2011distributed} can effectively distribute the computational load across all robots by solving one centralized optimization problem in a decentralized manner. However, such methods may be limited to predefined constraints \cite{pereira2022decentralized, shorinwa2023distributed}. 
In flexible multi-robot connectivity maintenance with switching topology, it is often difficult to pre-determine the particular set of edges needed for establishing the corresponding constraints. 
This added complexity makes it particularly challenging to co-optimize connectivity constraints and desired robots' behaviors in a fully decentralized manner.

In this paper, we seek to minimally modify the nominal task-related multi-robot controller while ensuring global and subgroup LOS connectivity for the team of robots under positional noises. 
Our contributions are as follows.
\begin{itemize}
    \item A novel notion of Probabilistic Line-of-Sight Connectivity Barrier Certificate (PrLOS-CBC) is proposed to define the admissible control space, from which the existing pairwise LOS can be preserved with high probability under noisy positional information;
    \item A decentralized algorithm, referred to as Uncertainty-Aware Decentralized Line-of-Sight Least Constraining Tree (Dec-LOS-LCT), is proposed to interleave the bi-level optimization process of selective LOS constraint specification and constrained control optimization;
    \item The theoretical proofs, simulation and real-world experiments are given to justify the performance of our method.
\end{itemize}

\section{Preliminaries}
Consider a robotic team $\mathcal{S}$ with $N$ robots moving in a $d$-dimensional shared workspace, which consists of free space and occupied space $\mathcal{C}_{\mathrm{obs}}=\bigcup\limits_{k=1}^K \mathcal{O}_k$ by $K$ static polyhedral obstacles $\mathcal{O}_k \subset \mathbb{R}^{d},\forall k$. The positions of the static obstacles are assumed to be known by the robots. Each robot $i\in \mathcal{I}= \{1,..,N\}$ is centered at the position $\mathbf{x}_i\!\in\!\mathcal{X}\!\subset\!\mathbb{R}^{d} $. The system dynamics $\dot{\mathbf{x}}_i$ affine in control and the noisy observation $\hat{\mathbf{x}}_{i} \in \mathbb{R}^{d}$ of each robot $i$ are described as follows.
\begin{equation}
\begin{split}
\label{dynamics}
    &\dot{\mathbf{x}}_i = F_i(\mathbf{x}_i) +G_i(\mathbf{x}_i)\mathbf{u}_i,\;\hat{\mathbf{x}}_i = \mathbf{x}_i + \epsilon_i, \;\epsilon_i \sim \mathcal{N}(0,{\Sigma}_i)
\end{split}
\end{equation}
where $\mathbf{u}_i\!\in\!\mathcal{U}\!\subset\!\mathbb{R}^{q}$ denotes the control input. $F_i:\mathbb{R}^{d} \mapsto \mathbb{R}^{d}$ and $G_i\!:\!\mathbb{R}^{d}\!\mapsto\!\mathbb{R}^{d\times q}$ are locally Lipschitz continuous. $\epsilon_i\!\in\! \mathbb{R}^{d}$ is the measurement noise and considered as a continuous independent random variable with a Gaussian distribution that can vary at each time step. 

\subsection{Safety and Range Limited Line-of-Sight Connectivity}
We assume the $K$ static polyhedral obstacles can be commonly represented by $L$ discretized obstacles modeled as rigid spheres along the boundary of the static obstacles \cite{thirugnanam2021duality}. Each discretized obstacle can be denoted as $o \in \{1,...,L\}$. Consider the joint robot states $\mathbf{x} = \{\mathbf{x}_1,...,\mathbf{x}_N\} 
\in \mathcal{X} 
\subset \mathbb{R}^{dN}$, the discretized joint obstacle states $\mathbf{x}^\mathrm{obs}=\{\mathbf{x}^\mathrm{obs}_1,\ldots,\mathbf{x}^\mathrm{obs}_L\}\in\mathcal{X}^\mathrm{obs}\subset \mathbb{R}^{dL}$, the minimum inter-robot safe distance as $R_\mathrm{s}\in\mathbb{R}$, and the minimum obstacle-robot safe distance as $R_\mathrm{obs}\in\mathbb{R}$, the desired sets for any pairwise robots $i$, $j$ and obstacle $o$ satisfying inter-robot or robot-obstacle collision avoidance can be defined as: 
\begin{align}
&h^\mathrm{s}_{i,j}(\mathbf{x}) = ||\mathbf{x}_i -\mathbf{x}_j||^2 -R_\mathrm{s}^2, \forall i>j, \notag\\
&\mathcal{H}^\mathrm{s}_{i,j} = \{\mathbf{x} \in \mathbb{R}^{dN} | h^\mathrm{s}_{i,j}(\mathbf{x}) \geq 0 \} \label{eq:h_safe_per}\\
&h^\mathrm{obs}_{i,o}(\mathbf{x},\mathbf{x}^\mathrm{obs}) = ||\mathbf{x}_i -\mathbf{x}^\mathrm{obs}_o||^2 -R_\mathrm{obs}^2, \forall i,o,\notag\\
&\mathcal{H}^\mathrm{obs}_{i,o} = \{\mathbf{x} \in \mathbb{R}^{dN}, \mathbf{x}^\mathrm{obs}\in\mathbb{R}^{dL}| h^\mathrm{obs}_{i,o}(\mathbf{x},\mathbf{x}^\mathrm{obs}_{o}) \geq 0 \} \label{eq:h_o_safe}
 \end{align}
 
Pairwise robots $i$ and $j$ are said to be Line-of-Sight (LOS) connected when they are \textit{not only} within the limited communication range (\textbf{communication distance condition}), \textit{but also} have an unobstructed LOS between the two for all obstacles (\textbf{occlusion-free condition}). The undirected LOS edge between such two robots can be denoted by 
$(v_i,v_j) \in \mathcal{E}^\mathrm{los}$ (i.e., $(v_i,v_j) \in \mathcal{E}^\mathrm{los} \Leftrightarrow (v_j,v_i) \in \mathcal{E}^\mathrm{los}$), where each node $v_i,v_j \in \mathcal{V}$ represents a robot, and we denote LOS communication graph \footnote{For ease of notation, the dependency of $\mathcal{G}^\mathrm{los}(t)$ on time $t$ may be omitted.} as $\mathcal{G}^\mathrm{los}= (\mathcal{V},\mathcal{E}^\mathrm{los})$ for the entire team.

Considering the communication range $R_\mathrm{c}\in\mathbb{R}$, the desired set of $\mathbf{x}$ for pairwise robots $i$ and $j$ satisfying the \textbf{communication distance condition} can be defined as:
\begin{align}
&h^\mathrm{c}_{i,j}(\mathbf{x}) = R_\mathrm{c}^2 - ||\mathbf{x}_i - \mathbf{x}_j||^2, \;\forall (v_i,v_j)\in\mathcal{E}^\mathrm{c}\subseteq\mathcal{E}^\mathrm{los},\;\notag\\
&\mathcal{H}^\mathrm{c}_{i,j} = \{\mathbf{x} \in \mathbb{R}^{dN}| h^\mathrm{c}_{i,j}(\mathbf{x}) \geq 0 \} \label{eq:h_conn_per}
\end{align}

Accordingly, the desired set satisfying \textbf{occlusion-free condition} for pairwise LOS connectivity of robots $i,j$ can be described as follows.
\begin{align}\begin{split}
\label{eq:h_los}
    \mathcal{H}^\mathrm{los}_{i,j} = &\{\mathbf{x} \in \mathbb{R}^{dN}|\; h^\mathrm{los}_{i,j}(\mathbf{x},\mathcal{C}_\text{obs})\geq 0 \}=\\
    &\{\mathbf{x} \in \mathbb{R}^{dN}|\; \mathbf{x}_i(1-\Omega)+\mathbf{x}_j\Omega \notin \mathcal{C}_{\text{obs}},\;\forall \Omega\in [0,1]\}
\end{split}\end{align}

Therefore, for the entire team with any given LOS connectivity communication graph $\mathcal{G}^{\mathrm{slos}}=(\mathcal{V},\mathcal{E}^\mathrm{slos})\subseteq \mathcal{G}^{\mathrm{los}}$ with $\mathcal{E}^\mathrm{slos}\subseteq\mathcal{E}^\mathrm{los}$ to enforce, the desired set implying the satisfying LOS connectivity condition can be defined by the intersection of the connectivity set $\mathcal{H}^{\mathrm{c}}_{i,j}$ and occlusion-free set $\mathcal{H}^\mathrm{los}_{i,j}$ as follows.

\begin{align}
\label{eq:hlosset}
    \mathcal{H}^\mathrm{los}(\mathcal{G}^\mathrm{slos})= &\Bigl(\bigcap_{\{v_i,v_j \in \mathcal{V}:(v_i,v_j)\in\mathcal{E}^\mathrm{slos}\}} \mathcal{H}^\mathrm{los}_{i,j} \Bigr)\bigcap \notag\\
    &\Bigl(\bigcap_{\{v_i,v_j \in \mathcal{V}:(v_i,v_j)\in\mathcal{E}^\mathrm{slos}\}} \mathcal{H}^\mathrm{c}_{i,j} \Bigr)
\end{align}

\subsection{Probabilistic Safety and Line-of-Sight Connectivity using Control Barrier Function}

To guarantee the system's safety in a deterministic dynamical system, control barrier functions could be used to enforce the safety set forward invariant by defining an admissible control space. The results can be summarized as follows. 

\begin{lemma}
\label{lem:cbf}[Summarized from \cite{ames2019control}]
Given a deterministic dynamical system affine in control (i.e., $\dot{\mathbf{x}}=F(\mathbf{x})+G(\mathbf{x})\mathbf{u}$) and a desired set $\mathcal{H}$ as the 0-super level set of a continuously differentiable function $h: \mathcal{X} \mapsto \mathbb{R}$, the function $h$ is called a control barrier function, if there exists an extended class-$\mathcal{K}$ function\footnote{In the rest of this paper, we select the particular choice of $\kappa(h(\mathbf{x}))=\gamma h(\mathbf{x})$ with $\gamma$ as a user-defined parameter \cite{ames2019control}.}  $\kappa(\cdot)$ such that 
$\sup_{\mathbf{u}\in\mathcal{U}}\{\dot{h}(\mathbf{x}, \mathbf{u})\}\geq -\kappa(h(\mathbf{x}))$ for all $\mathbf{x}\in\mathcal{X}$. 
\end{lemma}

With Lemma~\ref{lem:cbf}, the admissible control space for any Lipschitz continuous controller $\mathbf{u}\in \mathcal{U}$ rendering $\mathcal{H}$ forward invariant (i.e., keeping the system state $\mathbf{x}$ staying in $\mathcal{H}$ overtime) thus becomes:
\begin{align}\label{eq:cbc_lemma}
    \mathcal{B}(\mathbf{x}) = \{ \mathbf{u}\in \mathcal{U} | \dot{h}(\mathbf{x},\mathbf{u}) + \kappa(h(\mathbf{x}))\geq 0 \}
\end{align}

Due to the unbounded noisy observations described in Eq.~(\ref{dynamics}), the desired sets for collision avoidance between pairwise robots can only be satisfied in a probabilistic manner. 
Motivated by Lemma~\ref{lem:cbf},
the probabilistic safety barrier certificates (PrSBC)~\cite{luo2020multi} has been presented
to specify the admissible control space $\mathcal{S}^{\sigma^\mathrm{s}}_\mathbf{u}(\hat{\mathbf{x}})$ that guarantees the probability of pairwise safety $\text{Pr} (\mathbf{x}_i,\mathbf{x}_j \in \mathcal{H}^\mathrm{s}_{i,j})$ above the user-defined confidence level $\sigma^\mathrm{s}\in(0,1)$, i.e., $\mathbf{u}\in\mathcal{S}^{\sigma^\mathrm{s}}_\mathbf{u}(\hat{\mathbf{x}}) \implies \text{Pr} (\mathbf{x}_i,\mathbf{x}_j \in \mathcal{H}^\mathrm{s}_{i,j}) \geq \sigma^\mathrm{s},\forall i>j$.
\begin{align}\begin{aligned}
\label{Prsbc}
\mathcal{S}^{\sigma^\mathrm{s}}_\mathbf{u}(\mathbf{\hat{x}}) = \{\mathbf{u} \in \mathbb{R}^{qN} | A^{\sigma^\mathrm{s}}_{i,j}\mathbf{u}\leq g^{\sigma^\mathrm{s}}_{i,j}, \forall i >j, \\
A^{\sigma^\mathrm{s}}_{i,j} \in \mathbb{R}^{1\times qN}, g^{\sigma^\mathrm{s}}_{i,j} \in \mathbb{R}\}
\end{aligned}\end{align}
where $A^{\sigma^\mathrm{s}}_{i,j}$ and $g^{\sigma^\mathrm{s}}_{i,j}$ are determined by observed robot states, distribution of system noise, and confidence level $\sigma^\mathrm{s}$. 
Likewise, the admissible control space $S^{\sigma^\mathrm{obs}}_\mathbf{u}(\hat{\mathbf{x}},\mathbf{x}^\mathrm{obs})$ for robot-obstacle collision avoidance could be derived in a similar form as Eq.~(\ref{Prsbc}) to guarantee the chance-constrained robot-obstacle safety (i.e., $\mathbf{u}\in S^{\sigma^\mathrm{obs}}_\mathbf{u}(\hat{\mathbf{x}},\mathbf{x}^\mathrm{obs}) \implies \text{Pr} (\mathbf{x}_i, \mathbf{x}^{\mathrm{obs}}_o\in \mathcal{H}^\mathrm{obs}_{i,o}) \geq \sigma^\mathrm{obs},\forall i,o$). Readers are referred to \cite{luo2020multi} for detailed derivations of Eq.~(\ref{Prsbc}).

Given the similar structure of pairwise inter-robot safety and communication distance constraints in Eq.~(\ref{eq:h_safe_per}) and Eq.~(\ref{eq:h_conn_per}), one can also obtain the set of pairwise connectivity control constraints in the form of $\mathcal{C}^{\sigma^\mathrm{c}}_\mathbf{u}(\hat{\mathbf{x}},\mathcal{G}^\mathrm{slos})$ as follows to enforce chance-constrained communication distance condition (i.e., $\mathbf{u}\in \mathcal{C}^{\sigma^\mathrm{c}}_\mathbf{u}(\hat{\mathbf{x}},\mathcal{G}^\mathrm{slos}) \implies \text{Pr} (\mathbf{x}_i, \mathbf{x}_j\in \mathcal{H}^\mathrm{c}_{i,j}) \geq \sigma^\mathrm{c},\forall (v_i,v_j)\in\mathcal{E}^\mathrm{slos}$) for every pairwise robots in a given LOS communication spanning graph $\mathcal{G}^\mathrm{slos}$. 
\begin{align}
\label{Prcbc}
\mathcal{C}^{\sigma^\mathrm{c}}_\mathbf{u}(\hat{\mathbf{x}},\mathcal{G}^\mathrm{slos}) = \{\mathbf{u} \in \mathbb{R}^{qN} | &B^{\sigma^\mathrm{c}}_{i,j}\mathbf{u}\leq f^{\sigma^\mathrm{c}}_{i,j}, \forall (v_i,v_j)\in\mathcal{E}^\mathrm{slos}, \notag \\
&B^{\sigma^\mathrm{c}}_{i,j} \in \mathbb{R}^{2d\times qN}, f^{\sigma^\mathrm{c}}_{i,j} \in \mathbb{R}^{2d}\}
\end{align}
where $B^{\sigma^\mathrm{c}}_{i,j}$ and $f^{\sigma^\mathrm{c}}_{i,j}$ are determined by observed robot states, distribution of system noise, and confidence level $\sigma^\mathrm{c}\in(0,1)$. See discussion about Eq.~(\ref{Prcbc}) in Section~\ref{app:eq:prcbc}.

Considering the occlusion-free condition in Eq.~(\ref{eq:h_los}), in Section~\ref{Sec:PrLOS_CBC} we first present a novel analytical form of occlusion-free condition. 
Based on that, we further propose a novel notion of Probabilistic Line-of-Sight Connectivity Barrier Certificates (PrLOS-CBC) as the control constraint for the pairwise robots to preserve the occlusion-free condition with satisfying probability.

\subsection{Problem Statement} \label{sec:problem}
We assume that the team of robots $\mathcal{S}$ is allocated $M$ simultaneous tasks ($M\leq N$) by $M$ subgroups $\mathcal{S}=\{\mathcal{S}_1,\ldots,\mathcal{S}_M\}$. 
Each robot is assigned to one of the subgroups $\mathcal{S}_{m},\forall m=1,\ldots,M$ with the individual task-related nominal controller $\mathbf{u}_i=\tilde{\mathbf{u}}_i\in \mathbb{R}^q$. 
To ensure overall task performance and efficient inter-robot collaboration within each subgroup,
the LOS connectivity should be preserved both globally and at the subgroup level \cite{luo2020behavior}. Here we formally define the required two levels of LOS connectivity as follows.
\begin{definition}\label{def:graph}
\textbf{Global LOS Connectivity}: A graph $\mathcal{G}^\mathrm{los}$ is \textit{LOS connected} if there exists at least one occlusion-free path between every pair of vertices on the graph. \textbf{Subgroup LOS Connectivity}: A graph $\mathcal{G}^\mathrm{los}$ is \textit{Subgroup LOS connected} if there exists at least one occlusion-free path between every pair of vertices in each induced LOS subgroup graph $\mathcal{G}^\mathrm{los}_m = \mathcal{G}^\mathrm{los}[\mathcal{V}_m]\subseteq\mathcal{G}^\mathrm{los}, \forall m = 1,...,M$,   where $\mathcal{V}_m\subseteq \mathcal{V}$ includes all robots within the same subgroup.    
\end{definition}

\noindent\textbf{Assumption:} \textit{Without loss of generality, we assume that the communication range $R_\mathrm{c}$ is much larger than the inter-robot safety distance $R_\mathrm{s}$ and obstacle-robot safety distance $R_\mathrm{obs}$, i.e., $R_\mathrm{c} \gg R_\mathrm{s}, R_\mathrm{obs}$.}\\

Given the real-time LOS communication graph $\mathcal{G}^\mathrm{los}$, the objective is to optimize the joint multi-robot controller $\mathbf{u}\in\mathbb{R}^{qN}$ such that (i) the global and subgroup LOS connectivity of the resulting LOS communication graph will be preserved, and (ii) the deviation of the LOS constrained joint controller $\mathbf{u}$ for all robots will be minimized from their nominal task-related joint controller $\tilde{\mathbf{u}}$. We assume $\mathcal{G}^\mathrm{los}$ satisfies the global and subgroup LOS connectivity initially. Then, the problem can be formally defined as follows.
\begin{align}
 &\mathbf{u}^* = \argmin_{\mathcal{G}^\mathrm{slos},\mathbf{u}} \sum_{i=1}^{N}||\mathbf{u}_i-\tilde{\mathbf{u}}_i||^2 \label{eq:rawobj}\\
 \text{s.t.} &\quad \mathcal{G}^\mathrm{slos}=(\mathcal{V},\mathcal{E}^\mathrm{slos})\subseteq \mathcal{G}^\mathrm{los} \label{eq:rawglobal}
 \\ & \quad\mathcal{G}^\mathrm{slos}_m=\mathcal{G}^\mathrm{slos}[\mathcal{V}_m],\; \forall m=1,...,M \label{eq:rawconn} \\
 &\quad \text{where $\mathcal{G}^\mathrm{slos}$ and all $\mathcal{G}^\mathrm{slos}_m$ are LOS connected and} \notag\\ & \text{edges in $\mathcal{G}^\mathrm{slos}$ stay LOS connected with high probability.} \notag\\
&\quad \mathbf{u}\in\mathcal{S}^{\sigma^\mathrm{s}}_\mathbf{u}(\mathbf{\hat{x}})\bigcap\mathcal{S}^{\sigma^\mathrm{obs}}_\mathbf{u}(\hat{\mathbf{x}},\mathbf{x}^\mathrm{obs}), \notag\\
&\quad ||\mathbf{u}_i|| \leq \alpha_i,\forall i = 1,...,N \label{eq:rawconst}
\end{align}

In Section~\ref{sec:tree}, we will reformulate problem Eq.~(\ref{eq:rawobj}) as a standard bilevel optimization form.
The optimization problem Eq.~(\ref{eq:rawobj}) involves the lower-level task in Eq.~(\ref{eq:rawglobal}) and Eq.~(\ref{eq:rawconn}) of finding an optimal subset of edges from $\mathcal{G}^\mathrm{los}$ forming globally and subgroup LOS connected $\mathcal{G}^\mathrm{slos}$ to preserve, and the upper-level task in Eq.~(\ref{eq:rawobj}) of minimizing the accumulative deviations from the predefined task-related nominal controllers $\tilde{\mathbf{u}}_i,i=1, \ldots, N$. This allows for co-optimization between   
connectivity constraints to enforce and the overall control modification for staying as close to nominal robots' behaviors.
Due to the noisy observation of robots' positional information, we will discuss in Section~\ref{sec:method_new} how to derive deterministic robot controllers so that the selected $\mathcal{G}^\mathrm{slos}$ will remain LOS connected with satisfying probability.

\section{Method}\label{sec:method_new}
\subsection{Probabilistic Line-of-Sight Connectivity Barrier Certificates (PrLOS-CBC)}\label{Sec:PrLOS_CBC}
Considering the occlusion-free condition defined in Eq.~(\ref{eq:h_los}) alongside the noisy observations of robots' positions, deriving an analytical expression for $h^\mathrm{los}_{i,j}(\mathbf{x},\mathcal{C}_\text{obs})$ that quantifies the satisfaction of the condition $h^\mathrm{los}_{i,j}(\mathbf{x},\mathcal{C}_\text{obs})\geq 0$ is a non-trivial task.
Hence, we propose to use approximation methods such as the ellipsoidal representation to characterize the satisfying occlusion-free condition for
LOS communication edge $(v_i,v_j)\in\mathcal{E}^\mathrm{los}$ and to analytically determine with satisfying probability $\sigma^\mathrm{los}\in(0,1)$ whether any obstacle is intersecting with this edge.
To prevent overly conservative approximation, we formulate the ellipsoidal approximation as a Minimum Volume Covering Ellipsoid (MVCE) \cite{mittal2021finding} circumscribing the edge $(v_i,v_j)\in\mathcal{E}^\mathrm{los}$ as well as the $\sqrt{\sigma^\mathrm{los}}$-confidence error ellipsoids $\mathcal{Q}^{\sqrt{\sigma^\mathrm{los}}}_i,\mathcal{Q}^{\sqrt{\sigma^\mathrm{los}}}_j$ defined by the Gaussian distributions of the noisy positions of the robots $i,j$.

Given the observed positions and uncertainty covariance of pairwise robots ${\mathbf{x}}_i\sim \mathcal{N}({\hat{\mathbf{x}}}_i,{\Sigma_i}), {\mathbf{x}}_j\sim \mathcal{N}({\hat{\mathbf{x}}}_j,{\Sigma_j})$ that are initially LOS connected and the corresponding $\sqrt{\sigma^\mathrm{los}}$-confidence error ellipsoids $\mathcal{Q}^{\sqrt{\sigma^\mathrm{los}}}_i,\mathcal{Q}^{\sqrt{\sigma^\mathrm{los}}}_j$,  then the probability of the two robots' states both lie in those two $\sqrt{\sigma^\mathrm{los}}$-confidence error ellipsoids is $\sigma^\mathrm{los}$. 
With this, one can compute the instantaneous MVCE centered at $\hat{\mathbf{p}}^0_{i,j}=\frac{\hat{\mathbf{x}}_i+\hat{\mathbf{x}}_j}{2}\in\mathbb{R}^d$ as $\mathcal{Q}^{\sigma^\mathrm{los}}_{i,j}(Q^{\sigma^\mathrm{los}}_{i,j}, \hat{\mathbf{p}}^0_{i,j})=\{\mathbf{p}\in\mathbb{R}^d|(\mathbf{p}-\hat{\mathbf{p}}^0_{i,j})^TQ^{\sigma^\mathrm{los}}_{i,j}(\mathbf{p}-\hat{\mathbf{p}}^0_{i,j})\leq 1\}$ 
over the set of $\sigma^{\sqrt{\mathrm{los}}}$-confidence error ellipsoids $\{\mathcal{Q}^{\sqrt{\sigma^\mathrm{los}}}_i,\mathcal{Q}^{\sqrt{\sigma^\mathrm{los}}}_j\}$ \cite{mittal2021finding}, 
where $Q^{\sigma^\mathrm{los}}_{i,j}= \argmin_{\{Q^{\sigma^\mathrm{los}}_{i,j}\}} \text{det} (Q^{\sigma^\mathrm{los}}_{i,j})^{-1}$ subjects to 
$\mathcal{Q}^{\sqrt{\sigma^\mathrm{los}}}_i,\mathcal{Q}^{\sqrt{\sigma^\mathrm{los}}}_j\subset \mathcal{Q}^{\sigma^\mathrm{los}}_{i,j}$ and $Q^{\sigma^\mathrm{los}}_{i,j}\succ0$. 
Note that the approximated ellipsoid characterized by $Q^{\sigma^\mathrm{los}}_{i,j}$ will update as robots $i,j$ move over time.

Then the function of $h_{i,j}^\mathrm{los}$ for occlusion-free condition under uncertainty and its 0-superlevel set $\mathcal{H}_{i,j}^\mathrm{los}$ in Eq.~(\ref{eq:h_los}) could be analytically re-defined as follows. 
{\footnotesize
\begin{align}
        &h_{i,j,o}^\mathrm{los}(\hat{\mathbf{x}},\mathbf{x}^\mathrm{obs}) = (\mathbf{x}^\mathrm{obs}_o-\hat{\mathbf{p}}^{0}_{i,j})^{T}Q^{\sigma^\mathrm{los}}_{i,j}(\mathbf{x}^{\mathrm{obs}}_o-\hat{\mathbf{p}}^{0}_{i,j})-1, \forall (v_i,v_j)\in\mathcal{E}^\mathrm{los}\notag\\
       &\mathcal{H}_{i,j,o}^\mathrm{los}=\{{\hat{\mathbf{x}}}\in \mathbb{R}^{d N}|h^\mathrm{los}_{i,j,o}(\hat{\mathbf{x}},\mathbf{x}^\mathrm{obs})\geq 0\;,\; \forall (v_i,v_j)\in \mathcal{E}^\mathrm{los} ,\forall o\} \notag\\
       &\mathcal{H}_{i,j}^\mathrm{los} = \bigcap_{\forall o} \mathcal{H}^\mathrm{los}_{i,j,o}  \label{los1}
\end{align}}

Following Lemma~\ref{lem:cbf} and its probabilistic extension in \cite{luo2020multi}, we now formally define Probabilistic Line-of-Sight Connectivity Barrier Certificates (PrLOS-CBC) as follows.
\begin{lemma}
\label{loscbcdefinition}
\textbf{Probabilistic Line-of-Sight Connectivity Barrier Certificates (PrLOS-CBC):}
Given a LOS communication spanning graph $\mathcal{G}^\mathrm{slos}=(\mathcal{V},\mathcal{E}^\mathrm{slos})$, a desired set $\mathcal{H}^\mathrm{los}(\mathcal{G}^\mathrm{slos})$ in Eq.~(\ref{eq:hlosset}) with $h^\mathrm{los}_{i,j,o}$ from Eq.~(\ref{los1}), and a user-defined high probability ${\sigma^\mathrm{los}}\in(0,1)$, for any Lipschitz continuous controller $\mathbf{u}$, the Probabilistic Line-of-Sight connectivity barrier certificates (PrLOS-CBC) as admissible control space $ \mathcal{C}^{\sigma^\mathrm{los}}_\mathbf{u}(\hat{\mathbf{x}},\mathcal{C}_\mathrm{obs},\mathcal{G}^\mathrm{slos})$ 
defined below enforces system state to stay in 
$\{\bigcap_{\{v_i,v_j \in \mathcal{V}:(v_i,v_j)\in\mathcal{E}^\mathrm{slos}\}} \mathcal{H}^\mathrm{los}_{i,j}\}$ with high probability (by enforcing pairwise LOS separately with satisfying $\sigma^\mathrm{los}$): 
{\footnotesize
\begin{align}\label{eq:prlos_cbc}
&\mathcal{C}^{\sigma^\mathrm{los}}_\mathbf{u} (\hat{\mathbf{x}},\mathcal{C}_\mathrm{obs},\mathcal{G}^\mathrm{slos}) = \{\mathbf{u}\in\mathbb{R}^{qN}: \\
&\dot{h}^\mathrm{los}_{i,j,o}(\hat{\mathbf{x}}, \mathbf{x}^\mathrm{obs},\sigma^\mathrm{los},\mathbf{u})+\gamma h^\mathrm{los}_{i,j,o}(\hat{\mathbf{x}},\mathbf{x}^\mathrm{obs},\sigma^\mathrm{los})\geq 0,\forall (v_i,v_j)\in \mathcal{E}^\mathrm{slos},\forall o\} \notag
\end{align}}
where $\dot{h}_{i,j,o}^\mathrm{los}(\hat{\mathbf{x}},\mathbf{x}^\mathrm{obs}, \mathbf{u}) = -(\mathbf{x}^\mathrm{obs}_{o}-\frac{\hat{\mathbf{x}}_i+\hat{\mathbf{x}}_j}{2})^{T}Q^{\sigma^\mathrm{los}}_{i,j}(F_{i,j}(\mathbf{x})+G_{i,j}(\mathbf{x})\mathbf{u}_{ij})$, $F_{i,j}(\mathbf{x}) = F_{i}(\mathbf{x}_i)+F_{j}(\mathbf{x}_j)$ and $G_{i,j}(\mathbf{x})\mathbf{u}_{ij} = G_{i}(\mathbf{x}_i)\mathbf{u}_{i}+ G_{j}(\mathbf{x}_j)\mathbf{u}_{j}$. 
\end{lemma}

See detailed proofs and computation of $\mathcal{C}^{\sigma^\mathrm{los}}_\mathbf{u}(\hat{\mathbf{x}},\mathcal{C}_\mathrm{obs},\mathcal{G}^\mathrm{slos})$ in Section~\ref{app:sec:prooflemma2}. 

\begin{remark}\label{remark:ecbf}
   The computation of Probabilistic Line-of-Sight Connectivity Barrier Certificates (PrLOS-CBC) proposed in Lemma~\ref{loscbcdefinition} is mainly valid for systems with relative degree 1 when CBFs can be directly applied. To extend PrLOS-CBC on higher-order dynamics, one can employ variants of CBF such as Exponential Control Barrier Function (ECBF) \cite{nguyen2016exponential} or High Order Control Barrier Functions (HOCBFs) \cite{xiao2021high} to ensure the forward invariance of $\mathcal{H}^\mathrm{los}_{i,j}$.
\end{remark}

\subsection{Uncertainty-Aware Line-of-Sight Least Constraining Tree (ULOS-LCT)}\label{sec:tree}
We first characterize the desired graph $\mathcal{G}^\mathrm{slos} =\mathcal{G}^\mathrm{slos*}$ in Eq.~(\ref{eq:rawglobal}) and Eq.~(\ref{eq:rawconn}) to maintain at each time step. This selected optimal graph $\mathcal{G}^\mathrm{slos*}=(\mathcal{V},\mathcal{E}^\mathrm{slos*})\subseteq \mathcal{G}^\mathrm{los}$ should (a) have the least number of edges for $\mathcal{G}^\mathrm{slos*}$ to be globally and subgroup LOS connected (i.e., spanning trees of $\mathcal{G}^\mathrm{los}$), and (b) introduce minimum motion constraints around $\tilde{\mathbf{u}}$ for robots to keep edges in $\mathcal{E}^\mathrm{slos*}$ LOS connected, implying that nominal multi-robot controllers $\tilde{\mathbf{u}}$ are \textit{least likely} violating constraints $\mathcal{C}^{\sigma^\mathrm{c}}_\mathbf{u}(\hat{\mathbf{x}},\mathcal{G}^\mathrm{slos*})\cap\mathcal{C}^{\sigma^\mathrm{los}}_\mathbf{u}(\hat{\mathbf{x}},\mathcal{C}_\mathrm{obs},\mathcal{G}^\mathrm{slos*})$ prescribed by $\mathcal{G}^\mathrm{slos*}$ compared to other spanning graph $\mathcal{G}^\mathrm{slos}$ 
at each time step. 
Thus, we formally define the class of candidate $\mathcal{G}^\mathrm{slos}$ (trees) below with redefined edge weights to quantify the levels of violation if following $\tilde{\mathbf{u}}$ in order to find $\mathcal{G}^\mathrm{slos*}$.
\begin{definition}\label{def:uccst}
Given the current LOS communication graph $\mathcal{G}^\mathrm{los}=(\mathcal{V},\mathcal{E}^\mathrm{los})$, we can define the weights $\mathcal{W}'=\{w'_{i,j}\}$ for all edges in $\mathcal{E}^\mathrm{los}$ as follows, where $w^\mathrm{d+los}_{i,j} = w^\mathrm{los}_{i,j} + w^\mathrm{d}_{i,j}$ with occlusion condition weights $w^\mathrm{los}_{i,j}=\frac{1}{L}\sum_{o=1}^{L}({{\dot{h}_{i,j,o}^\mathrm{los}(\hat{\mathbf{x}},\mathbf{x}^\mathrm{obs},\tilde{\mathbf{u}})+\gamma h_{i,j,o}^\mathrm{los}(\hat{\mathbf{x}},\mathbf{x}^\mathrm{obs})}})$ and connectivity condition weight $w^\mathrm{d}_{i,j}=f^{\sigma^\mathrm{c}}_{i,j}-B^{\sigma^\mathrm{c}}_{i,j}\tilde{\mathbf{u}}$.
\begin{align}
w'_{i,j} = \left\{ \begin{gathered}
  -\beta\cdot w^\mathrm{d+los}_{i,j}, \text{if} \; \text{$v_i$ and $v_j$ are in the same sub-group } \hfill \\
 - w^\mathrm{d+los}_{i,j}, \text{if} \;\text{$v_i$ and $v_j$ are in different sub-groups} \hfill \\
\end{gathered} \label{eq:neww} \right.
\end{align}
where $B^{\sigma^\mathrm{c}}_{i,j}$ and $f^{\sigma^\mathrm{c}}_{i,j}$, are from Eq.~(\ref{Prcbc}). $\beta\in\{\beta \gg 1| -\beta \cdot w_{i,j}\ll -w_{i',j'},\forall v_i,v_i',v_j,v_j' \in \mathcal{V}\}$ is a unique user-defined constant for the entire graph $\mathcal
{G}^\mathrm{los}$. Then given the weight-modified graph denoted as $\mathcal{G}'$, its spanning trees $\{\mathcal{T}_w^\mathrm{los'}=(\mathcal{V},\mathcal{E}^T,\mathcal{W}^{T'})\}$ are defined as Uncertainty-Aware Line-of-Sight Spanning Trees (ULOS-ST).
\end{definition}

The communication distance condition weight $w^\mathrm{d}_{i,j}$, occlusion-free condition weight $w^\mathrm{los}_{i,j}$, and the edge weight $w^\mathrm{d+los}_{i,j}$ as the sum of the two are heuristically used to quantify how likely the pairwise LOS constraints are to be violated under the task-related controller $\tilde{\mathbf{u}}_i, \tilde{\mathbf{u}}_j$.  
With the higher value of $w^\mathrm{d+los}_{i,j}$ (i.e., the smaller value of $-w^\mathrm{d+los}_{i,j}$), there is less total violation of the LOS communication constraints in Eq.~(\ref{Prcbc}) and Eq.~(\ref{eq:prlos_cbc}) for pairwise robots $i$ and $j$ under the task-related controllers $\tilde{\mathbf{u}}_i$ and $\tilde{\mathbf{u}}_j$, heuristically implying less disruption to the nominal controller if enforcing the LOS edge between $v_i,v_j$. 
To satisfy the subgroup LOS connectivity requirement in Eq.~(\ref{eq:rawconn}), the parameter $\beta$ is used to regulate intra-group edge weights as significantly smaller than inter-group edge weights. 
Hence, the optimal $\mathcal{G}^\mathrm{slos*}$ satisfying global and subgroup LOS connectivity is equivalent to the particular ULOS-ST with minimum total weight (proved below), and we define such a minimum weight ULOS-ST $\bar{\mathcal{T}}_w^\mathrm{los'}= \argmin_{\{\mathcal{T}_w^\mathrm{los'}\}} \sum_{(v_i,v_j)\in \mathcal{E}^{T}}\{w'_{i,j}\}$ as the \textbf{Uncertainty-Aware Line-of-Sight Least Constraining Tree} (ULOS-LCT).
\begin{theorem}\label{theorem:LCT}
The Uncertainty-Aware Line-of-Sight Least Constraining Tree (ULOS-LCT) $\bar{\mathcal{T}}_w^\mathrm{los'}$ is (i) globally and subgroup LOS connected, and (ii) least violated if following nominal joint controller $\mathbf{u}$ compared to all other candidate spanning trees in $\{\mathcal{T}_w^\mathrm{los'}=(\mathcal{V},\mathcal{E}^T,\mathcal{W}^{T'})\}$.
\end{theorem}

See detailed proof in Section~\ref{app:sec:proof_theorem4}. 

\begin{remark}\label{rem:graph}
The result of Theorem~\ref{theorem:LCT} answers an important question: which existing edges in $\mathcal{E}^\mathrm{los}$ from current multi-robot LOS graph $\mathcal{G}^\mathrm{los}$ should be maintained from now to the future (by constraining corresponding robots' motion), so that the resulting multi-robot LOS graph at future timesteps remains globally and subgroup LOS connected through at least these enforced edges that also introduce the least interruption on the original multi-robot tasks?
In other words, the computed ULOS-LCT  $\bar{\mathcal{T}}_w^\mathrm{los'}\subseteq \mathcal{G}^\mathrm{los}$ at each time step specifies such desired $\mathcal{G}^\mathrm{slos}\subseteq \mathcal{G}^\mathrm{los}$ that translates the high-level LOS graph connectivity constraints Eq.~(\ref{eq:rawglobal}) and Eq.~(\ref{eq:rawconn}) into edge-preserving for $\bar{\mathcal{T}}_w^\mathrm{los'}$ in favor of original task-related controllers.   

\end{remark}
With that, we can reformulate the original problem in Eq.~(\ref{eq:rawobj}) into the following standard bilevel optimization:
\begin{align}
 &\mathbf{u}^* = \argmin_{\mathcal{G}^\mathrm{slos},\mathbf{u}} \sum_{i=1}^{N}||\mathbf{u}_i-\tilde{\mathbf{u}}_i||^2 \label{eq:obj_final}\\
 \text{s.t.} \quad 
 &\mathcal{G}^\mathrm{slos}\leftarrow\bar{\mathcal{T}}_w^\mathrm{los'}= \argmin_{\{\mathcal{T}_w^\mathrm{los'}\}} \sum_{(v_i,v_j)\in \mathcal{E}^{T}}\{w'_{i,j}\}\\
&\mathbf{u}\in \mathcal{S}^{\sigma^\mathrm{s}}_\mathbf{u} \textstyle \bigcap 
\mathcal{S}^{\sigma^\mathrm{obs}}_\mathbf{u}
\bigcap \mathcal{C}^{\sigma^\mathrm{c}}_\mathbf{u}(\hat{\mathbf{x}},\mathcal{G}^\mathrm{slos})\bigcap \mathcal{C}^{\sigma^\mathrm{los}}_\mathbf{u}(\hat{\mathbf{x}},\mathcal{C}_\mathrm{obs},\mathcal{G}^\mathrm{slos}),\notag\\
&\quad ||\mathbf{u}_i|| \leq \alpha_i,\forall i=1,\ldots,N  \label{eq:const_final}
\end{align}

Note that the resulting ULOS-LCT $\bar{\mathcal{T}}_w^\mathrm{los'}$ may vary over time due to the dynamically changing real-time graph $\mathcal{G}^\mathrm{los}$. 
Hence, the optimal controller can be obtained by solving the QP problem in Eq.~(\ref{eq:obj_final}) following the construction of the $\bar{\mathcal{T}}_w^\mathrm{los'}$ using a standard Minimum Spanning Tree (MST) algorithm at each time step in a centralized manner. 
Next, we introduce our Dec-LOS-LCT algorithm that solves bi-level optimization Eq.~(\ref{eq:obj_final}) with optimal $\mathcal{G}^\mathrm{slos*}$ computed in a fully \textit{decentralized and interleaved manner}. 

\subsection{Decentralized Algorithm Design}\label{sec:dec-lct}
Define neighborhood set $\mathcal{P}_{i},\forall i\in [1,N]$ that contains all robots that can communicate with each robot $i$. We assume that robots can communicate and share noisy state and deterministic control information with all neighbors only if they are \textbf{LOS connected}. Denote $n_{i}$ as the number of neighbors for robot $i$, we thus define the local control variable $\mathbf{u}^{i} = [\mathbf{u}_i, (\mathbf{u}^{i}_j)_{j\in\mathcal{P}_i}]\in \mathbb{R}^{q(n_{i}+1)}$, noisy observation state information $\hat{\mathbf{x}}^{i}=[\hat{\mathbf{x}}_i, (\hat{\mathbf{x}}^{i}_j)_{j\in\mathcal{P}_i}]\in \mathbb{R}^{d(n_{i}+1)}$ and pre-assigned nominal task-related controller information  $\tilde{\mathbf{u}}^{i}=[\tilde{\mathbf{u}}^{i}, \;(\tilde{\mathbf{u}}^{i}_j)_{j\in\mathcal{P}_i}] \in \mathbb{R}^{q(n_{i}+1)}$ for each robot $i$, where $\mathbf{u}^{i}_j$, $\hat{\mathbf{x}}^{i}_j$ and $\tilde{\mathbf{u}}^{i}_{j} $ are the local copy of the $\mathbf{u}_j$, $\hat{\mathbf{x}}_j$ and $\tilde{\mathbf{u}}_{j}$ for robot $i$.  
For each robot $i$, denote the reformulated inequality constraints in Eq.~(\ref{eq:const_final}) based on its own local information as $\mathcal{SC}_i=\mathcal{S}_{\mathbf{u}^{i}}^{\sigma^{\mathrm{s}}}(\hat{\mathbf{x}}^{i})\cap \mathcal{S}_{\mathbf{u}^{i}}^{\sigma^{\mathrm{obs}}}(\hat{\mathbf{x}}^{i},\mathbf{x}^\mathrm{obs})\cap\mathcal{C}^{\sigma^\mathrm{c}}_{\mathbf{u}^{i}}(\hat{\mathbf{x}}^{i},\mathcal{G}^\mathrm{slos}_i)\cap\mathcal{C}^{\sigma^\mathrm{los}}_{\mathbf{u}^{i}}(\hat{\mathbf{x}}^{i},\mathcal{C}_\mathrm{obs},\mathcal{G}^\mathrm{slos}_i)$, where $\mathcal{G}^\mathrm{slos}_{i}$ is the partial of $\mathcal{G}^\mathrm{slos}$ from robot $i$'s view. With this, the centralized QP problem in Eq.~(\ref{eq:obj_final}) with given $\mathcal{G}^\mathrm{slos}$ can be written as a compact decentralized form as follows. 
\begin{equation}
\begin{aligned}
     &\mathbf{u}^{*} = \argmin_{\mathbf{u}} \sum^{N}_{i=1}{\sum_{j\in{\{i,\mathcal{P}_i\}}}}\left(\|\mathbf{u}^{i}_j-\tilde{\mathbf{u}}^i_j\|^2 \right. \left. +\mathcal{I}_{\mathcal{SC}_i}\right)\\
    &\text{ s.t. } \quad 
    \|\mathbf{u}^{i}_j\|\leq \alpha_i, \quad \forall j \in \{i,\mathcal{P}\}, \quad \forall i =1,...,N 
\end{aligned}
\label{eq:distributed_ustar}
\end{equation}
where $\mathcal{I}_\psi:\mathbb{R}^{qN}\mapsto \mathbb{R}$ is the indicator function that $\mathcal{I}_\psi(\mathbf{u}) = 0$ if $\mathbf{u} \in \psi$, otherwise  $\mathcal{I}_\psi(\mathbf{u}) = \infty$. 
Thus, the decomposed problem from Eq.~(\ref{eq:distributed_ustar}) to each robot $i$ becomes: 
\begin{align}
\label{eq:each_distributed_ustar}
   \mathbf{u}^{i*} =  &\argmin_{\mathbf{u}^i} {\sum_{j\in\{i,\mathcal{P}_i\}}\norm{\mathbf{u}^{i}_j-\tilde{\mathbf{u}}^i_j}^2} \\
    \text{ s.t. }  \quad 
   & \mathbf{u}^{i}\in \mathcal{SC}_i,\quad
    ||\mathbf{u}^{i}_j||\leq \alpha_i, \quad \forall j \in \{i,\mathcal{P}\} \notag
\end{align}

To enforce the consensus within the robot system, the augmented Lagrangian is defined as $\mathcal{L}^{i}_{\rho}(\mathbf{u}^{i},\mathbf{\bar{u}}^{i},\eta^{i}) =  OF_i  +(\eta^{i})^{\top}(\mathbf{u}^{i} - \mathbf{\bar{u}}^{i}) + \frac{\rho}{2}\norm{\mathbf{u}^{i} - \mathbf{\bar{u}}^{i}}^{2}_{2}$ \cite{boyd2011distributed}, where $OF_i = {\sum_{j\in\{i,\mathcal{P}_i\}}\norm{\mathbf{u}^{i}_j-\tilde{\mathbf{u}}^i_j}^2}$ and $\rho$ is the penalty parameter. $\eta^{i}=\rho(\mathbf{u}^{i} - \mathbf{\bar{u}}^{i})$ is the Lagrangian multiplier with $\mathbf{\bar{u}}^{i}$ as the global average variable which can be defined as:
\begin{align}
&\mathbf{\bar{u}}^{i} = [\mathbf{\bar{u}}_{i},(\mathbf{\bar{u}}_{j})_{j\in\mathcal{P}_i}]\in \mathbb{R}^{q(n_{i}+1)},\; \notag\\
&\mathbf{\bar{u}}_{i} = \frac{1}{n_i}\sum_{j\in\mathcal{P}_i}\mathbf{u}^{j}_i \in \mathbb{R}^{q} \label{eq:update_global}
\end{align}

The augmented Lagrangian incorporates a penalty term to ensure robot consensus. Finally, the decomposed problem in Eq.~(\ref{eq:each_distributed_ustar}) for each robot $i$ could be reformulated as:
\begin{align}
\label{eq:local_update}
  &\mathbf{u}^{i+}=\argmin_{\mathbf{u}^{i}}{\mathcal{L}^{i}_{\rho}(\mathbf{u}^{i},\mathbf{\bar{u}}^{i},\eta^{i})}\\
  \text{s.t.} \quad &\mathbf{u}^{i}\in{\mathcal{SC}_i},\quad
  ||\mathbf{u}^{i}_j||\leq \alpha_i, \quad \forall j \in \{i,\mathcal{P}\} \notag
\end{align}

In the standard iteration of C-ADMM \cite{boyd2011distributed}, each robot $i$ sends its local update $\mathbf{u}^{i+}_j$ from $\mathbf{u}^{i+}$ in Eq.~(\ref{eq:local_update}) to all neighbors $j\in\mathcal{P}_i$, and calculates the average variable $\mathbf{\bar{u}}_{i}^{+}$ according to Eq.~(\ref{eq:update_global}). Then it collects the $\mathbf{\bar{u}}^{+}_{j}$ and constructs the global average variable $\mathbf{\bar{u}}^{i+}$ Eq.~(\ref{eq:update_global}) from its neighbors $j\in\mathcal{P}_i$. At the end of each C-ADMM iteration, each robot will update the Lagrangian multiplier by $\eta^{i+} = \mathbf{u}^{i+}-\mathbf{\bar{u}}^{i+}$. The agreement feasibility $\norm{\mathbf{u}^{i}-\mathbf{\bar{u}}^{i}}_2$ is applied to determine the convergence during the C-ADMM iteration \cite{boyd2011distributed}. 

Next, we propose a novel decentralized approach in Algorithm~\ref{alg:dis_LCT} to solve the bi-level optimization problem Eq.~(\ref{eq:obj_final}) without specifying $\mathcal{G}^\mathrm{slos}$ beforehand. Note that our Algorithm~\ref{alg:dis_LCT} interleaves the two processes of finding ULOS-LCT $\bar{\mathcal{T}}_w^\mathrm{los'}$ and solving the resultant C-ADMM based optimization Eq.~(\ref{eq:local_update}) \emph{at each time step}.

In general, our algorithm starts with a graph where each node (robot) is isolated from each other. By following the Algorithm~\ref{alg:dis_LCT}, robots keep updating their LOS connectivity information (Line~\ref{alg:update_leader}-\ref{algorithm:update_control}). Once new control constraints are established, the robots are collaboratively solving the resultant optimization problem (Line~\ref{alg:u_initial}-\ref{alg:u_final}). Iteratively, the robots will join and form the optimal tree $\bar{\mathcal{T}}_w^\mathrm{los'}$ and reach the consensus on the local control variable $\mathbf{u}^{i}$. Then the converged $\mathbf{u}_i^*,\forall i=1, \cdots, N$ extracted from $\mathbf{u}^{i}$ is the solution of Eq.~(\ref{eq:obj_final}) that guarantees the safety and global and subgroup LOS connectivity with satisfying probability.

Specifically, at the beginning of Algorithm~\ref{alg:dis_LCT}, each robot initializes two adjacency matrices – the local adjacency matrix $Ad_i \in \mathbb{R}^{(n_{i}+1)\times(n_{i}+1)}$ and a local copy of the global adjacency matrix $Ad \in \mathbb{R}^{N\times N}$. $Ad_i$ records the connectivity within neighbors $j\in \mathcal{P}_i$, while $Ad$ records connectivity among the team. Each robot collects information from all neighbors to construct (a) local variables (i.e., $\hat{\mathbf{x}}^i$ and $\tilde{\mathbf{u}}^i$), and (b) safety constraints $\mathcal{S}_{\mathbf{u}^{i}}^{\sigma^{\mathrm{s}}}(\hat{\mathbf{x}}^{i})$ and $\mathcal{S}_{\mathbf{u}^{i}}^{\sigma^{\mathrm{obs}}}(\hat{\mathbf{x}}^{i},\mathbf{x}^\mathrm{obs})$. Each robot is assigned a unique ID, and the leader ID is initialized to be itself. Each robot evaluates the matrix $Ad$ to determine the LOS connectivity of the team. If the team is not LOS connected, each robot selects the leader among its fragment, which is a subtree of the MST. It then reports the \textit{Minimum-Weight Outgoing Edge (MWOE)} information to the leader, which is the least-weighted edge among all outgoing edges of a fragment linking a fragment's node to an external node. The leader selects a robot for connection and updates the fragment's global connectivity information. Upon establishing a new connection, each robot updates its leader ID, local adjacency matrix ($Ad_i$), and LOS connectivity constraints. In summary, the leader is required to guarantee consistency within each fragment and utilized to avoid unnecessary computation. The robot then achieves a consensus within its fragment and iteratively between different fragments, while selecting a new leader for the merged fragments. 
In this way, each robot is able to collectively construct the optimal tree $\bar{\mathcal{T}}_w^\mathrm{los'}$ (Line~\ref{algorithm:update_gaph}-\ref{algorithm:update_graph_local}) and establish the corresponding control constraints $\mathcal{C}^{\sigma^\mathrm{c}}_\mathbf{u}(\hat{\mathbf{x}},\mathcal{G}^\mathrm{slos}_i)\bigcap \mathcal{C}^{\sigma^\mathrm{los}}_\mathbf{u}(\hat{\mathbf{x}},\mathcal{C}_\mathrm{obs},\mathcal{G}^\mathrm{slos}_i)$ (Line~\ref{algorithm:update_control}) using local adjacency matrix ($Ad_i$), where $\mathcal{G}^\mathrm{slos}_i=(\bar{\mathcal{T}}_w^\mathrm{los'})_i$ is the optimal tree $\bar{\mathcal{T}}_w^\mathrm{los'} $ from each robot's view. With this, the robots can accordingly update their local control variable $\mathbf{u}^{i}$ based on the updated control constraints (Line~\ref{alg:u_initial}-\ref{alg:u_final}). The final goal is to form a single leader and achieve team-wide consensus on the global adjacency matrix ($Ad$), representing the optimal tree $\bar{\mathcal{T}}_w^\mathrm{los'}$. The algorithm terminates when global consensus is reached respecting safety and LOS connectivity constraints by the optimal tree $\bar{\mathcal{T}}_w^\mathrm{los'}$. The optimal controller $\mathbf{u}_i^{*}$ is then derived from the converged $\mathbf{u}^i$.  

\begin{algorithm}[t]
    \caption{Uncertainty-Aware Dec-LOS-LCT}
    \label{alg:dis_LCT}
    \begin{algorithmic}[1]
    \Input{$a_i$: adjacency edge weight list, $\tilde{\mathbf{u}}_i$: pre-assigned nominal task-related controller, and $\mathbf{\hat{x}}_i$: noisy state information.}
    \Output{$\mathbf{u}^{*}_i$ optimal controller for robot $i$}
    \Function{Dec-LOS-LCT }{$a_i$,
    $\mathbf{\hat{x}}_i$ and $\tilde{\mathbf{u}}_i$}
    \State $Ad_i$, $Ad$ $\gets$ empty adjacency matrix
    \State initialize $\bar{\mathbf{u}}^{i} = 0$, $\eta^i = 0$
    \State $\mathbf{\hat{x}}^i$ and $\tilde{\mathbf{u}}^i$ $\gets$ from all neighbors $j\in\mathcal{P}_i$
    \State{construct $\mathcal{S}_{\mathbf{u}^{i}}^{\sigma^{\mathrm{s}}}(\hat{\mathbf{x}}^{i})$ and $\mathcal{S}_{\mathbf{u}^{i}}^{\sigma^{\mathrm{obs}}}(\hat{\mathbf{x}}^{i},\mathbf{x}^\mathrm{obs})$ from neighbour message} \label{algorithm:update_safe}
    \While {$Ad$ is not connected}          \label{alg:outer_start}
    \State $leader\_id$ $\gets$ min($self\_id$, $neighbor\_id$) \label{alg:update_leader}
    \State {report the MWOE to leader}
    \If {$leader\_id$ is $self\_id$}
    \State connect with MWOE and update MWOE \label{algorithm:update_gaph}
    \State update $Ad$ and broadcast $Ad$ within the fragment \label{algorithm:update_ad}
    \EndIf
    \State update the $Ad_i$ according to $Ad$ \label{algorithm:update_graph_local}
    \State{compute $\mathcal{C}^{\sigma^\mathrm{c}}_{\mathbf{u}^{i}}(\hat{\mathbf{x}}^{i},\mathcal{G}^\mathrm{slos}_i)\cap\mathcal{C}^{\sigma^\mathrm{los}}_{\mathbf{u}^{i}}(\hat{\mathbf{x}}^{i},\mathcal{C}_\mathrm{obs},\mathcal{G}^\mathrm{slos}_i)$ from $Ad_i$} \label{algorithm:update_control}
    \While {not converged} \label{alg:inner_start}
      \State solve Eq.~(\ref{eq:local_update}) and send $\mathbf{u}^{i+}_{j}$ to all neighbors $j\in\mathcal{P}_i$ \label{alg:u_initial}
    \State collect $\mathbf{u}^{j+}_{i}$ ($j\in\mathcal{P}_i$), $\mathbf{\bar{u}}^{+}_{i}$ $\leftarrow$ Eq.~\eqref{eq:update_global}
    \State send $\mathbf{\bar{u}}^{+}_{i}$ to  neighbors $j\in\mathcal{P}_i$
    \State construct $\mathbf{\bar{u}}^{i+}$ $\leftarrow$ Eq.~\eqref{eq:update_global} 
    \State update $\eta^{i+} = \eta^{i} + \rho(\mathbf{u}^{i+} -\bar{\mathbf{u}}^{i+})$
    \State check convergence with agreement feasibility \label{alg:u_final}
    \EndWhile
    \State {update $\mathbf{u}^i$}
    \EndWhile
   \Return {$\mathbf{u}_i^{*}$ from $\mathbf{u}^i$}\label{alg:outer_end}
    \EndFunction
    \end{algorithmic}
\end{algorithm}

\subsection{Theoretic Analysis}
\begin{proposition}\label{proposition:graph}
By Algorithm~\ref{alg:dis_LCT}, each robot agrees with the same LOS communication graph $\mathcal{G}^\mathrm{slos}$, which is the real-time ULOS-LCT $\bar{\mathcal{T}}_w^\mathrm{los'}$.
\end{proposition}
See detailed proof in Section~\ref{app:sec:proof_proposition5}.
\begin{proposition}\label{proposition:guarantee}
By following Algorithm~\ref{alg:dis_LCT} at each time step, robots reach consensus regarding the admissible control space prescribed by $\mathcal{C}^{\sigma^\mathrm{c}}_\mathbf{u}(\hat{\mathbf{x}},\mathcal{G}^\mathrm{slos})\bigcap \mathcal{C}^{\sigma^\mathrm{los}}_\mathbf{u}(\hat{\mathbf{x}},\mathcal{C}_\mathrm{obs},\mathcal{G}^\mathrm{slos})$ where
$\mathcal{G}^\mathrm{slos}=\bar{\mathcal{T}}_w^\mathrm{los'}$, and 
the derived $\mathbf{u}_i^*,\forall i=1, \cdots, N$ is the solution of Eq.~(\ref{eq:obj_final}), rendering 
the resulting $\mathcal{G}^\mathrm{los} \supseteq\bar{\mathcal{T}}_w^\mathrm{los'}$ globally and subgroup LOS connected with satisfying probability at all times.
\end{proposition}
See detailed proof in Section~\ref{app:sec:proof_proposotion6}.
\begin{proposition}\label{proposition:probability}
By choosing $\sigma^\mathrm{los}=1-\frac{1-\sigma^\mathrm{graph}}{N-1}$ as the pair-wise robots occlusion-free confidence level to guarantee the occlusion-free condition for ULOS-LCT $\bar{\mathcal{T}}_w^\mathrm{los'}$, the resultant LOS communication graph $\mathcal{G}^\mathrm{los} \supseteq\bar{\mathcal{T}}_w^\mathrm{los'}$ satisfies the occlusion-free condition, ensuring at least one occlusion-free path between every pair of vertices with a probability greater than $\sigma^\mathrm{graph}$.
\end{proposition}
See detailed proof in Section~\ref{app:sec:proof_proposotion8}.

Proposition~\ref{proposition:graph} guarantees that robots converge to a uniform LOS graph $\bar{\mathcal{T}}_w^\mathrm{los'}$ via edge updates in Line~\ref{algorithm:update_gaph} of Algorithm~\ref{alg:dis_LCT}.
Proposition~\ref{proposition:guarantee} ensures that updating control constraints in Line~\ref{algorithm:update_control} of Algorithm~\ref{alg:dis_LCT} confines the robot team's control space to $\mathcal{C}^{\sigma^\mathrm{c}}_\mathbf{u}(\hat{\mathbf{x}},\bar{\mathcal{T}}_w^\mathrm{los'})\bigcap \mathcal{C}^{\sigma^\mathrm{los}}_\mathbf{u}(\hat{\mathbf{x}},\mathcal{C}_\mathrm{obs},\bar{\mathcal{T}}_w^\mathrm{los'})$ based on the optimal graph $\mathcal{G}^\mathrm{slos*}$. Hence, with Lemma~\ref{loscbcdefinition} all the robots remain globally and subgroup LOS connected (with satisfying probability) through preserving edges in $\bar{\mathcal{T}}_w^\mathrm{los'}$ overtime. Finally, if each pair-wise robot in $\bar{\mathcal{T}}_w^\mathrm{los'}$ satisfies the occlusion-free condition with $\sigma^\mathrm{los}=1-\frac{1-\sigma^\mathrm{graph}}{N-1}$ probability, Proposition~\ref{proposition:probability} guarantees that the resultant LOS communication graph has at least $\sigma^\mathrm{graph}\in (0,1)$ to satisfy the occlusion-free condition. 

\begin{remark} \label{remark:time_horizion}
[summarized from \cite{luo2020multi}] The probability of collision avoidance between two robots throughout an entire trajectory, denoted by $n_t$ time steps, is the product of independent probabilities at each step, with a lower bound of \( (\sigma^\mathrm{s})^{n_t} \). For longer trajectories, a discount factor $\zeta < 1$ could be introduced to moderate the step-wise threshold, ensuring the overall probability of collision avoidance remains within acceptable limits.
\end{remark}

\section{Results}
\begin{figure*}[h!]
\centering  
\begin{subfigure}[b]{0.28\textwidth}
  \includegraphics[width=\linewidth]{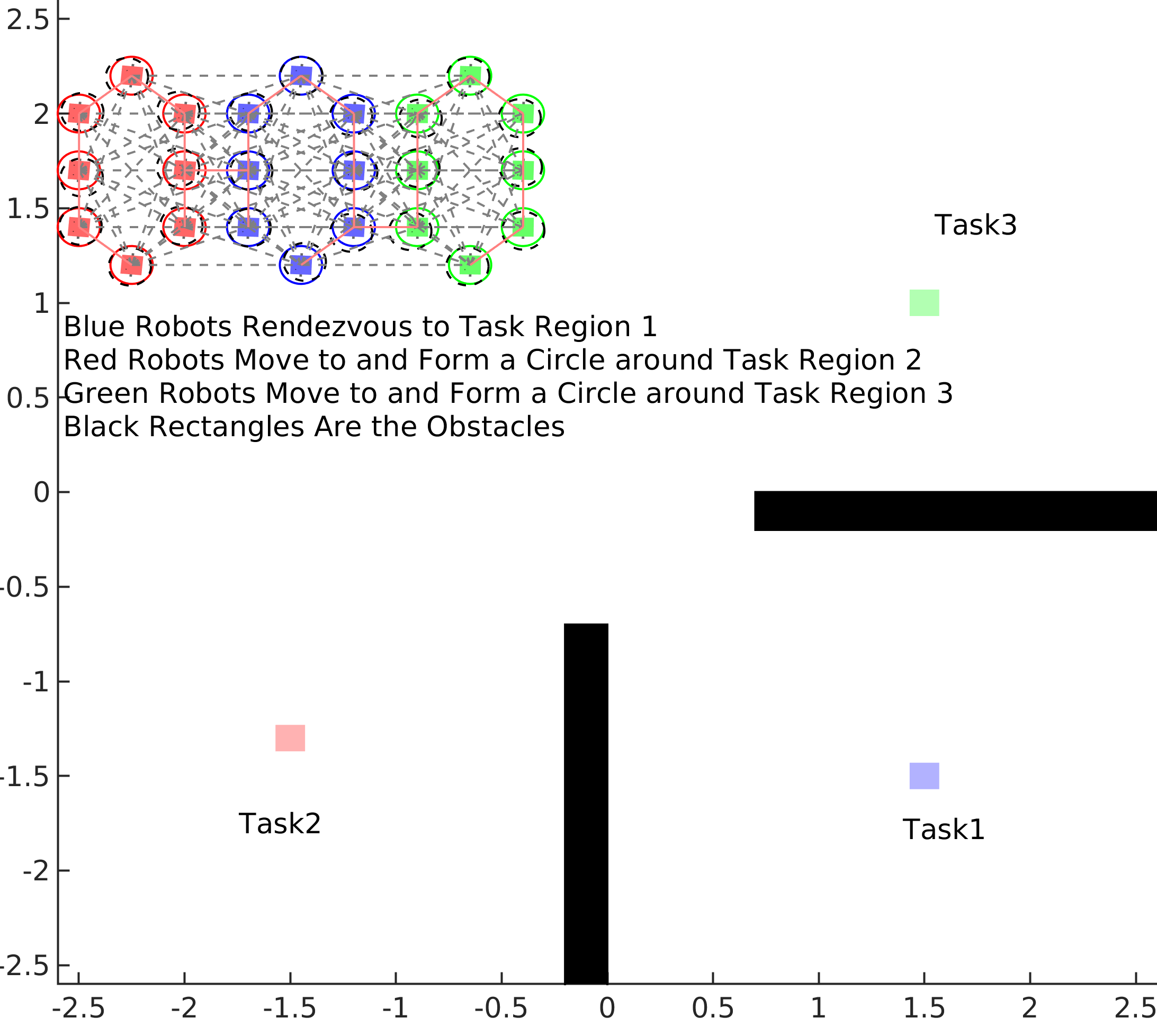}
 \caption{Our Method t = 0}
  \label{fig1:subfiga}
  \end{subfigure}
\begin{subfigure}[b]{0.28\textwidth}
  \includegraphics[width=\linewidth]{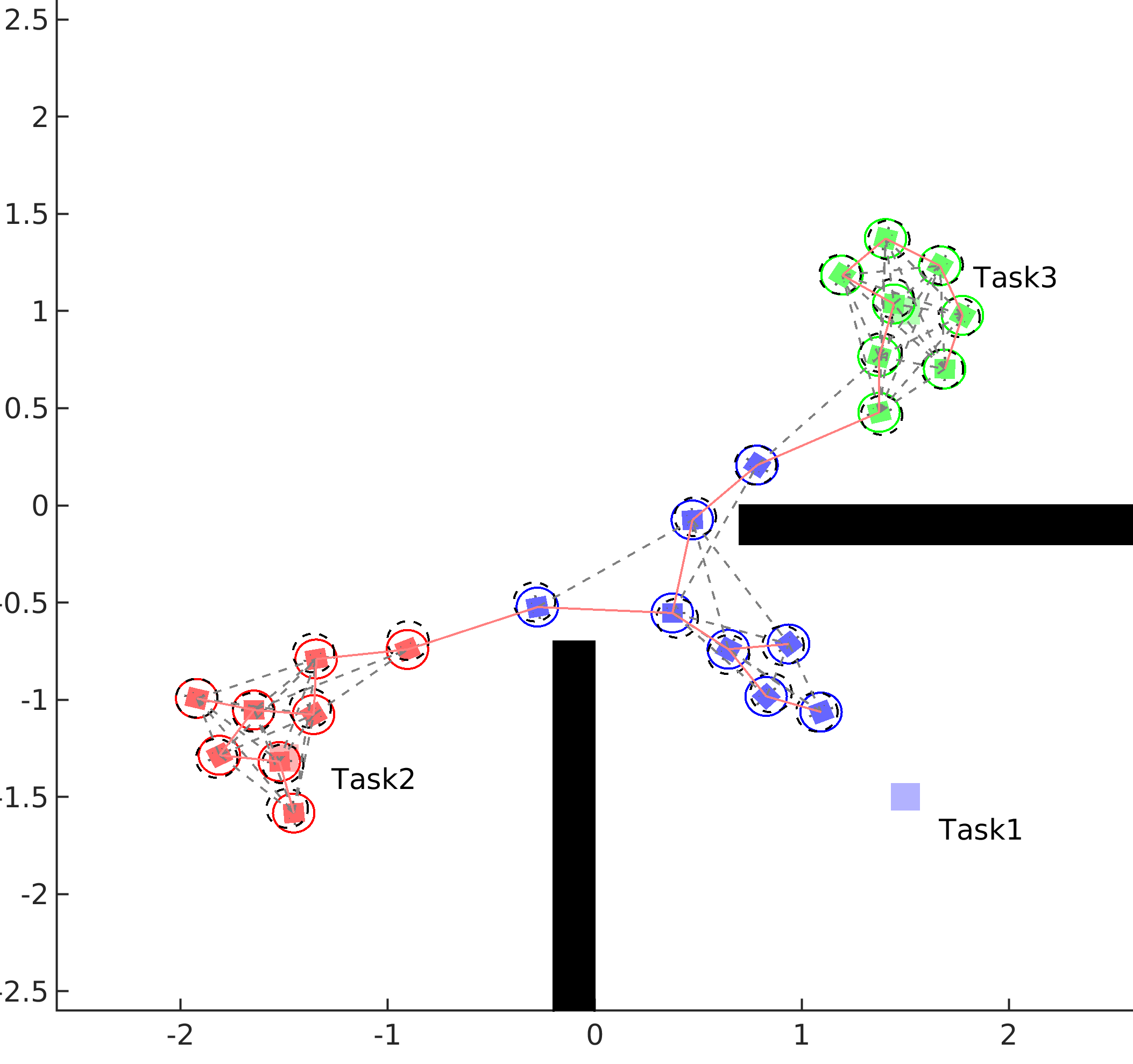}
 \caption{Our Method t = 1000}
  \label{fig1:subfigb}
   \end{subfigure}
   \begin{subfigure}[b]{0.28\textwidth}
\includegraphics[width=\linewidth]{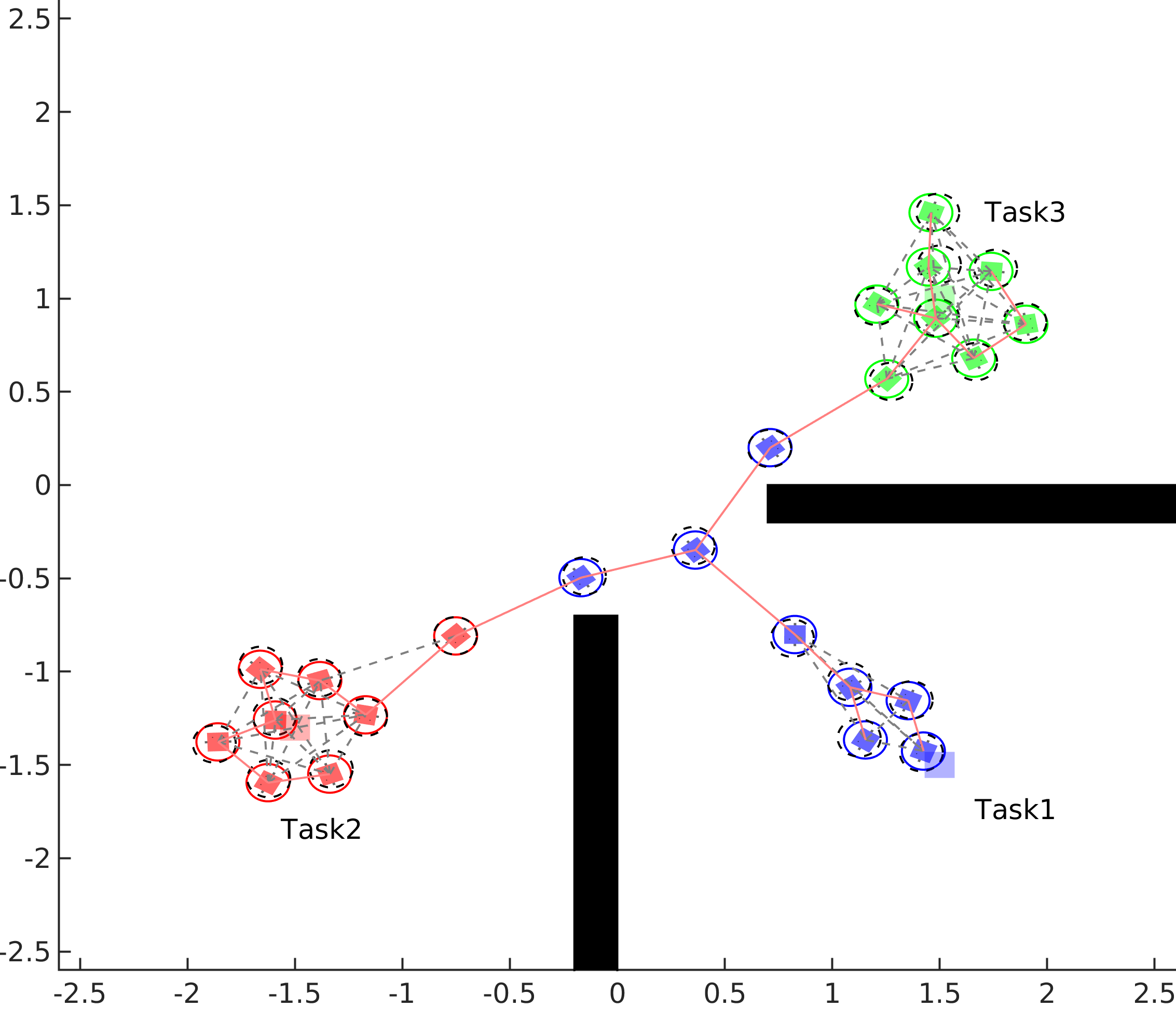}
 \caption{Our Method t = 2000 (Converged)}
  \label{fig1:subfigc}
  \end{subfigure}\\
\begin{subfigure}{0.24\textwidth}
\includegraphics[width=\linewidth]{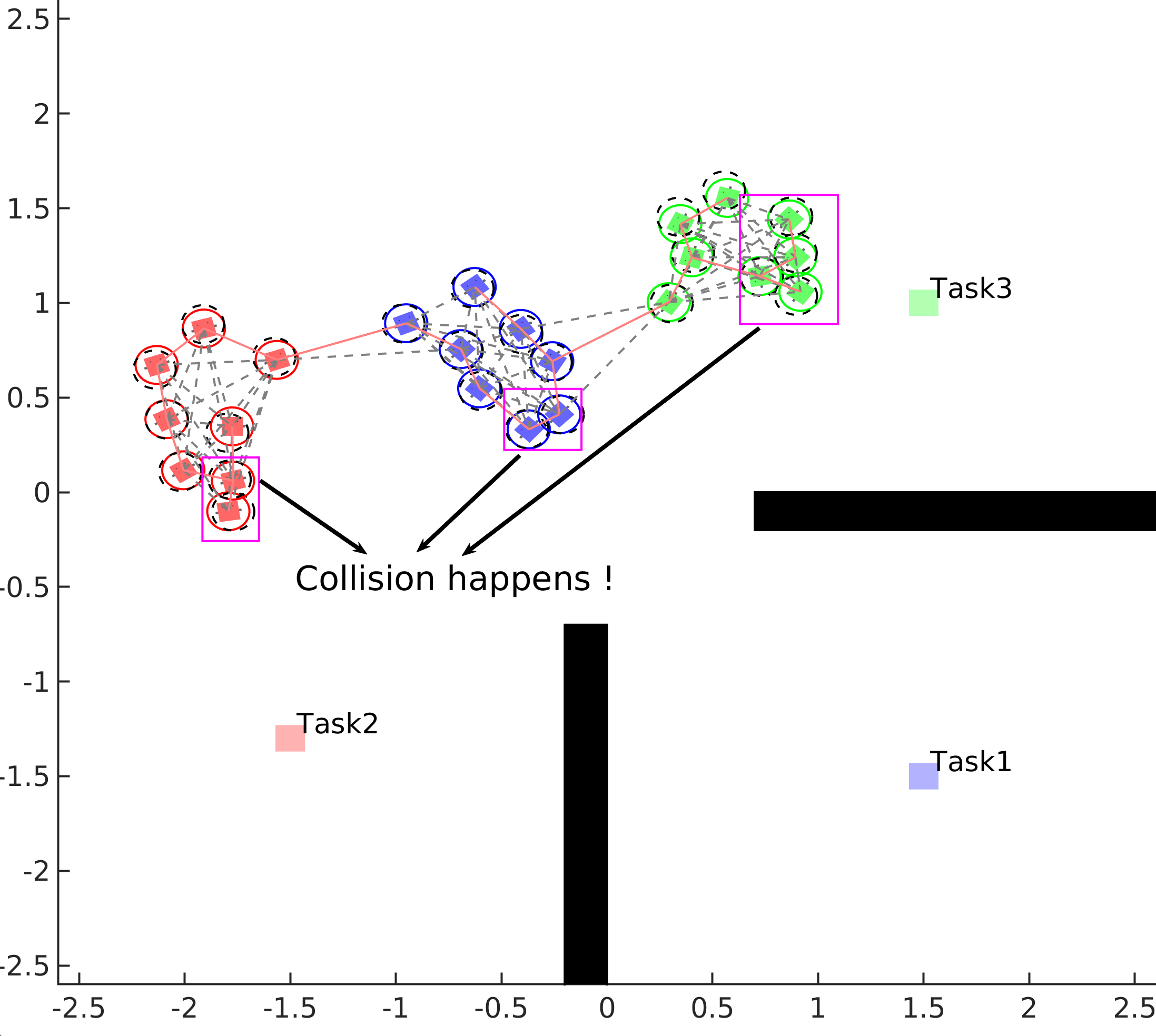}
\caption{}
  \label{fig2:subfigure1}
  \end{subfigure}
    \begin{subfigure}{0.24\textwidth}
\includegraphics[width=\linewidth]{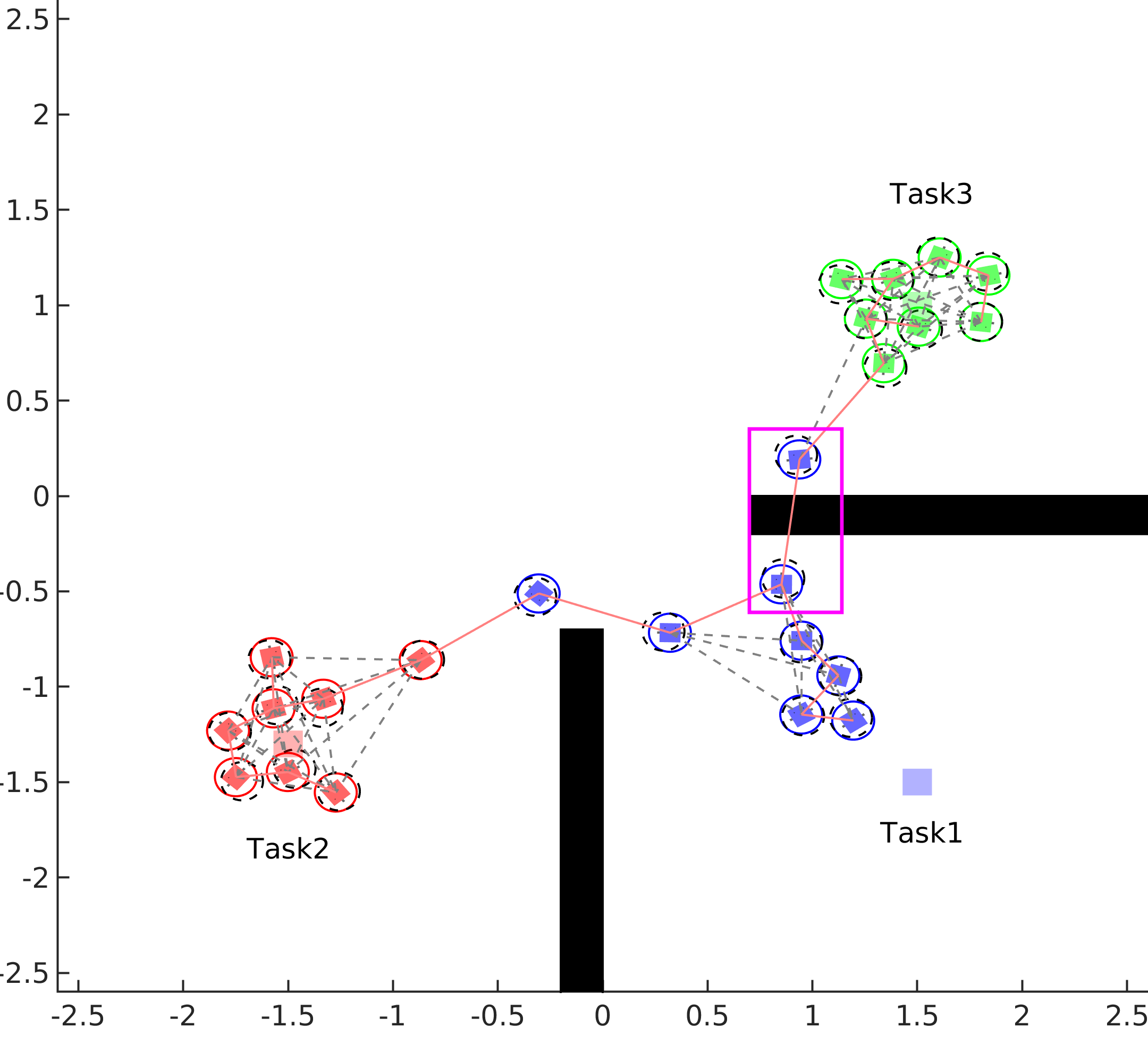}
 \caption{}
  \label{fig2:subfigure2}
  \end{subfigure}
  \begin{subfigure}{0.24\textwidth}
\includegraphics[width=\linewidth]{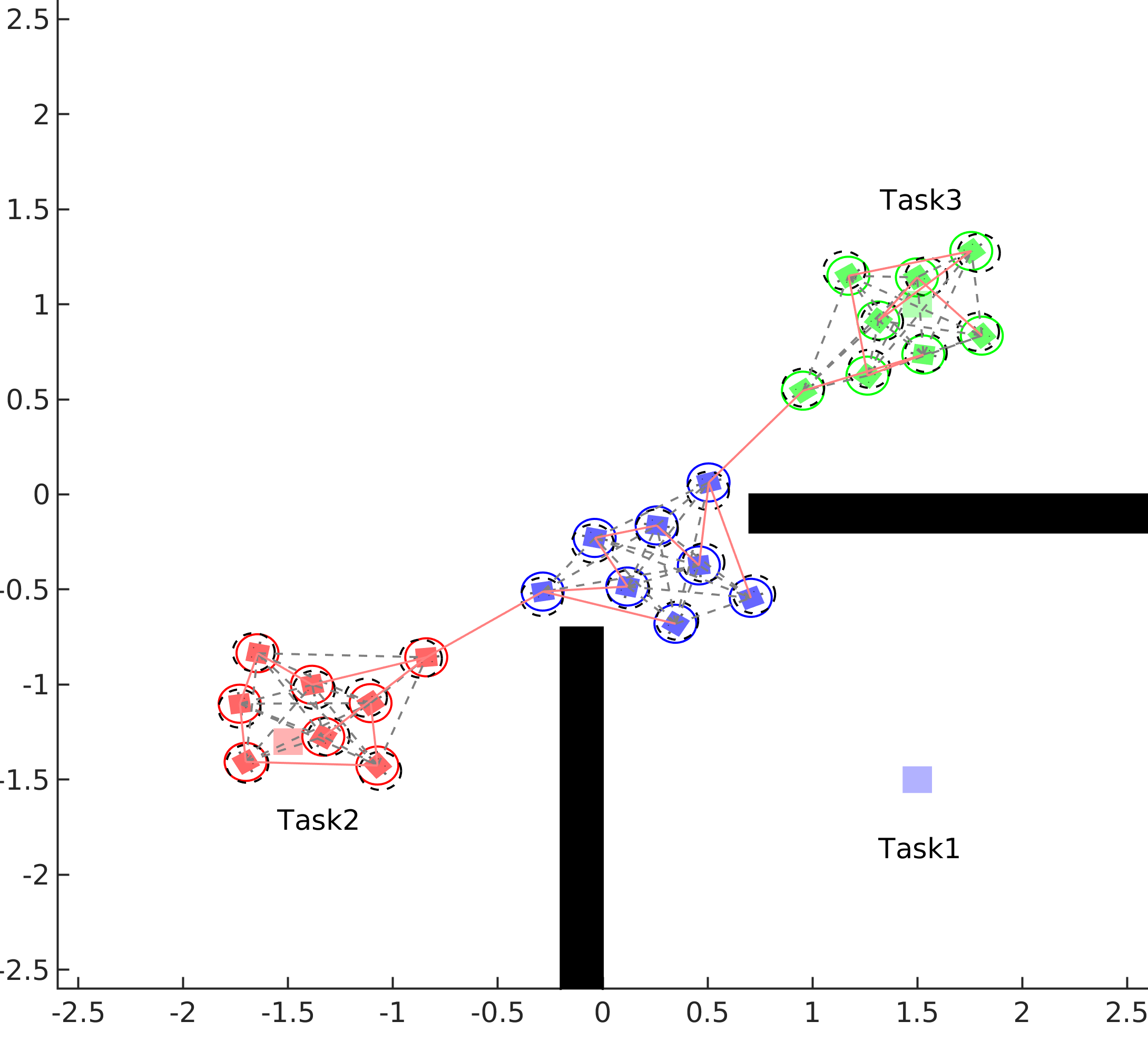}
\caption{}
  \label{fig2:subfigure3}
  \end{subfigure}
  \begin{subfigure}{0.24\textwidth}
\includegraphics[width=\linewidth]{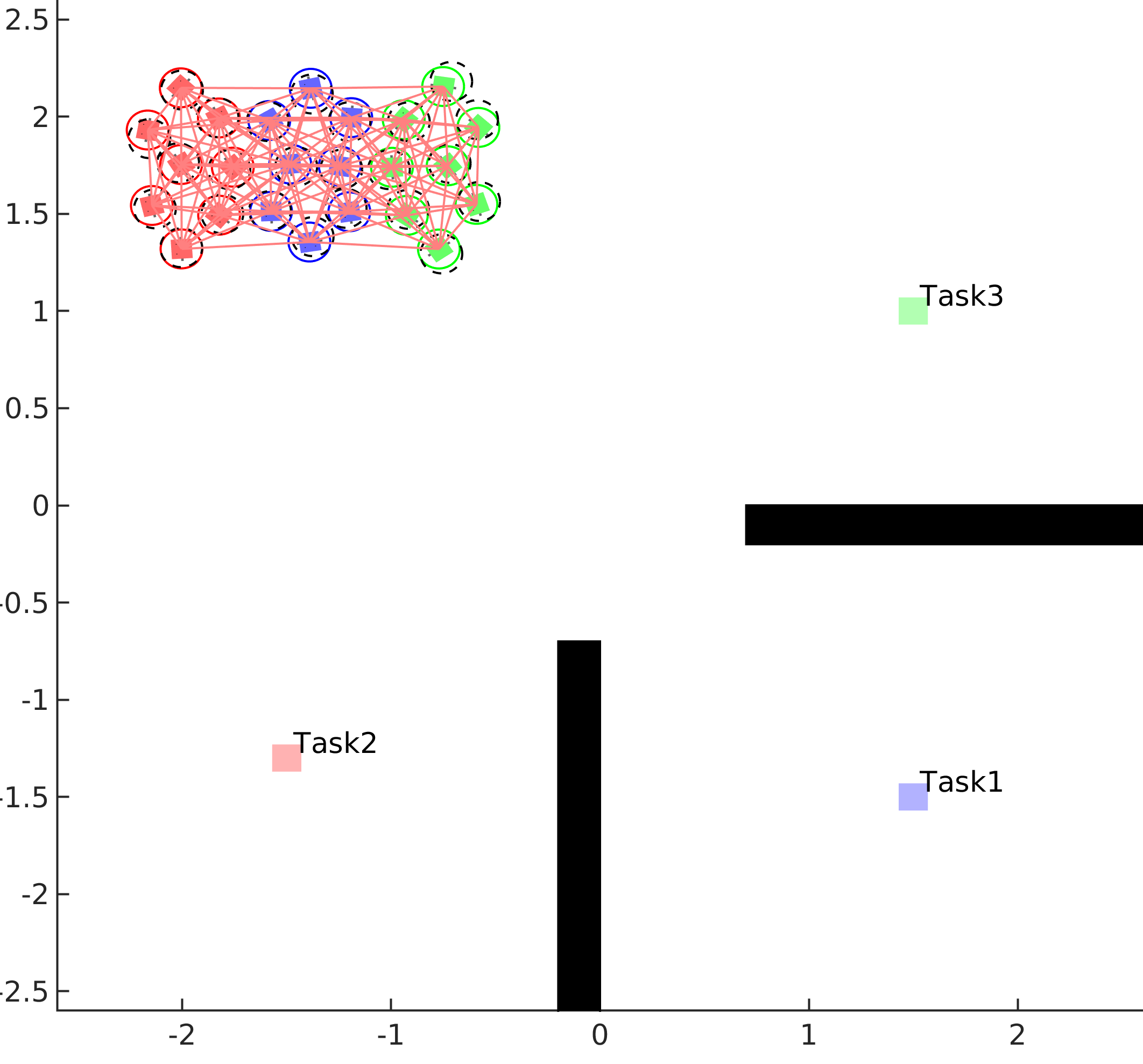}
 \caption{}
  \label{fig2:subfigure4}
  \end{subfigure}
  \caption{Simulation example of 24 robots divided into $M=3$ subgroups with different colors and tasked to three different places. Robots in blue subgroup 1 execute biased rendezvous behaviors towards the blue task site 1, while robots in red subgroup 2 and green subgroup 3 perform circle formation behaviors around the red task site 2 and green task site 3, respectively. The $R_\mathrm{s},\; R_\mathrm{obs}$ and $R_\mathrm{c}$ are 0.2 m, 0.2 m and 0.8 m in this experiment. The red lines in this figure denote the currently optimal LOS communication graph $\mathcal{G}^\mathrm{slos*}$ and the gray dash lines are the current LOS communication graph $\mathcal{G}^\mathrm{los}$. The black boxes represent the obstacles. The confidence level in this experiment is set as $\sigma^\mathrm{s}=\sigma^\mathrm{obs} = \sigma^\mathrm{c} = 0.90,\; \sigma^\mathrm{los} = 0.99\; (i.e.,\; \sigma^\mathrm{graph}=0.9 )$. The robot diameter is 0.16 m. The Multivariate Gaussian covariance matrix for measurement noise is  $\mathrm{diag}[0.03,0.04]$. Compared baseline methods include (d) MCCST \cite{luo2020behavior}, (e) Our method without considering occlusion avoidance, (f) Enforcing edges from fixed initial ULOS-LCT (red edges in Figure~\ref{fig1:subfiga}), and (g) Enforcing edges from fixed initial LOS connectivity graph (gray edges in Figure~\ref{fig1:subfiga}).}
  \label{fig:our_simulation}
\end{figure*}

\begin{figure*}[h!]
\centering
\begin{subfigure}[b]{0.24\textwidth}
  \includegraphics[width=\linewidth,height = 0.85\linewidth]{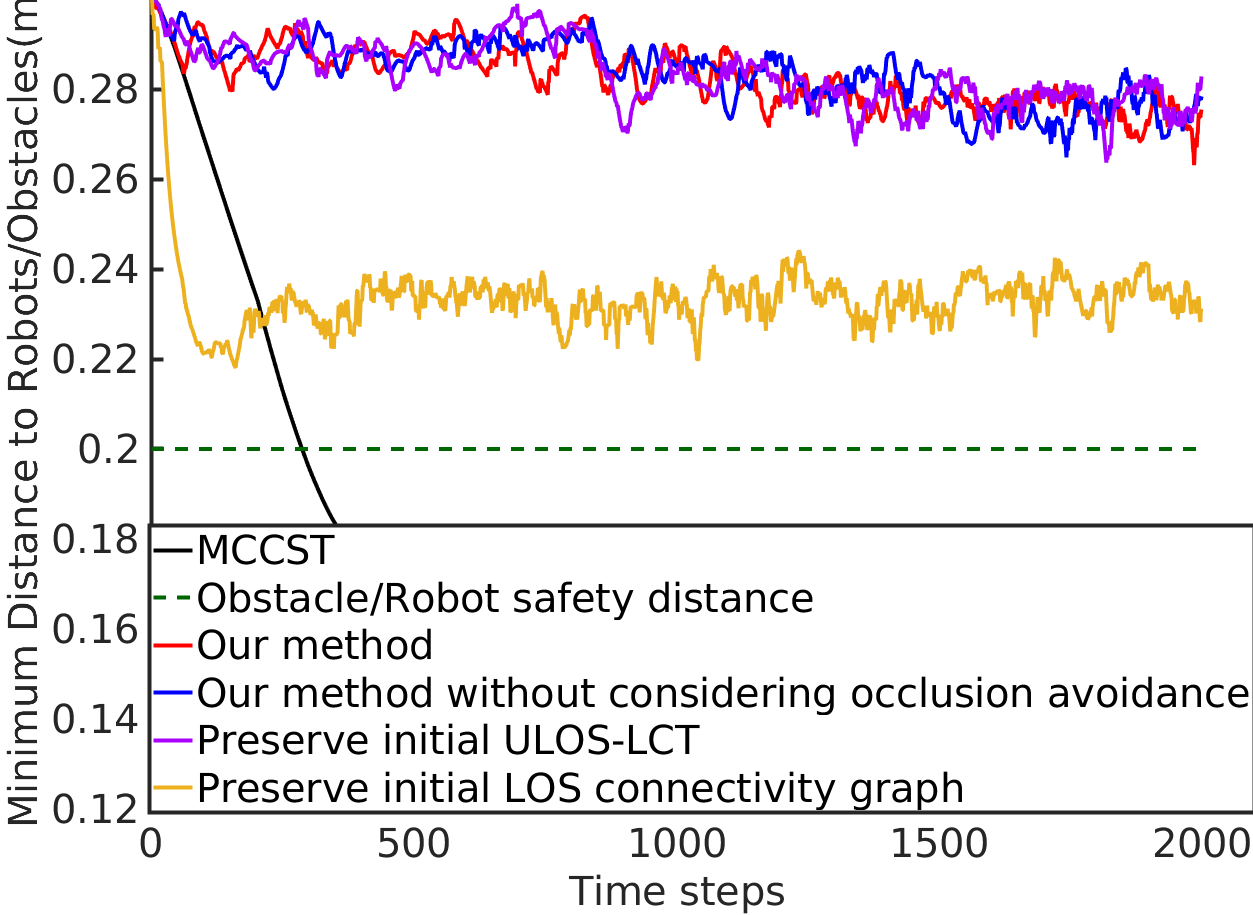}
 \caption{Minimum distance between robot/Obstacle}
  \label{fig3:subfigure1}
  \end{subfigure}
\begin{subfigure}[b]{0.24\textwidth}
  \includegraphics[width=\linewidth,height = 0.85\linewidth]{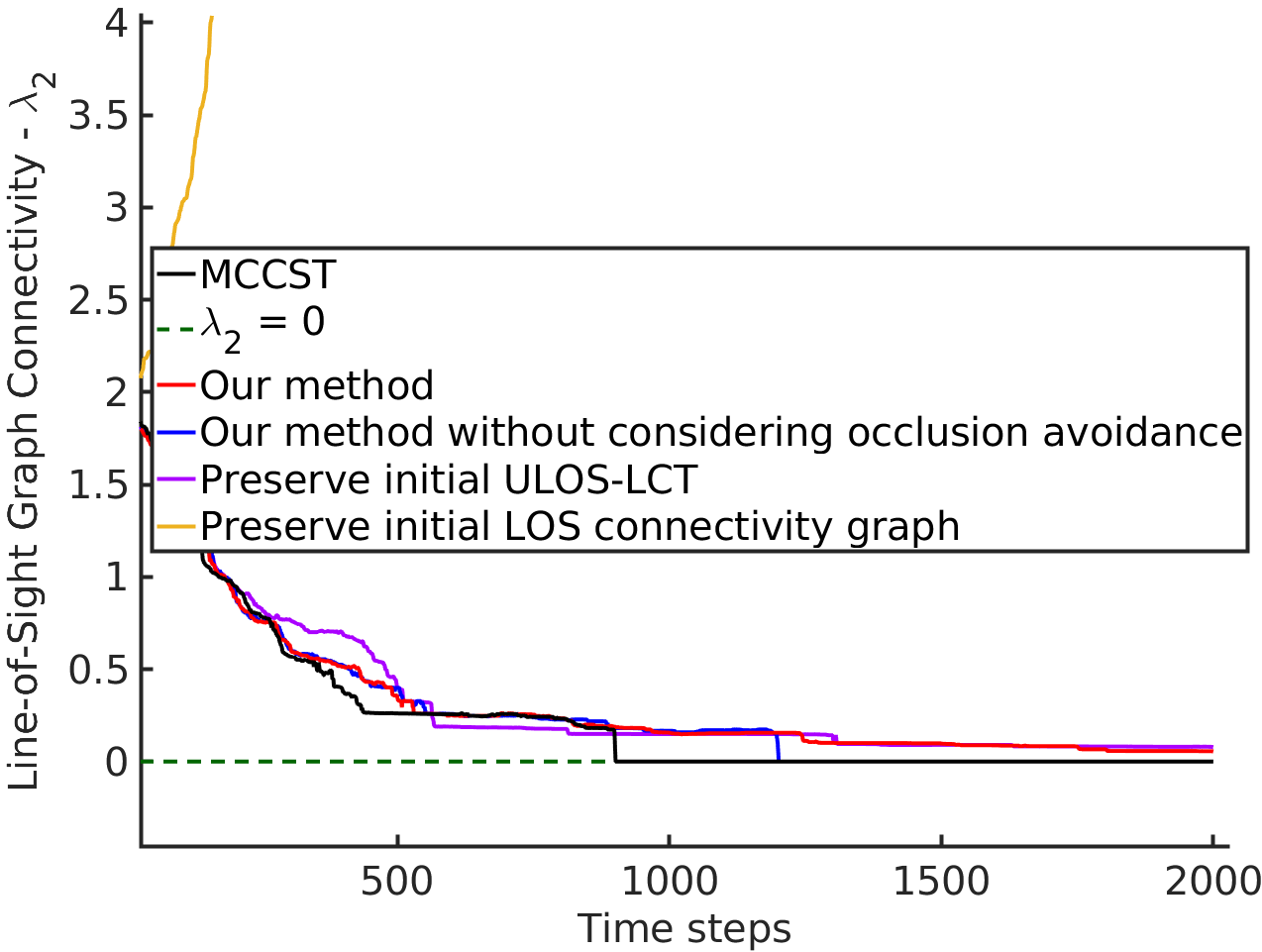}
 \caption{Algebraic LOS connectivity}
  \label{fig3:subfigure2}
\end{subfigure}
\begin{subfigure}[b]{0.24\textwidth}
  \includegraphics[width=\linewidth,height = 0.85\linewidth]{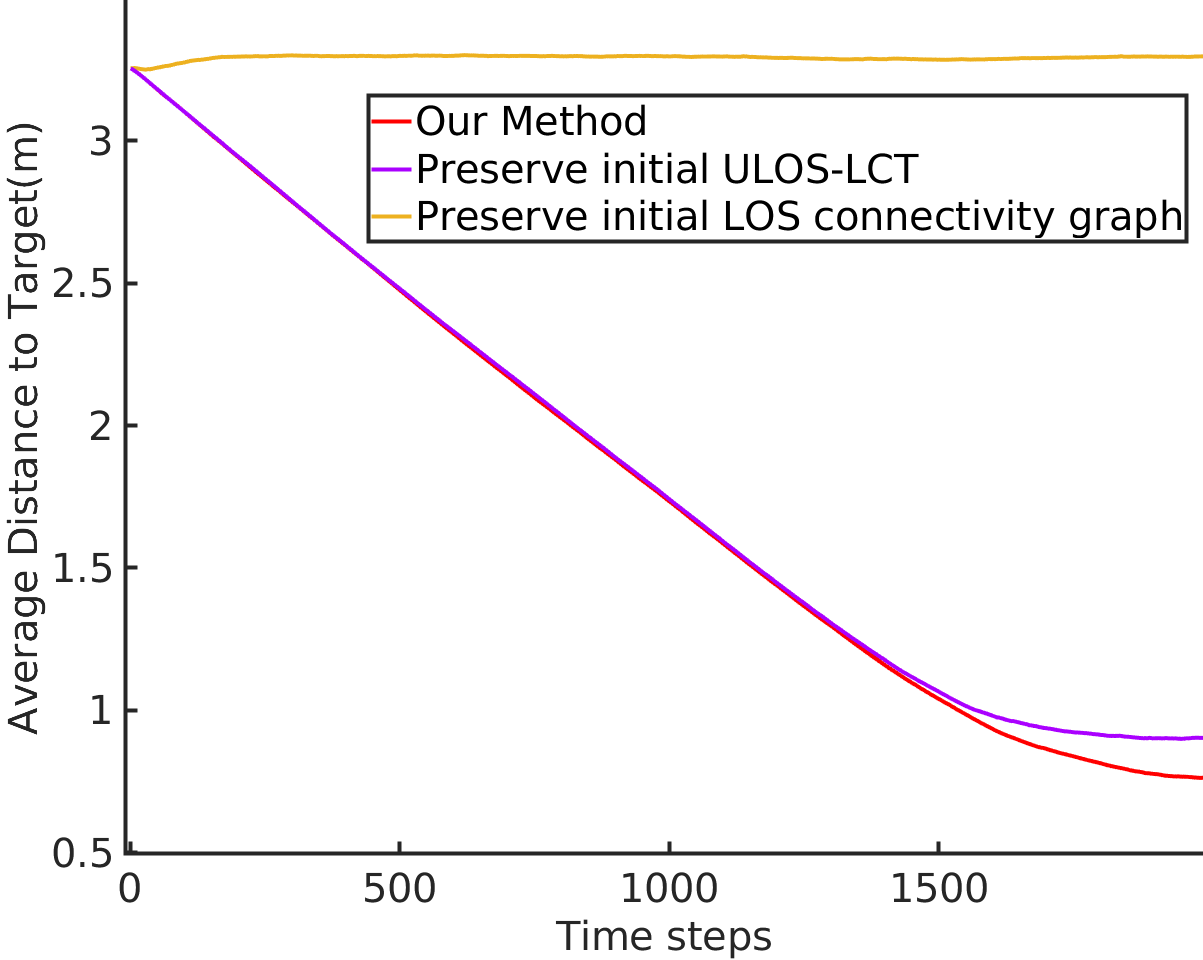}
 \caption{$D_{ave}$ to target region}
  \label{fig3:subfigure3}
   \end{subfigure}
\begin{subfigure}[b]{0.24\textwidth}
  \includegraphics[width=\linewidth,height = 0.85\linewidth]{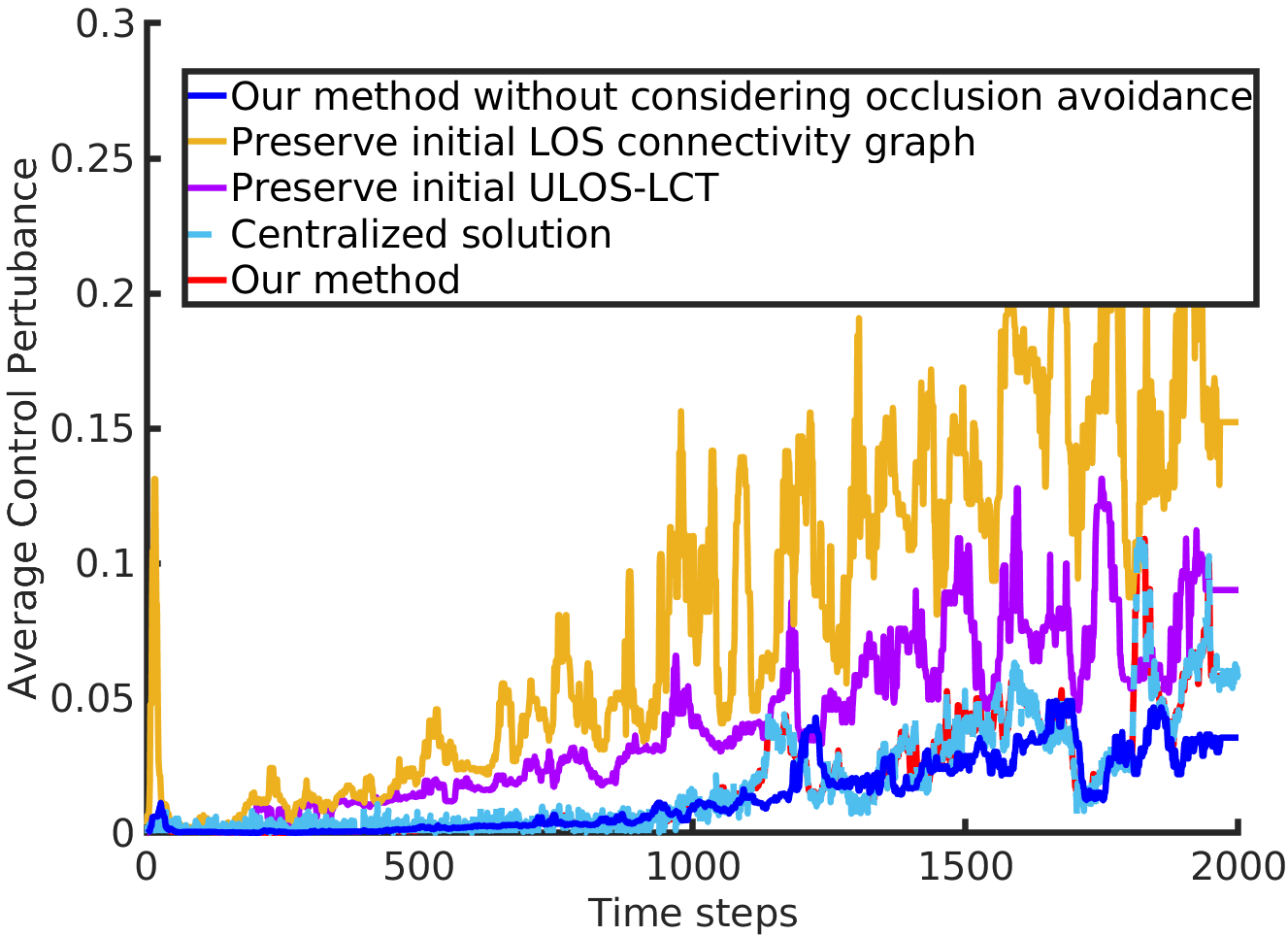}
 \caption{Average control perturbation}
  \label{fig3:subfigure4}
\end{subfigure}
  \caption{Performance comparison of the simulation example in Fig.~\ref{fig:our_simulation} w.r.t. different metrics: a) Minimum distance to robots/obstacles to verify the safety constraints' satisfaction, b) Average algebraic LOS connectivity to indicate whether the LOS graph is LOS connected ($\lambda_2 >0$) or not $\lambda_2 =0$, where $\lambda_2$ is the second-smallest eigenvalue of the LOS Laplacian matrix calculated from the LOS adjacency matrix. The elements in the LOS adjacency matrix indicate whether the pairwise robots are LOS connected), c) Average distance to target region to indicate the overall task efficiency, and (d) Average control perturbation (computed by $\frac{1}{N}\sum_{i=1}^N|| \mathbf{u}_i-\tilde{\mathbf{u}}_i||^{2}$ measuring the accumulated deviation from nominal controllers).}
  \label{fig:performance_comparsion}
\end{figure*}

\begin{figure*}[h!]
\centering  
\begin{subfigure}[b]{0.22\textwidth}
\centering
\includegraphics[width=\linewidth]{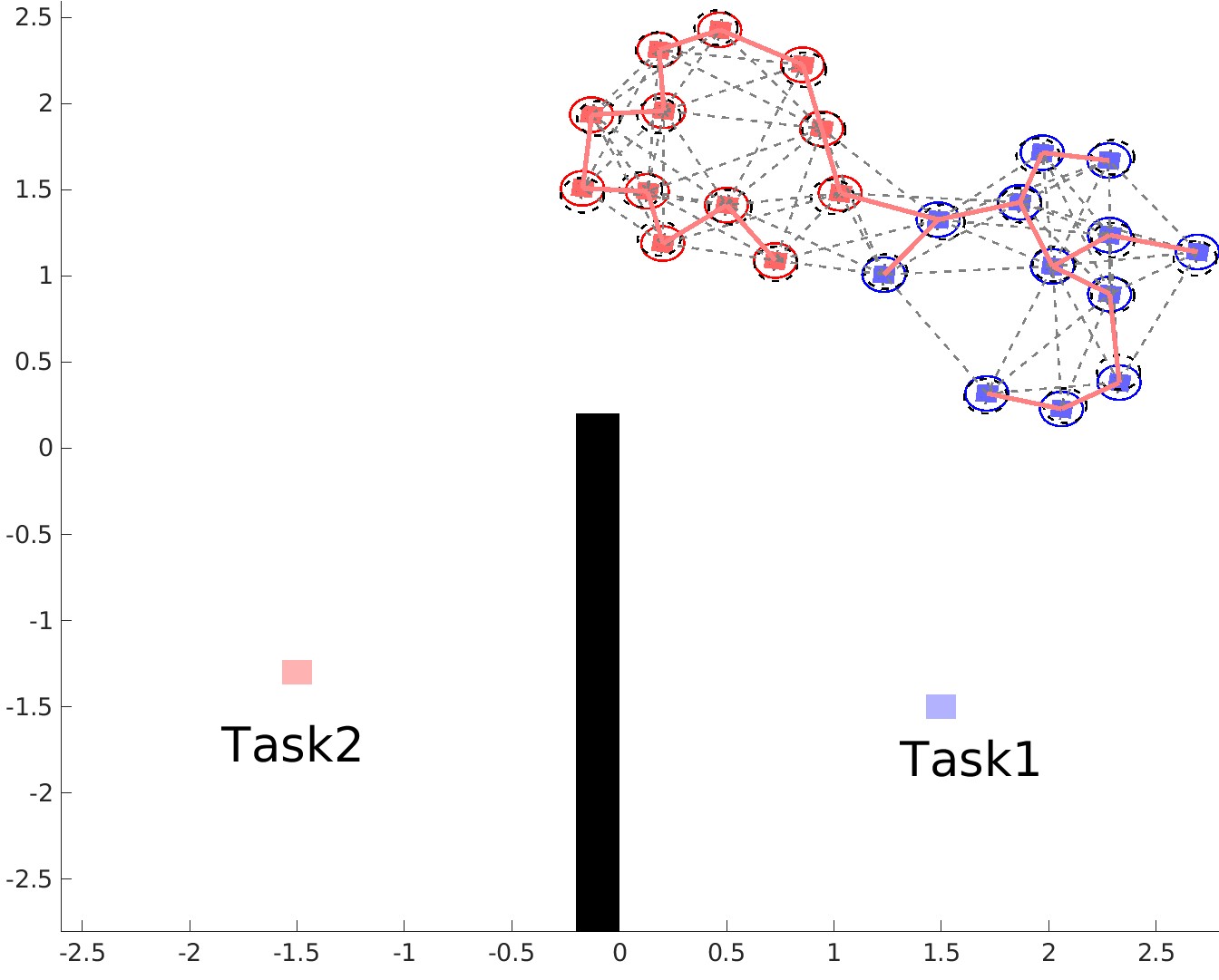}
  \caption{}
  \label{fig:init_5}
  \end{subfigure}
\begin{subfigure}[b]{0.22\textwidth}
\centering
  \includegraphics[width=\linewidth]{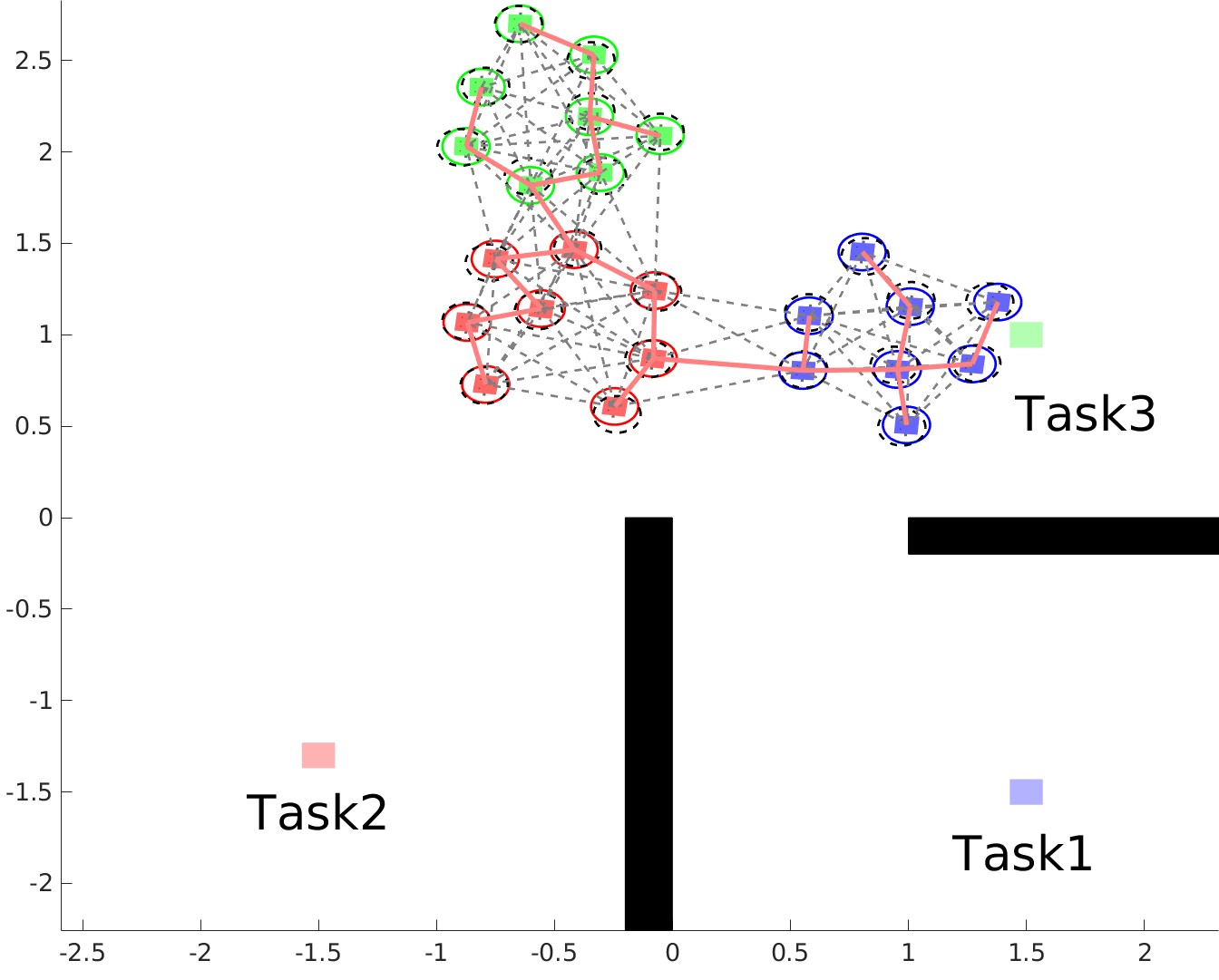}
 \caption{}
  \label{fig:init_2}
  \end{subfigure}
\begin{subfigure}[b]{0.22\textwidth}
\centering 
  \includegraphics[width=\linewidth]{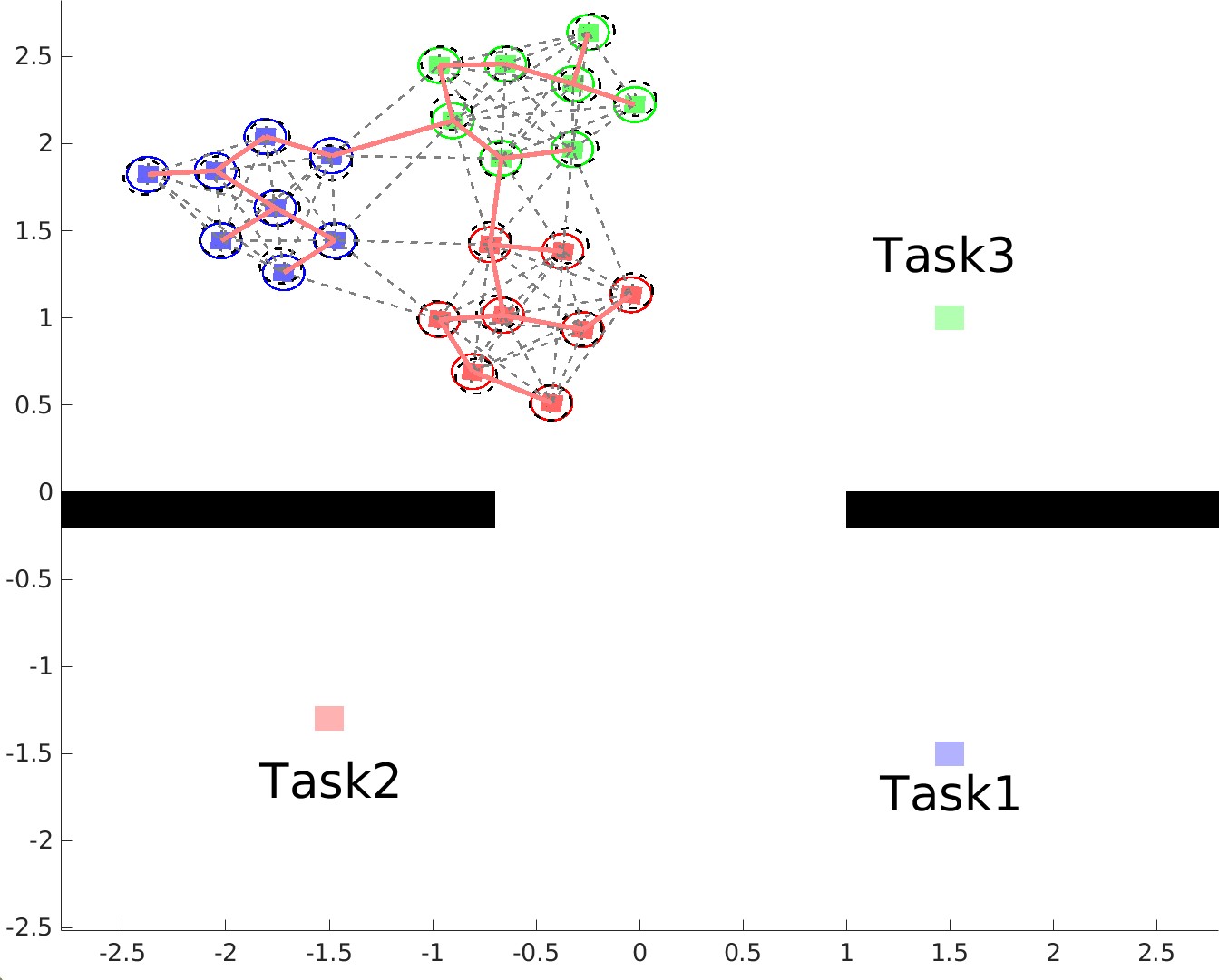}
 \caption{}
  \label{fig:init_3}
   \end{subfigure}
   \begin{subfigure}[b]{0.22\textwidth}
\centering
\includegraphics[width=\linewidth]{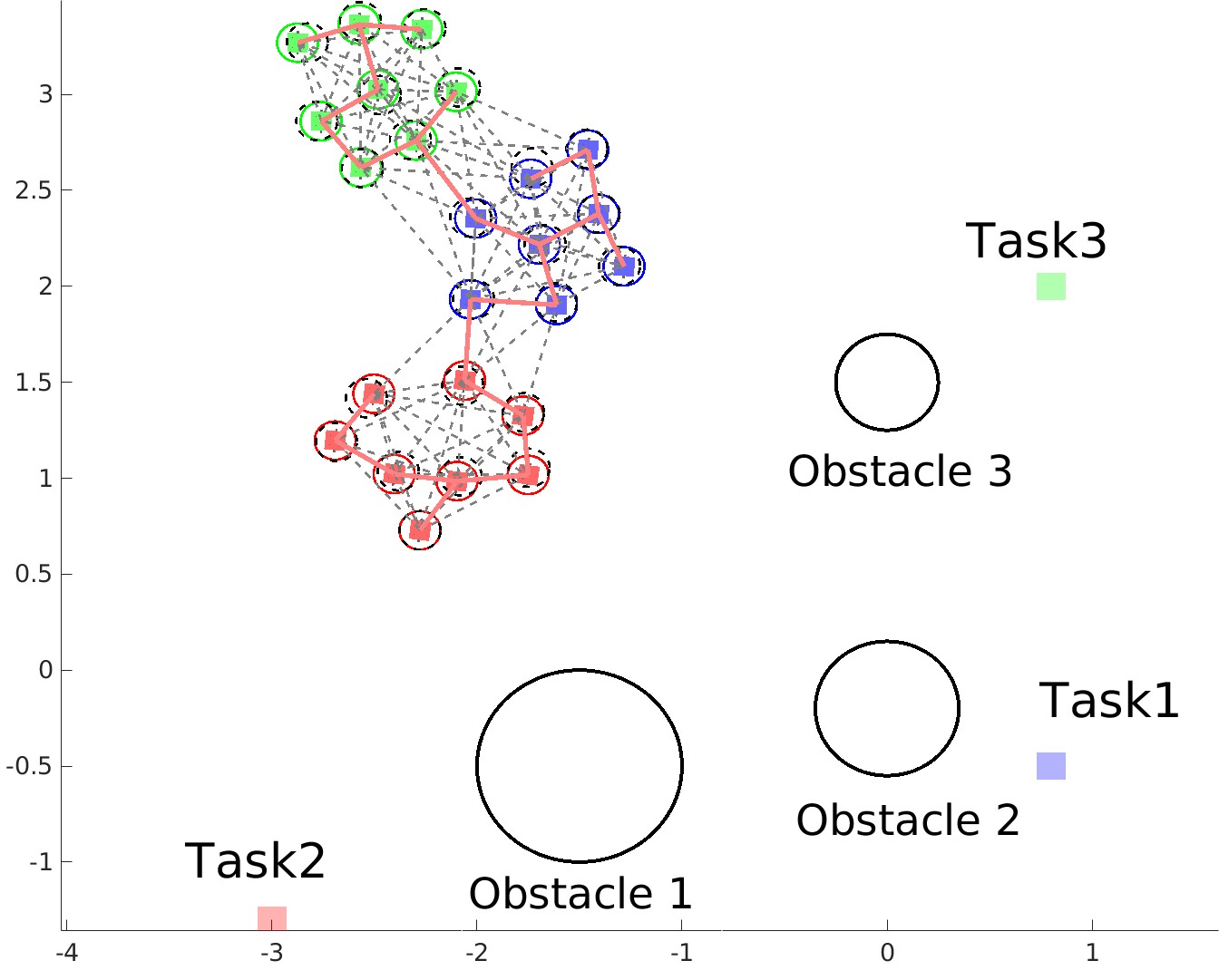}
  \caption{}
  \label{fig:init_4}
  \end{subfigure}
  \caption{Testing environments with different initial positions of robots, numbers of robot subgroups, and obstacles with different lengths, positions, and shapes.}
  \label{fig:init_different}
\end{figure*}

\begin{figure*}[h!]
\centering  
\begin{subfigure}[b]{0.23\textwidth}
  \includegraphics[width=\linewidth]{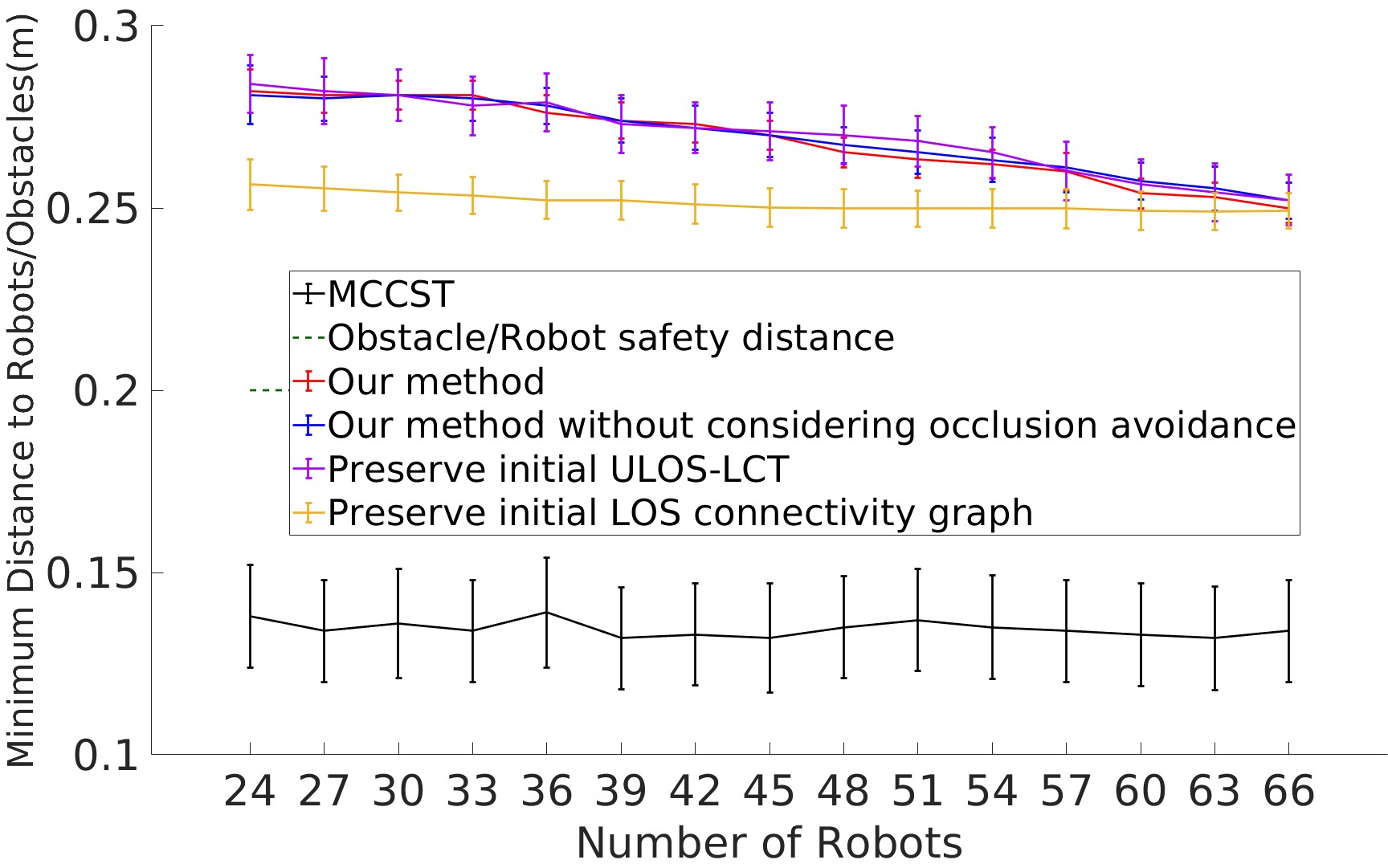}
 \caption{Minimum distance between robot/Obstacle}
  \label{fig4:subfig2}
\end{subfigure}
\begin{subfigure}[b]{0.23\textwidth}
  \includegraphics[width=\linewidth]{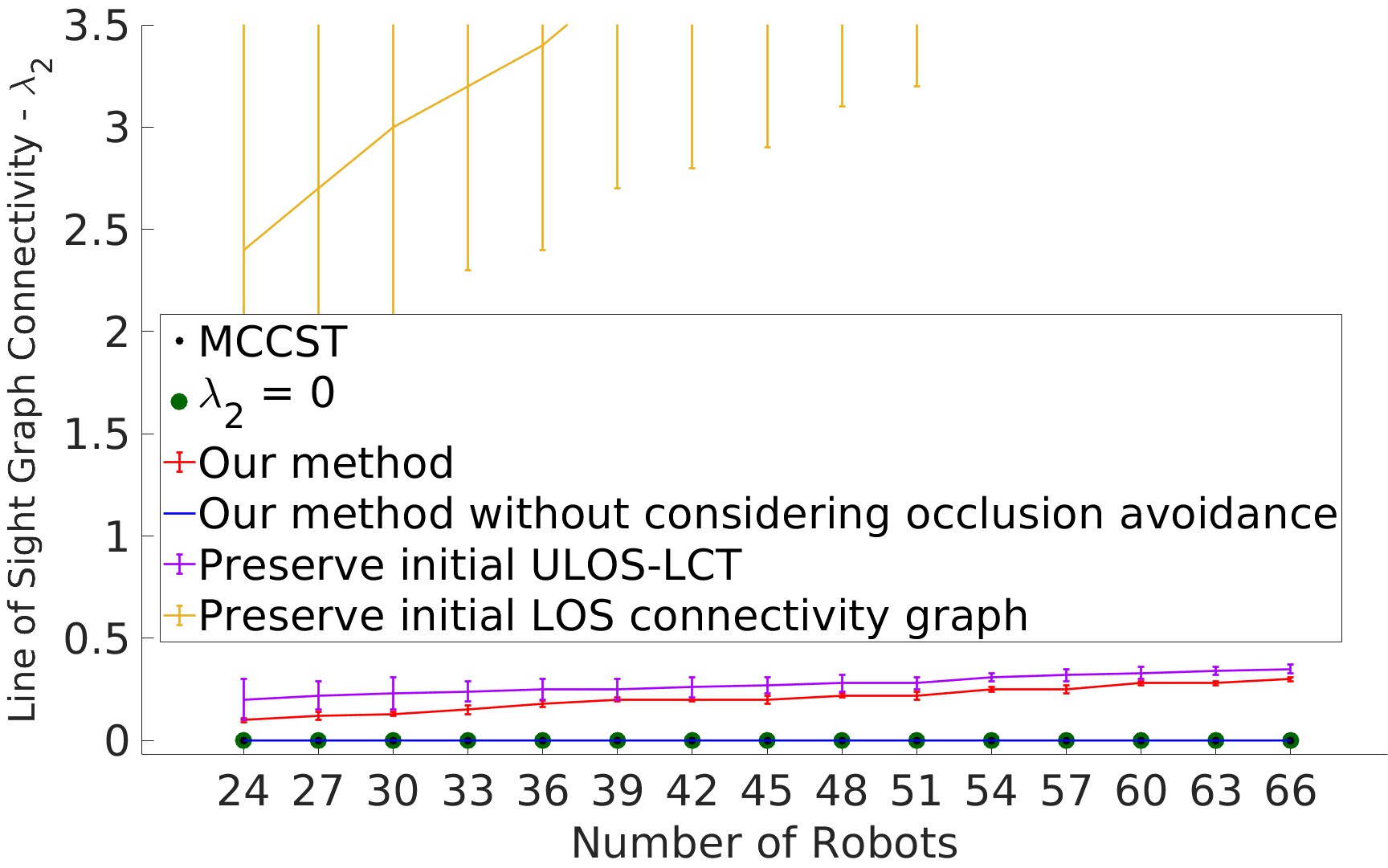}
 \caption{Algebraic LOS connectivity}
  \label{fig4:subfig3}
   \end{subfigure}
\begin{subfigure}[b]{0.23\textwidth}
  \includegraphics[width=\linewidth]{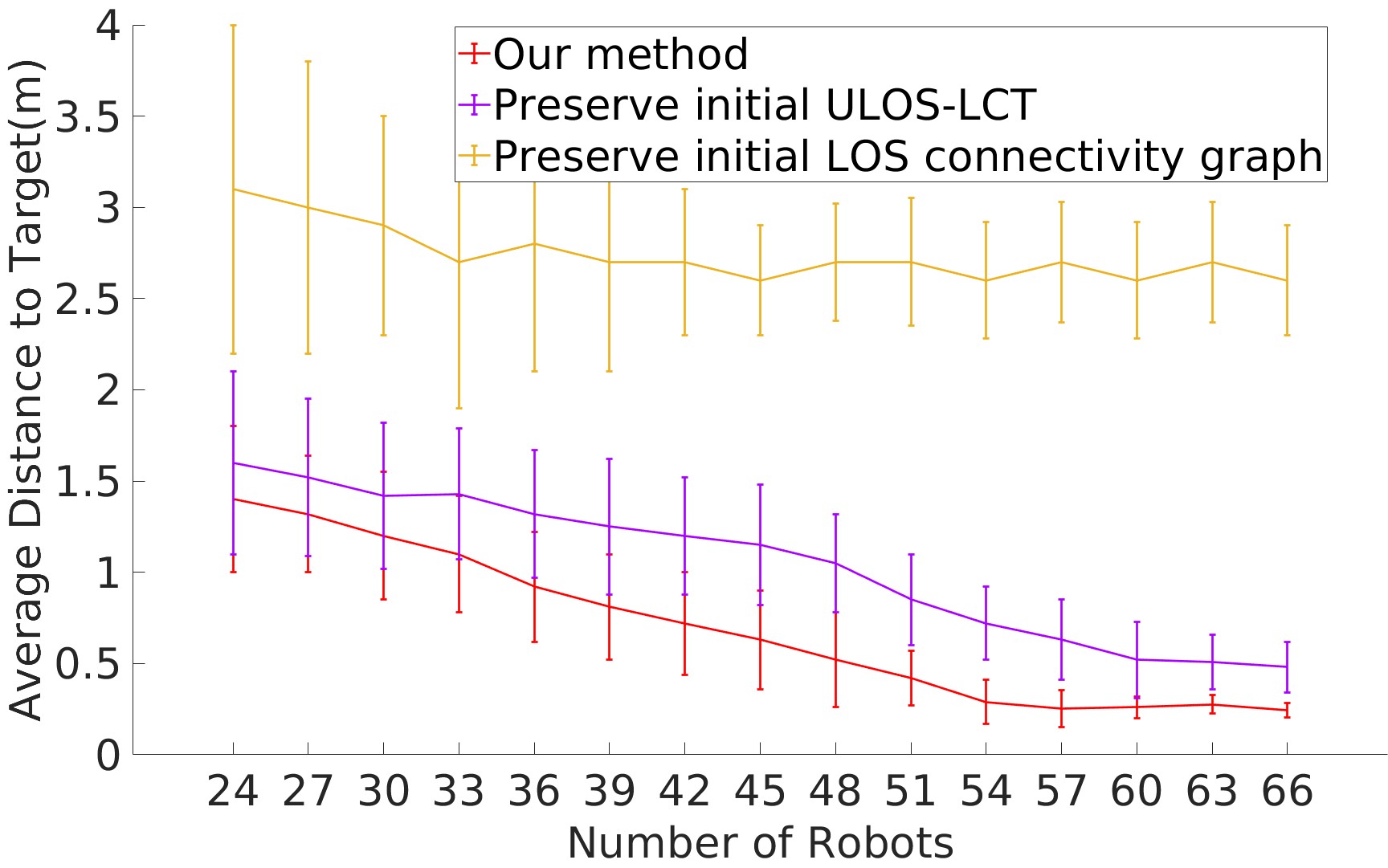}
 \caption{Average distance to target}
  \label{fig4:subfig4}
\end{subfigure}
\begin{subfigure}[b]{0.23\textwidth}
  \includegraphics[width=\linewidth]{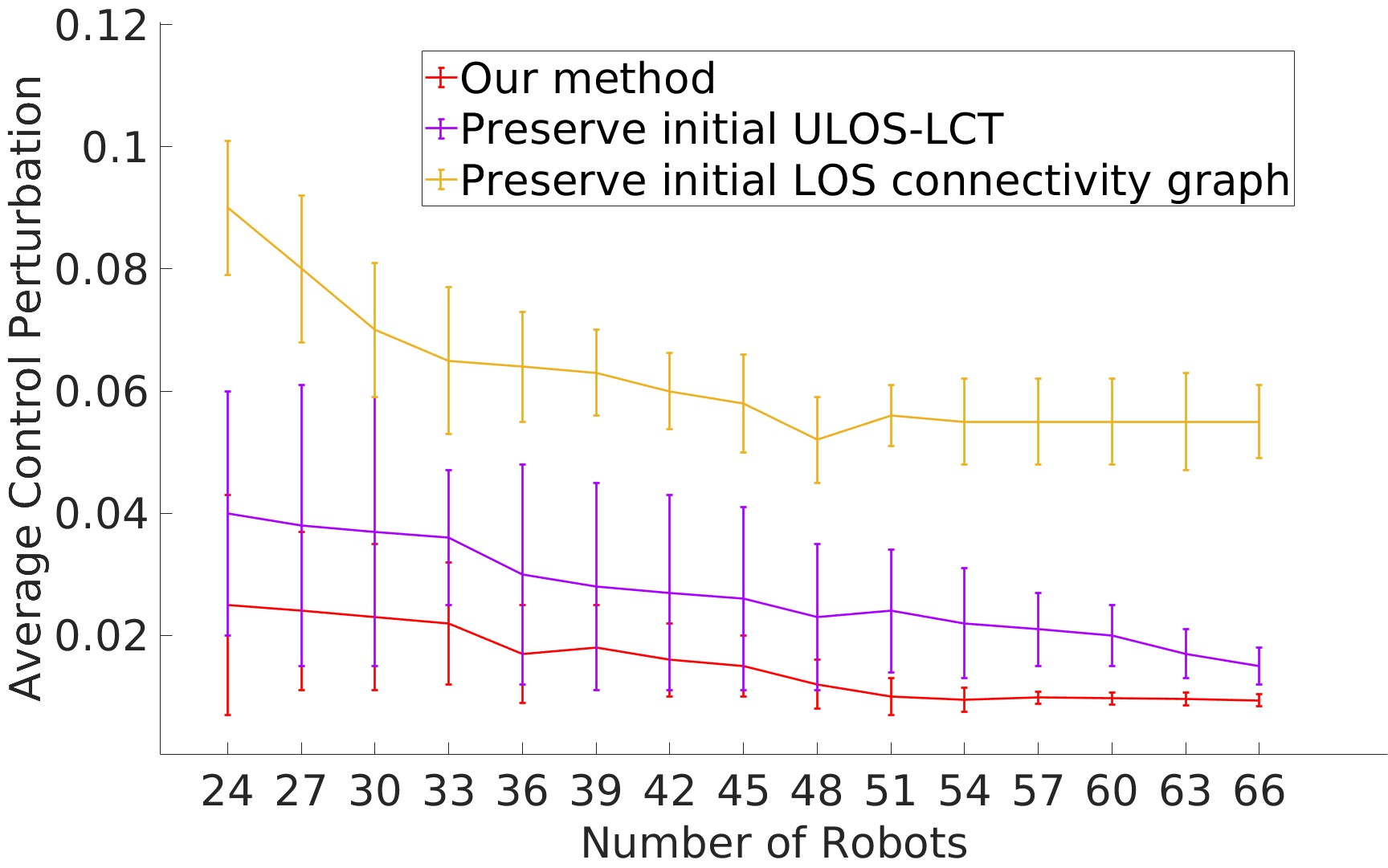}
 \caption{Average control perturbation}
  \label{fig4:subfig5}
\end{subfigure}
\caption{Quantitative results on different sizes of robot team. Error bars show the
standard deviation.}
\label{fig:quantitative_results}
\end{figure*}

\begin{figure*}[h!]
\centering
\begin{subfigure}[b]{0.24\textwidth}
  \includegraphics[width=\linewidth]{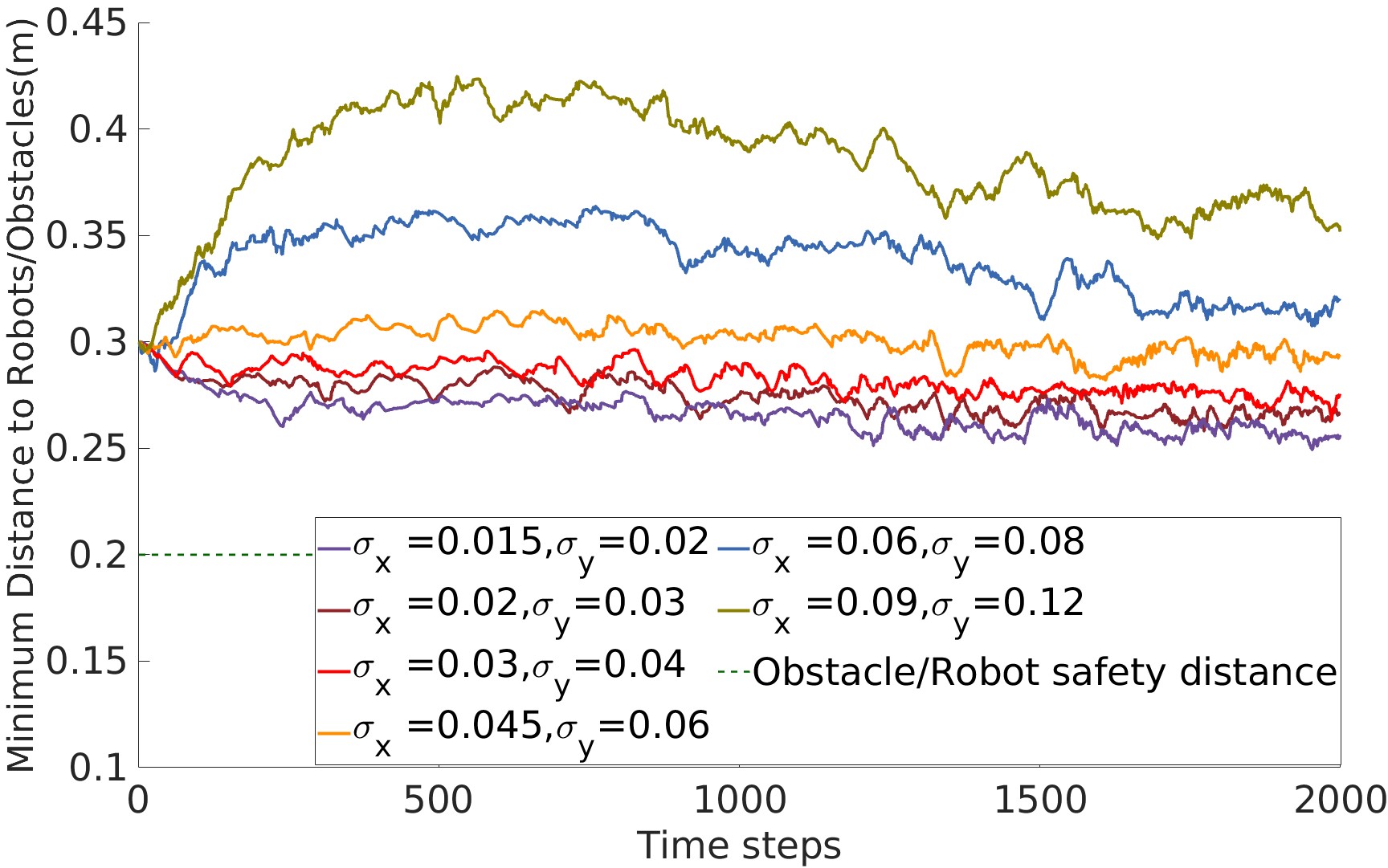}
 \caption{Minimum distance between robot/Obstacle}
  \label{fign:subfigure1}
  \end{subfigure}
\begin{subfigure}[b]{0.24\textwidth}
  \includegraphics[width=\linewidth]{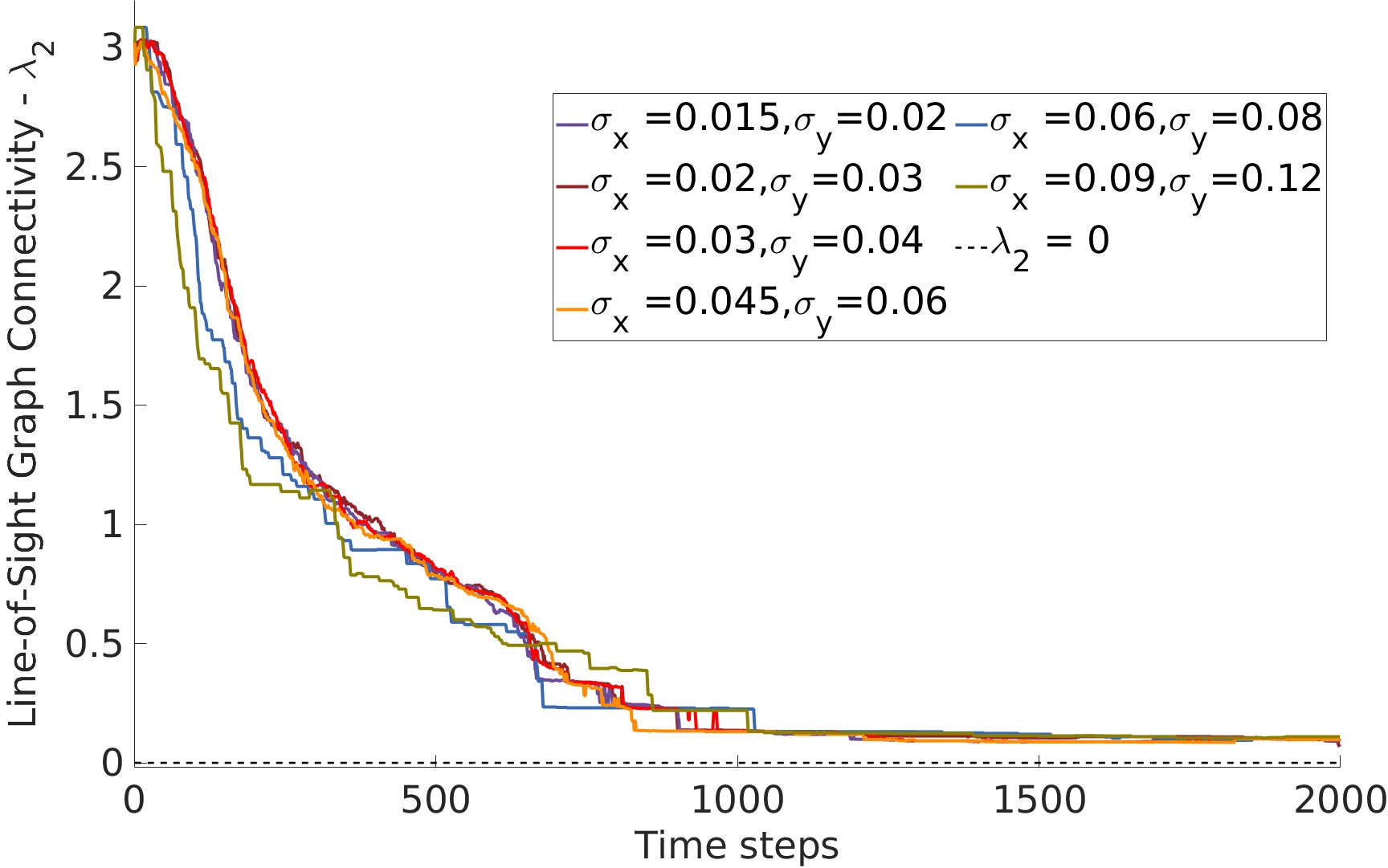}
 \caption{Algebraic LOS connectivity}
  \label{fign:subfigure2}
\end{subfigure}
\begin{subfigure}[b]{0.24\textwidth}
  \includegraphics[width=\linewidth]{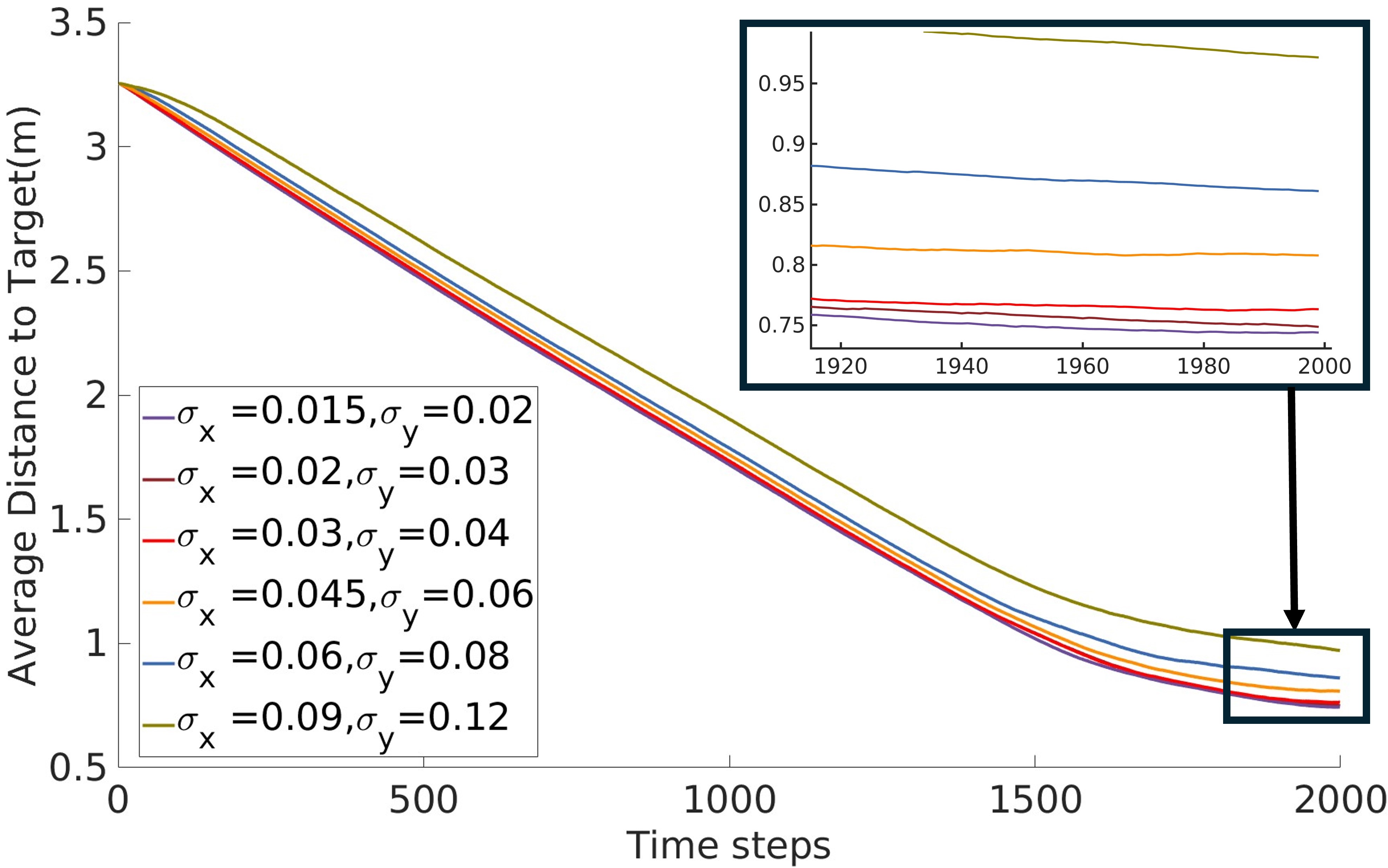}
 \caption{Average distance to target}
  \label{fign:subfigure3}
   \end{subfigure}
\begin{subfigure}[b]{0.24\textwidth}
  \includegraphics[width=\linewidth]{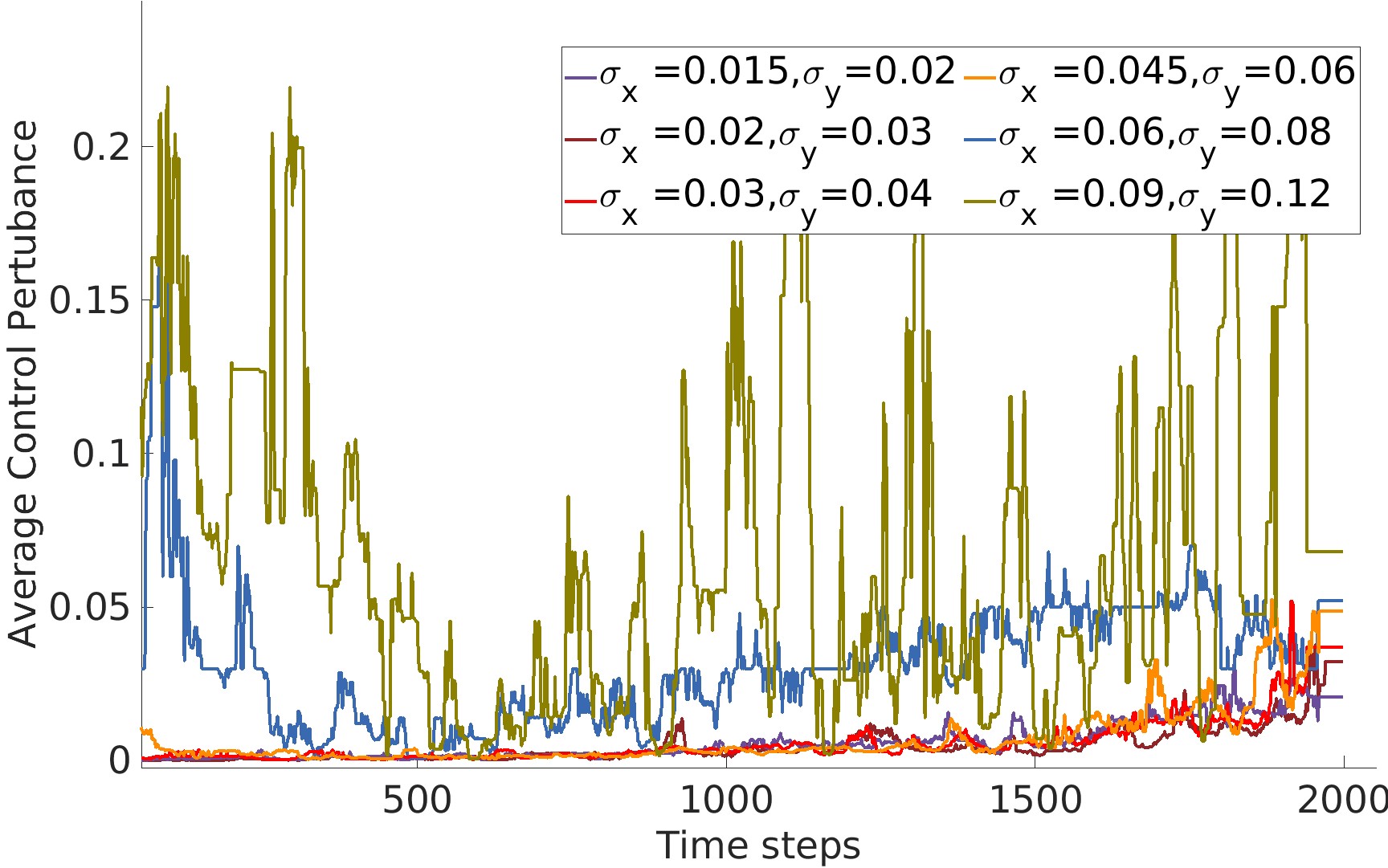}
 \caption{Average control perturbation}
  \label{fign:subfigure4}
\end{subfigure}
  \caption{Numerical results under different levels of noisy observations.}
  \label{fign:performance_comparsion}
\end{figure*}
\begin{figure*}[h!]
\centering  
\begin{subfigure}[b]{0.24\textwidth}
  \includegraphics[width=0.9\linewidth]{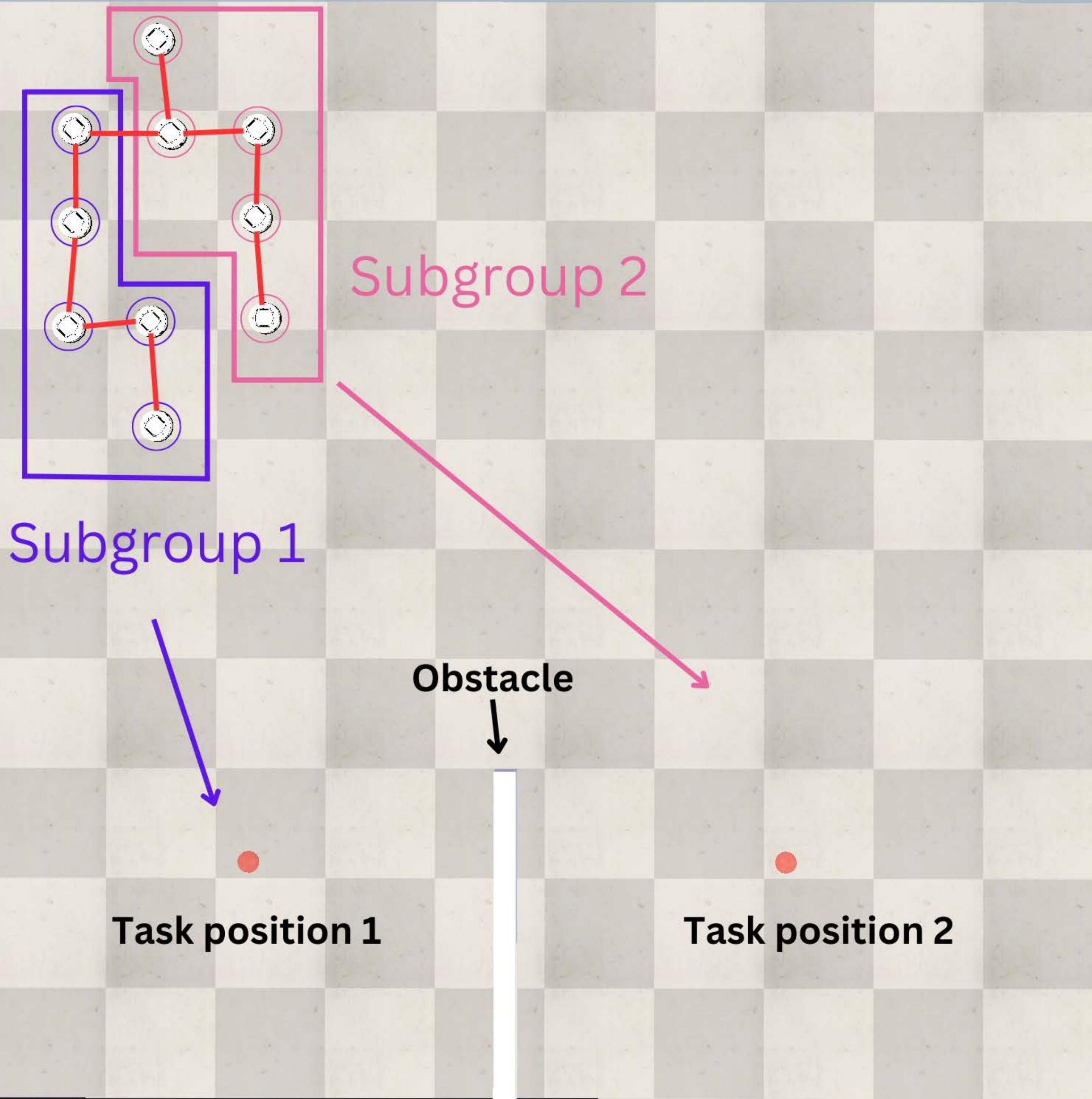}
 \caption{Time Step = 0}
  \label{fig:cop1}
\end{subfigure}
\begin{subfigure}[b]{0.24\textwidth}
  \includegraphics[width=0.9\linewidth]{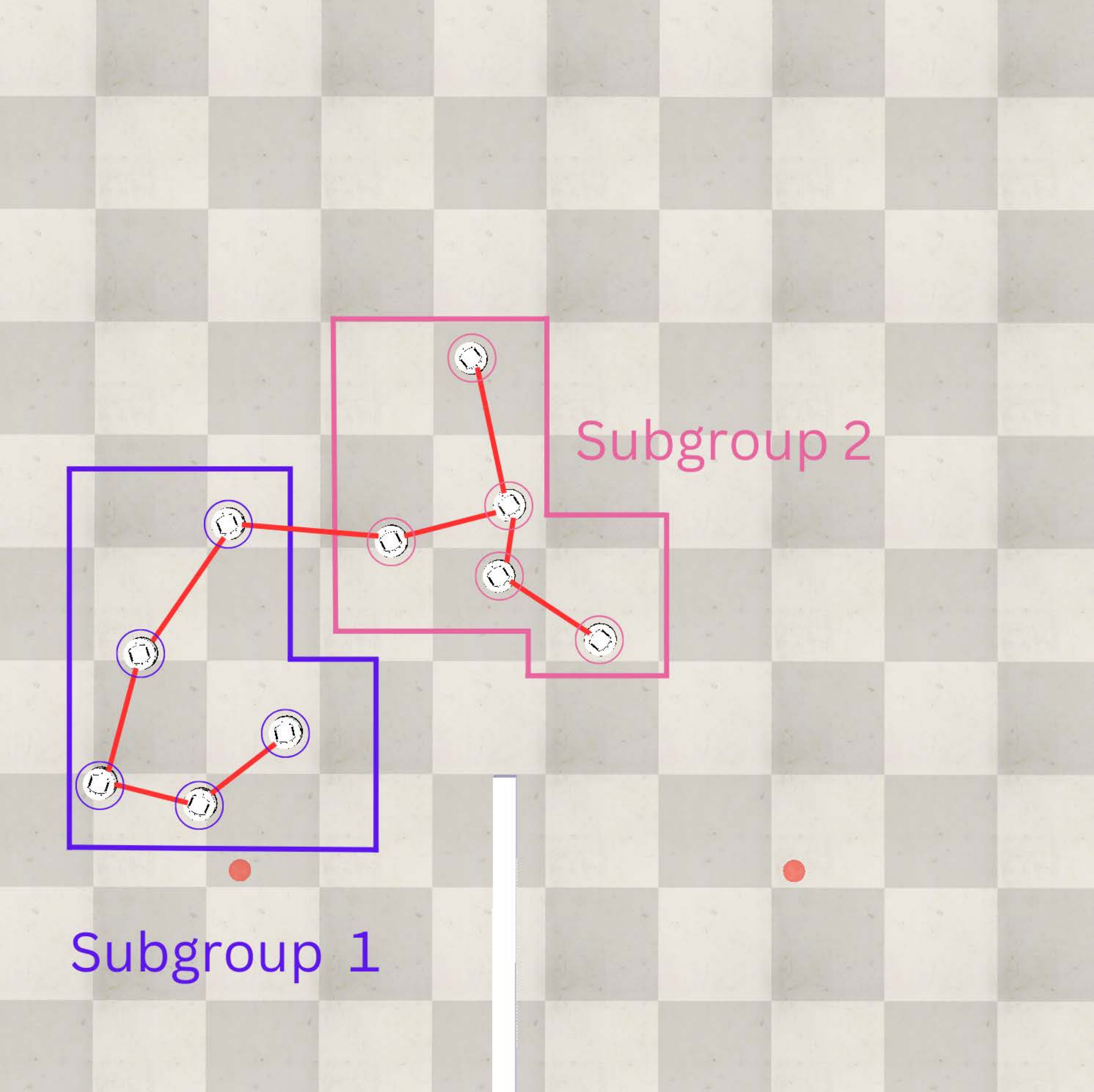}
 \caption{Time Step = 1000}
  \label{fig:cop2}
   \end{subfigure}
\begin{subfigure}[b]{0.24\textwidth}
  \includegraphics[width=0.9\linewidth]{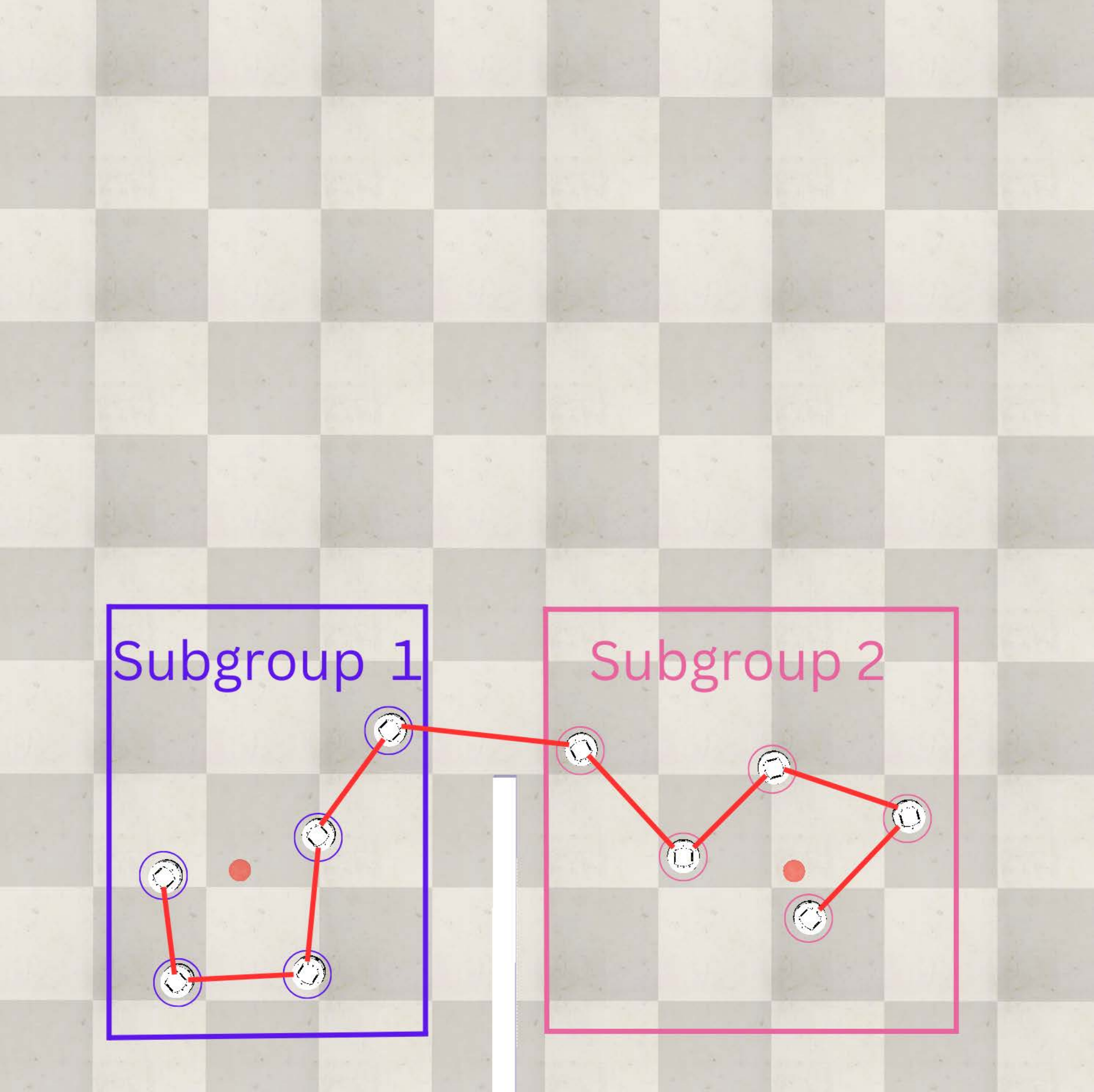}
 \caption{Time Step = 2000 (Converged)}
  \label{fig4:scop3}
\end{subfigure}
\caption{Simulation examples using CoppeliaSim. The confidence level in this experiment is set as $\sigma^\mathrm{s}=\sigma^\mathrm{obs} = \sigma^\mathrm{c} = 0.90,\; \sigma^\mathrm{los} = 0.99$. The Multivariate Gaussian covariance matrix for measurement noise is  $\mathrm{diag}[0.03,0.04]$.  The robot's diameter is 0.14 m. The $R_\mathrm{s},\; R_\mathrm{obs}$ and $R_\mathrm{c}$ are 0.28 m, 0.34 m and 0.9 m, respectively. Red edges denote the current active LOS communication graph $\mathcal{G}^\mathrm{slos*}$ to preserve.}
\label{fig:copsim}
\end{figure*}
\begin{figure*}[h!]
\centering
\begin{subfigure}[b]{0.24\textwidth}
  \includegraphics[width=\linewidth]{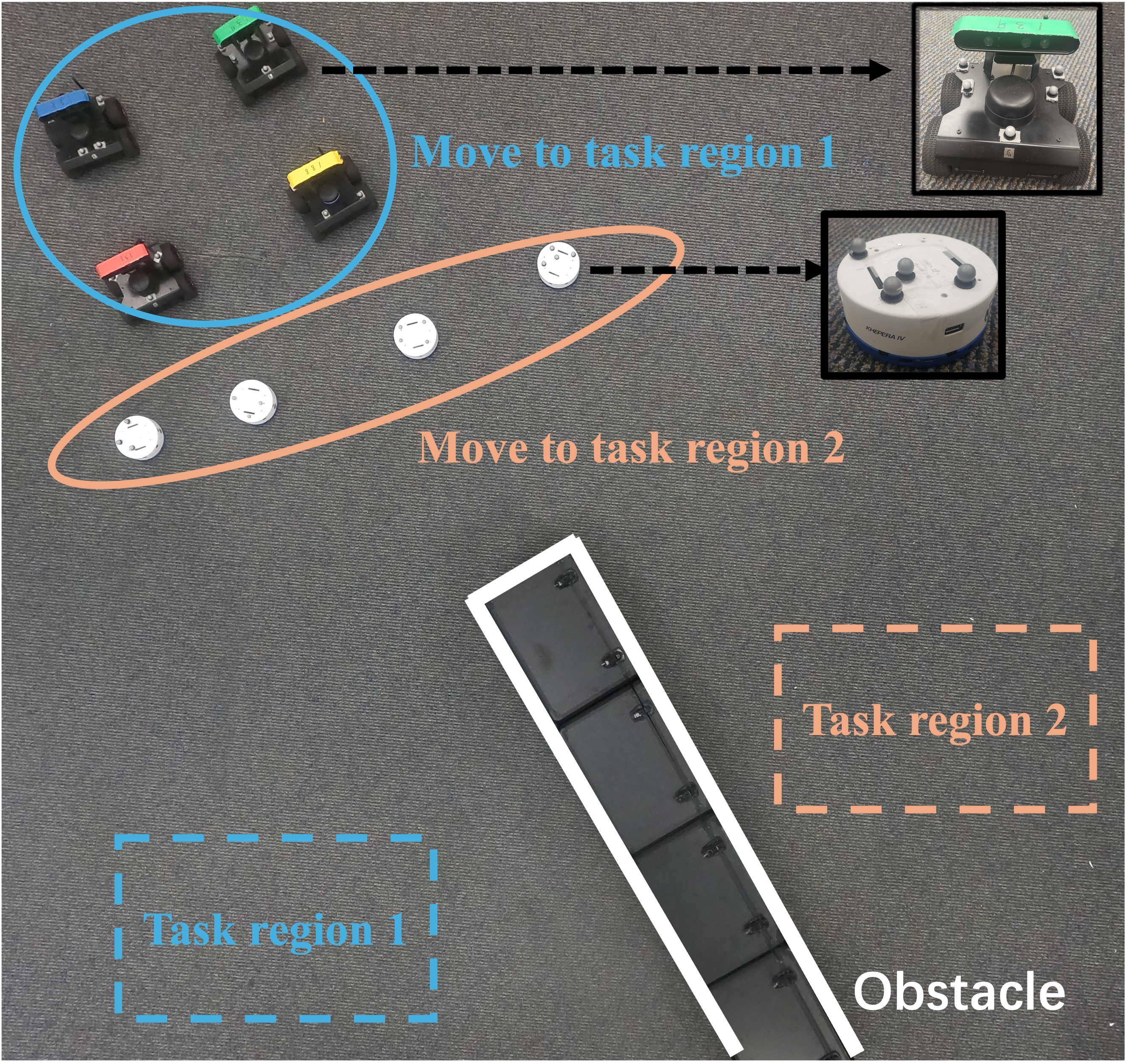}
 \caption{Initial configuration}
  \label{figr:subfigure1}
  \end{subfigure}
  \begin{subfigure}[b]{0.24\textwidth}
  \includegraphics[width=\linewidth]{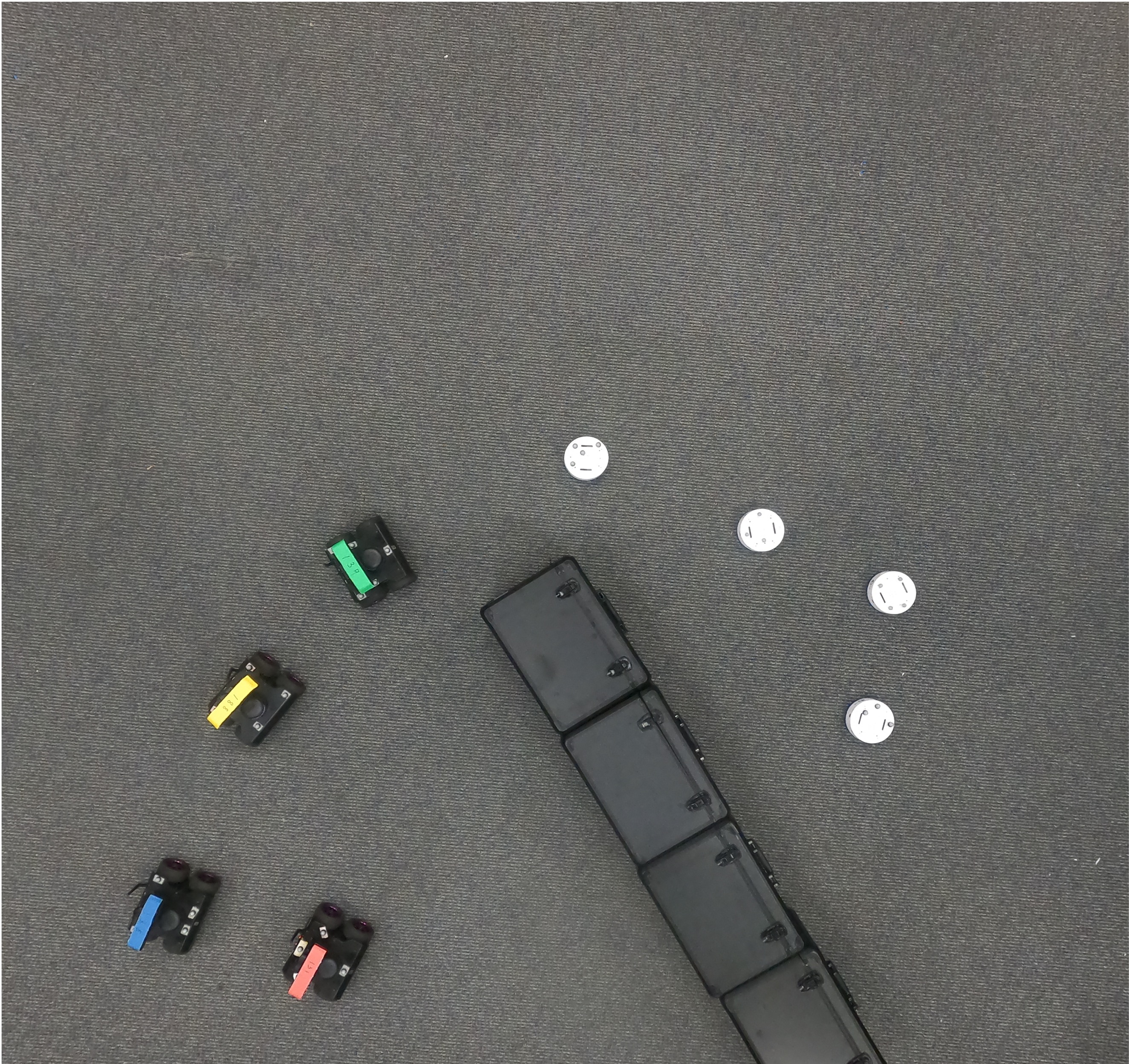}
 \caption{Converged configuration}
  \label{figr:subfigure2}
  \end{subfigure}
  \begin{subfigure}[b]{0.24\textwidth}
  \includegraphics[width=\linewidth]{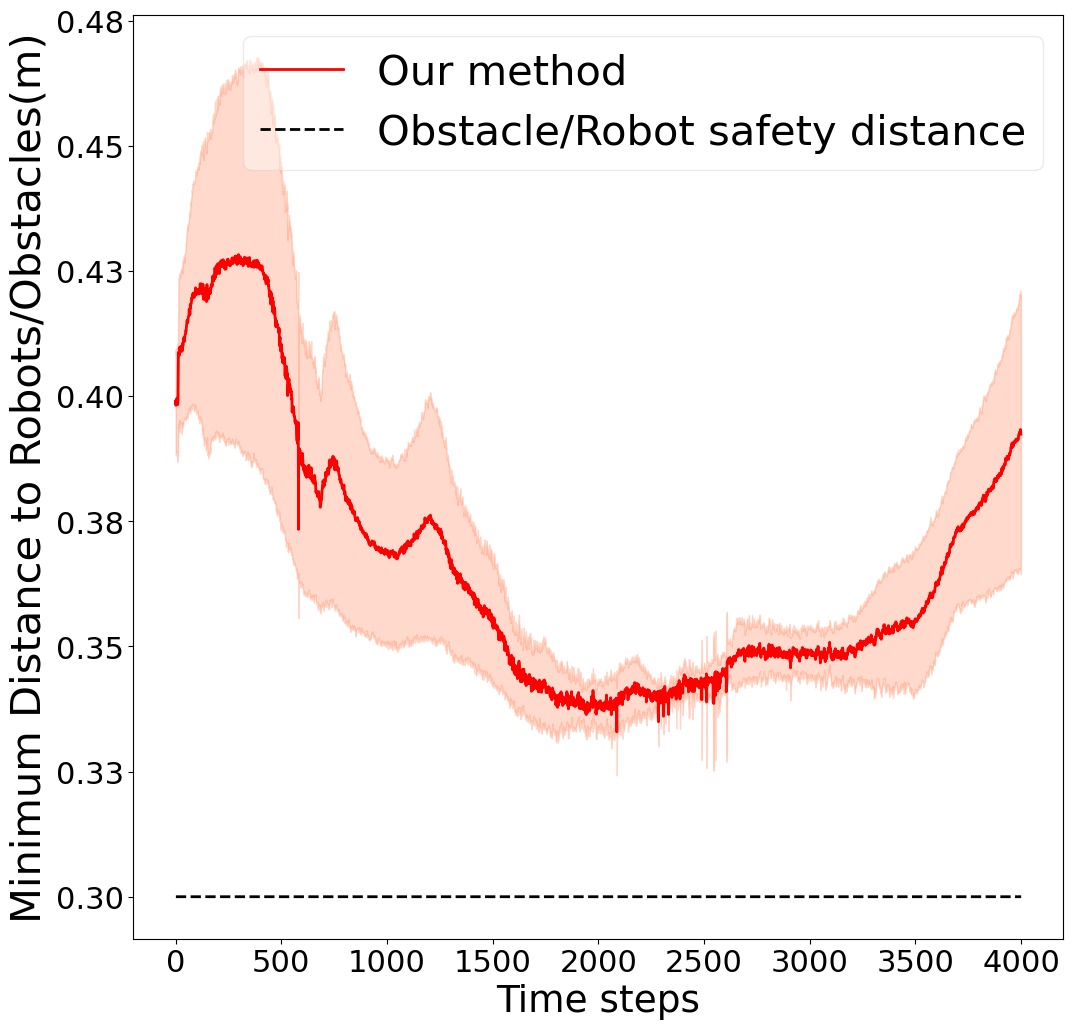}
 \caption{Minimum distance between robot/Obstacle}
  \label{figr:safety}
  \end{subfigure}
   \begin{subfigure}[b]{0.24\textwidth}
  \includegraphics[width=\linewidth]{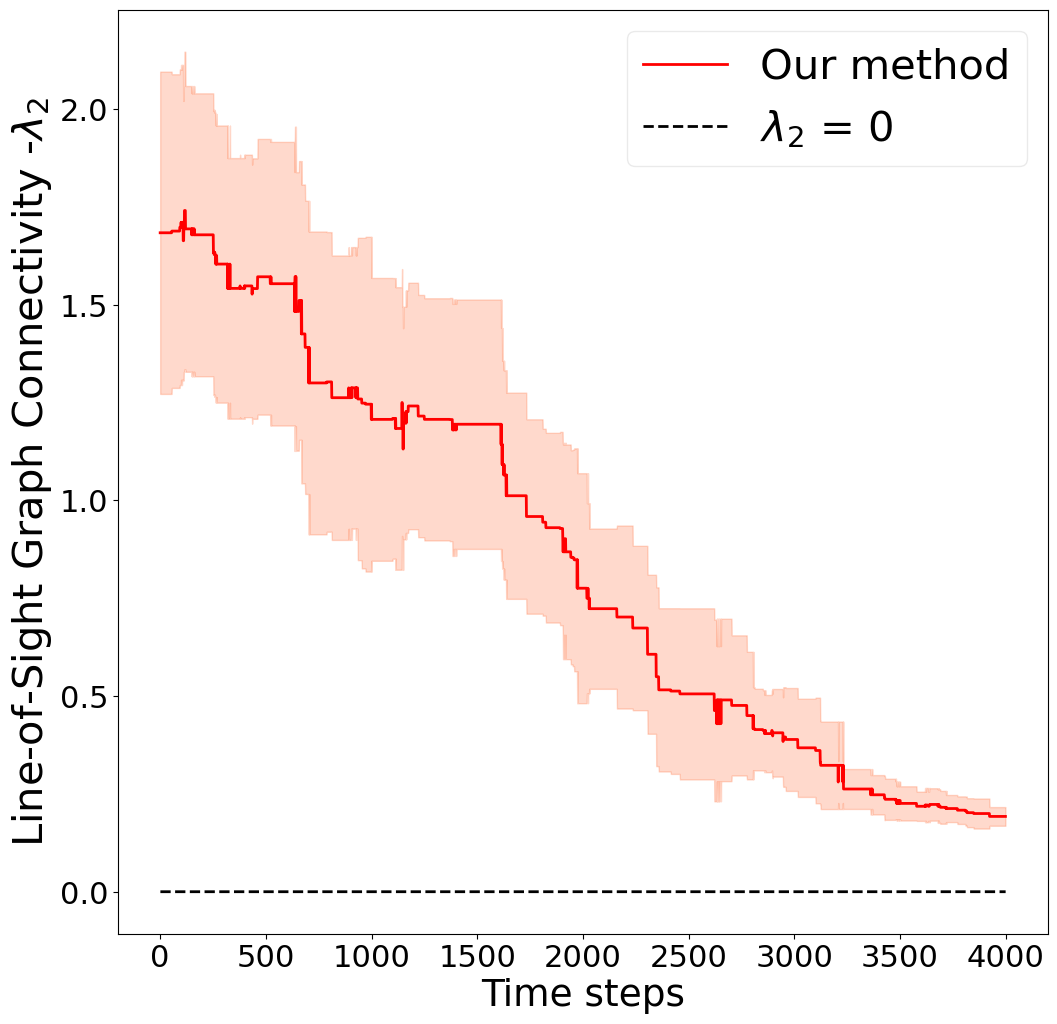}
 \caption{Algebraic LOS connectivity}
  \label{figr:connectivity}
  \end{subfigure}
  \caption{Hardware experiment. The confidence level in this experiment is set as $\sigma^\mathrm{s}=\sigma^\mathrm{obs} = \sigma^\mathrm{c} = 0.90,\; \sigma^\mathrm{los} = 0.90$. The Multivariate Gaussian covariance matrix for measurement noise is  $\mathrm{diag}[0.01,0.02]$. $R_\mathrm{s},\; R_\mathrm{obs}$ and $R_\mathrm{c}$ are 0.3 m, 0.3 m and 0.8 m, respectively. Red lines show the average performance. The shade area shows the standard derivation.}
    \label{fig:real_world_all}
\end{figure*}

\begin{figure}[htbp!]
    \centering   \includegraphics[width=0.6\linewidth]{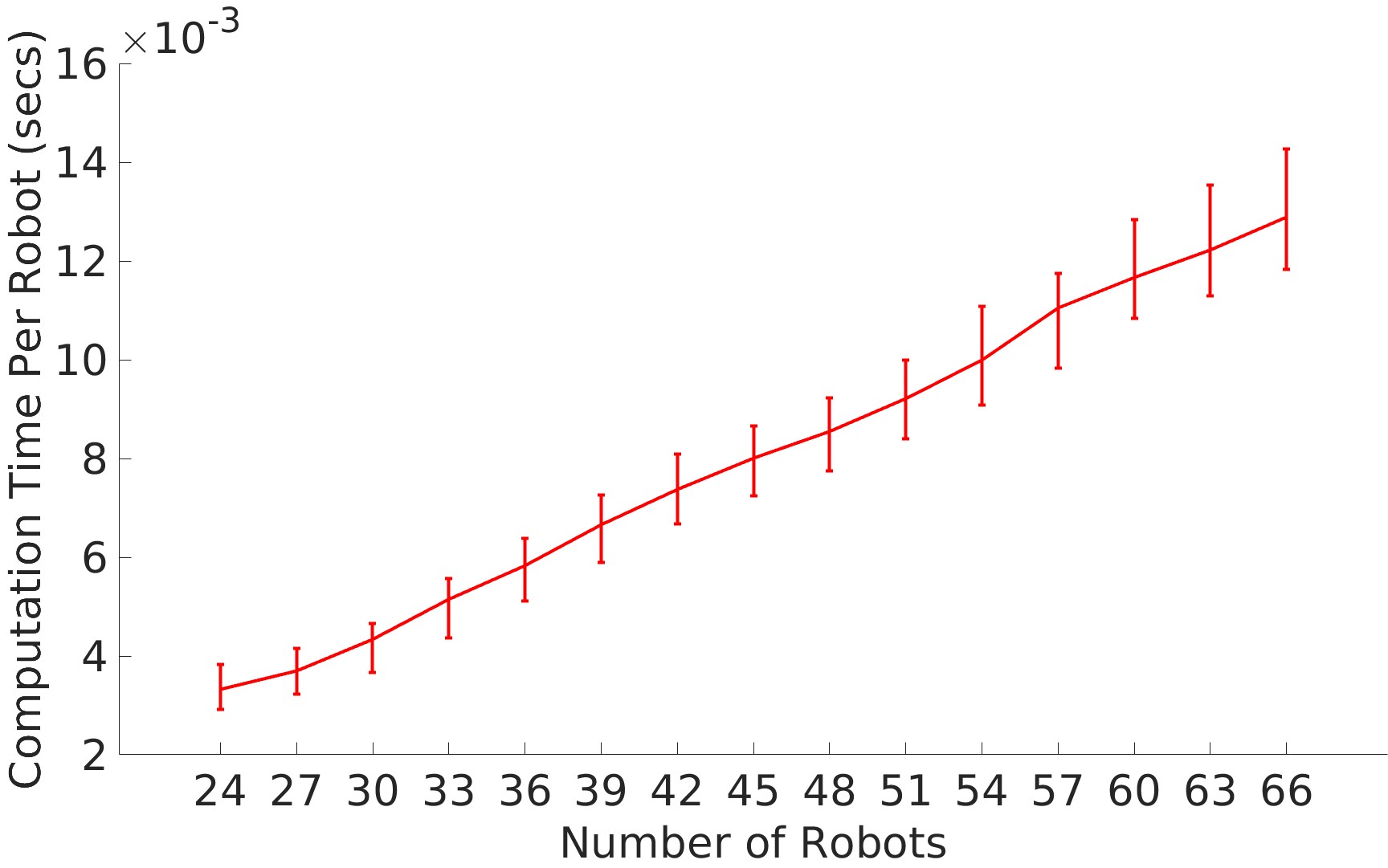}
    \caption{Computation time. Error bar shows the min and max values.}
    \label{fig:time}
\end{figure}
\subsection{Simulation Example}
Fig.~\ref{fig:our_simulation} shows the first set of simulations performed on a team of $N=24$ mobile robots with unicycle dynamics, where controllers are applied using kinematics mapping from \cite{wang2017safety}. Fig.~\ref{fig1:subfiga} to~\ref{fig1:subfigc} demonstrate that our proposed method consistently maintains Line-of-Sight (LOS) connectivity over time. 
Besides, we implemented four baseline methods as shown in Fig.~\ref{fig2:subfigure1}-\ref{fig2:subfigure4}. In Fig.~\ref{fig2:subfigure1} (MCCST \cite{luo2020behavior}), robots fail to maintain a safe distance due to the lack of consideration for observation uncertainty. In Fig.~\ref{fig2:subfigure2} (our method without occlusion avoidance), robots lose inter-robot LOS communication due to obstacles. In Fig.~\ref{fig2:subfigure3} and \ref{fig2:subfigure4} where the enforced LOS communication graph remains unchanged from the beginning, LOS communication is preserved but overly constrains robots' motion with poor task performance as a result. 
In summary, our proposed method effectively maintains safety and LOS connectivity despite observation uncertainty, while achieving the best task performance.

In Fig.~\ref{fig3:subfigure1}, all methods, except MCCST which doesn't account for observation uncertainty, meet the required safety criteria. Fig.~\ref{fig3:subfigure2} shows that both MCCST and our method without occlusion avoidance can lead to a disconnected Line-of-Sight configuration ($\lambda_2 = 0$). In Fig.~\ref{fig3:subfigure3} and Fig.~\ref{fig3:subfigure4}, task performance and control perturbation for MCCST are not included, as it fails to ensure both safety and LOS connectivity under positional uncertainty. These results illustrate that our method achieves the highest task efficiency through minimal control perturbation while maintaining safety and LOS connectivity under observation uncertainty. 
Finally, Fig.~\ref{fig3:subfigure4} demonstrates that our method performs similarly to the centralized solution in terms of task performance.

\subsection{Quantitative Results}\label{sec:quantitative}
In this section, we provide three case studies to verify the effectiveness of our algorithm in different environments, evaluate our algorithm under different noisy observation settings, and analyze the scalability of our algorithm.

\noindent\textbf{\textit{Case study 1:}} To assess the robustness of our algorithm, we conducted experiments under \textit{not only} different environments which are provided in Fig.~\ref{fig1:subfiga} and Fig.~\ref{fig:init_different}, \textit{but} with different initial positions up to 66 robots. 
Specifically, for each batch of robots with different sizes, we conduct experiments in these five different environments, testing five different initial positions within each environment, and repeating each position five times, totaling 125 experiments per number of robots. Finally, $5\times 5\times5\times 15 =1875$ number of experiments are conducted to verify the robustness of our algorithm (15 is the number of batches from 24 robots to 66 robots). The results are summarized in Fig.~\ref{fig:quantitative_results}. Fig.~\ref{fig4:subfig2} demonstrates that safety is guaranteed except for MCCST. Fig.~\ref{fig4:subfig3} shows that our method, fixed initial ULOS-LCT, and fixed LOS communication graph algorithm guarantee the required LOS connectivity, while the MCCST and our method without considering occlusion avoidance fail ($\lambda_2 = 0 $) as it does not consider the LOS connectivity. Fig.~\ref{fig4:subfig4} demonstrates our algorithm's superior task performance in reducing distance to the target. Fig.~\ref{fig4:subfig5} indicates that our method achieves less average control perturbance than fixed-graph algorithms, highlighting the importance of dynamic LOS graph updates.\\

\noindent\textit{\textbf{Case study 2:}} To evaluate the performance of our algorithm under varying levels of observation noise, represented by the matrix $\begin{bmatrix}
\sigma_x & 0 \\
0 & \sigma_y
\end{bmatrix}$, we conducted tests using the same configuration depicted in Fig.~\ref{fig1:subfiga}, but with different observation noise parameters. The results are illustrated in Fig.~\ref{fign:performance_comparsion}. In Fig.~\ref{fign:subfigure1}, the results clearly show that our algorithm reliably guarantees the desired safety, even as observation noise grows. It is worth noting that as the observation noise increases, the algorithm enforces the robots to stay further away from the other robots and obstacles, effectively compensating for the increase in localization error.
Fig.~\ref{fign:subfigure2} verifies that our algorithm preserves LOS connectivity under varying noisy observation parameters. Fig.~\ref{fign:subfigure4} reveals that handling larger observation noise necessitates greater control deviations, which can complicate the execution of original tasks, as indicated by the increased average distances to targets in Fig.~\ref{fign:subfigure3}.\\

\noindent\textbf{\textit{Case study 3:}} To evaluate the efficiency and scalability of our algorithm, we conduct another case study. Our algorithm's running time is dominated by the topology of the graph. The more complex topology costs more iteration to ensure the robots to agree with the same optimal graph $\bar{\mathcal{T}}_w^\mathrm{los'}$. Hence, to evaluate the efficiency and scalability of our method\footnote{The experiments were conducted on a 2.9 GHz Intel Core i7 processor with 16 GB RAM.}, we conducted 20 trials for each number of robots, each trial using a different graph topology. The experiment environment is the same as described in Fig.~\ref{fig1:subfiga}. The results are shown in Fig.~\ref{fig:time}. The worst computation time is evaluated based on \textit{complete graph}\footnote{In this paper, a complete graph is a graph in which each pair of graph vertices is LOS connected with an edge.} $\mathcal{G}^\mathrm{los}$, while the best computation time is evaluated based on the "\textit{minimally connected graph}"\footnote{In this paper, a minimally connected graph is a graph that is LOS connected and there is no edge that can be removed while still leaving the graph LOS connected.}  $\mathcal{G}^\mathrm{los}$.  The results demonstrate that our algorithm can support up to 48 robots updating control commands at 100 Hz. This indicates that our algorithm is capable of real-time implementation with a large number of robots.

\subsection{CoppeliaSim Simulation}
To further validate our algorithm in realistic scenarios, we conducted simulations using CoppeliaSim platform \cite{6696520} with 10 Khepera IV robots under differential drive dynamics, where controllers from our Dec-LOS-LCT method are applied using kinematics mapping from \cite{farias2017khepera}. As depicted in Fig.~\ref{fig:copsim}, robots with observation noise were divided into two subgroups, forming a circle with a radius of 0.5 m 
around two task locations. Our algorithm ensures flexible behaviors while remaining collision-free with the required LOS connectivity for the robots.

\subsection{Real-world Experiment}\label{sec:real_world}
We also validated our algorithm through hardware experiments using 4 Husarion ROSbot 2 PRO robots and 4 Khepera IV robots. The robots with observation noise are grouped based on different types of robots and tasked to two different task regions. Positions of the robots and obstacles are acquired from OptiTrack system with manually added Multivariate Gaussian distributed noise. The initial configuration of the robots is shown in Fig.\ref{figr:subfigure1}. We conducted 10 experiments using our method with this setup. The numerical results, presented in Fig.\ref{figr:safety} and Fig.~\ref{figr:connectivity}, confirm that our algorithm consistently ensures the required safety and maintains Line-of-Sight (LOS) connectivity. The red lines in the figures represent the average results from the 10 experiments, while the shaded areas depict the standard deviation, with ground truth position data provided by the OptiTrack system. Finally, the average running time per robot is 1.25 ms 
at each time step, which also verifies the efficiency of our proposed method. We provide video demonstrations for all simulation and real-world experiments at \textcolor{red}{\url{https://youtu.be/LxTxjxLWjx4?si=_wEq1AHUSgyT_IzI}}.

\section{Conclusion}

In this paper, we introduced a novel decentralized algorithm to address global and subgroup Line-of-Sight (LOS) connectivity maintenance with collision avoidance for robotic teams under positional uncertainty. Probabilistic Line-of-Sight Connectivity Barrier Certificates (PrLOS-CBC) are proposed to ensure a lower bounded probability of inter-robot LOS occlusion avoidance with closed-from expression. 
By formulating the problem as a bi-level optimization process, it enables robots to first compute the least constraining set of LOS edges as the lower-level task, and the resulting composition of PrLOS-CBC is integrated into the upper-level task to derive the minimally modified task-related controllers satisfying the global and subgroup LOS constraints.
With our proposed Dec-LOS-LCT algorithm, the optimization formulation can be solved in a fully decentralized manner, with superior performance demonstrated in simulation and real-world experiments. 
Future work includes extending the method and algorithm to more general nonlinear systems with higher relative degrees, by leveraging variants of CBFs such as Exponential Control Barrier Function (ECBF) \cite{nguyen2016exponential} and High Order Control Barrier Functions (HOCBFs) \cite{xiao2021high}.

\bibliographystyle{unsrt}  
\bibliography{ref}
\newpage
\clearpage
\newpage
\section{appendix}
\subsection{Discussion on Eq.~(\ref{Prcbc})}\label{app:eq:prcbc}
Here, we provide the brief discussion on the proof of existence of $\mathcal{C}^{\sigma^\mathrm{c}}_\mathbf{u}(\hat{\mathbf{x}},\mathcal{G}^\mathrm{slos})$  and how to construct  $B^{\sigma^\mathrm{c}}_{i,j}$ and $f^{\sigma^\mathrm{c}}_{i,j}$ in $\mathcal{C}^{\sigma^\mathrm{c}}_\mathbf{u}(\hat{\mathbf{x}},\mathcal{G}^\mathrm{slos})$ in Eq.~(\ref{Prcbc}). The desired set of $\mathbf{x}$ for pairwise robots $i$ and $j$ satisfying the communication distance condition is defined in Eq.~(\ref{eq:h_conn_per}): 
\begin{align}
&h^\mathrm{c}_{i,j}(\mathbf{x}) = R_\mathrm{c}^2 - ||\mathbf{x}_i - \mathbf{x}_j||^2, \;\forall (v_i,v_j)\in\mathcal{E}^\mathrm{c}\subseteq\mathcal{E}^\mathrm{los},\;\notag\\
&\mathcal{H}^\mathrm{c}_{i,j} = \{\mathbf{x} \in \mathbb{R}^{dN}| h^\mathrm{c}_{i,j}(\mathbf{x}) \geq 0 \} \tag{4}
\end{align}
Following \cite{luo2020multi}, to ensure the probability of pairwise robot satisfying the communication distance condition above the user-defined confidence level $\sigma^\mathrm{c}$, we can derive the following sufficient condition:
\begin{equation}\label{eq:connectivity_sufficient}
    \Pr (\mathbf{u}_i,\mathbf{u}_j \in \mathcal{B}^\mathrm{c}_{i,j}(\mathbf{x})) \geq \sigma^\mathrm{c}\Rightarrow  \Pr (\mathbf{x}_i,\mathbf{x}_j \in \mathcal{H}^\mathrm{c}_{i,j}(\mathbf{x})) \geq \sigma^\mathrm{c}
\end{equation}
where $ \mathcal{B}^\mathrm{c}_{i,j}(\mathbf{x}) = \{ \mathbf{u}\in \mathbb{R}^{qN} | \dot{h}^\mathrm{c}_{i,j}(\mathbf{x},\mathbf{u}) + \gamma h^\mathrm{c}_{i,j}(\mathbf{x})\geq 0 \}$. Considering $\dot{h}^\mathrm{c}_{i,j}(\mathbf{x},\mathbf{u}) = \frac{\partial{h}^{c}_{i,j}}{\partial{\mathbf{x}}}(\mathbf{x})(\Delta F_{i,j}(\mathbf{x}) + G_{i,j}(\mathbf{x})\mathbf{u}_{i,j})$ with $\Delta F_{i,j}(\mathbf{x}) = F_i(\mathbf{x}_j)-F_j(\mathbf{x})$ and  $ G_{i,j}(\mathbf{x})\mathbf{u}_{i,j} = G_{i}(\mathbf{x}_i)\mathbf{u}_{i}- G_{j}(\mathbf{x}_j)\mathbf{u}_{j}$, we can re-write the sufficiency condition $\Pr (\mathbf{u}_i,\mathbf{u}_j \in \mathcal{B}^\mathrm{c}_{i,j}(\mathbf{x})) \geq \sigma^\mathrm{c}$ in Eq.~\eqref{eq:connectivity_sufficient} using Eq.~\eqref{eq:h_conn_per} as follows:
{\footnotesize\begin{align}\label{eq:app:connectivity}
        & \Pr (\mathbf{u}_i,\mathbf{u}_j \in \mathcal{B}^\mathrm{c}_{i,j}(\mathbf{x})) \geq \sigma^\mathrm{c}\quad \Longleftrightarrow \\
       & \Pr (\frac{\partial{h}^\mathrm{c}_{i,j}}{\partial{\mathbf{x}}}(\mathbf{x})G_{i,j}(\mathbf{x})\mathbf{u}_{i,j} \geq -\gamma h^\mathrm{c}_{i,j}(\mathbf{x}) - \frac{\partial{h}^\mathrm{c}_{i,j}}{\partial{\mathbf{x}}}(\mathbf{x})(\Delta F_{i,j}(\mathbf{x})) \geq \sigma^\mathrm{c}\notag
\end{align}}
Hence, following \cite{luo2020multi}, we can reorganize Eq.~\eqref{eq:app:connectivity} as: 
{\footnotesize\begin{align}\label{eq:proof_conn}
\text{Pr}\bigg(\bigg[\Delta \mathbf{x}_{i,j}+\overbrace{\frac{ G_{i,j}\mathbf{u}_{i,j}+\Delta F_{i,j}}{\gamma}}^{\dot{\mathbf{x}}_i-\dot{\mathbf{x}}_j}\bigg]^2 \leq R_\mathrm{c}^2+\bigg[\overbrace{\frac{G_{i,j}\mathbf{u}_{i,j}+\Delta F_{i,j}}{\gamma}}^{\dot{\mathbf{x}}_i-\dot{\mathbf{x}}_j}\bigg]^2 \bigg)\geq \sigma^\mathrm{c}
\end{align}}
where $\Delta \mathbf{x}_{i,j} = \mathbf{x}_i -\mathbf{x}_j \sim \mathcal{N}(\mathbf{\hat{x}}_i-\mathbf{\hat{x}}_j,\Sigma_i+\Sigma_j)$.

Eq.~\eqref{eq:proof_conn} follows the same format as Eq.~(19) in \cite{luo2020multi} when without motion uncertainty.
With this, the rest of proof for existence of $\mathcal{C}^{\sigma^\mathrm{c}}_\mathbf{u}(\hat{\mathbf{x}},\mathcal{G}^\mathrm{slos})$ could directly follow the proof of Existence of PrSBC in Appendix A.1 in  \cite{luo2020multi}, which guarantees that the $\mathcal{C}^{\sigma^\mathrm{c}}_\mathbf{u}(\hat{\mathbf{x}},\mathcal{G}^\mathrm{slos})$ defined in Eq.~\eqref{Prcbc} is always non-empty.

Then, one can also follow \cite{luo2020multi} to derive the sufficient condition, so that ensuring $\mathbf{u}\in \mathcal{C}^{\sigma^\mathrm{c}}_\mathbf{u}(\hat{\mathbf{x}},\mathcal{G}^\mathrm{slos}) \implies  \text{Eq.}~\eqref{eq:proof_conn}$. 
Specifically, we need to modify the Eq.~(24) in \cite{luo2020multi} by changing the sign $\geq$ in $\text{Pr}(\cdot)$
to the sign $\leq$. Note that in \cite{luo2020multi}, they consider the motion uncertainty, hence they have the extra term $B^{l}_{i,j}$. The rest of the computation follows the calculation of PrSBC. The only difference is that, rather than enforcing the condition $\text{Pr}(\Delta\mathbf{x}^{l}_{i,j}\leq\mathcal{A})\geq\sigma^\mathrm{c}$ or $\text{Pr}(\Delta\mathbf{x}^{l}_{i,j}\geq\mathcal{B})\geq\sigma^\mathrm{c}$, we aim to enforce the condition $\text{Pr}(\mathcal{A}\leq\Delta\mathbf{x}^{l}_{i,j}\leq\mathcal{B})\leq\sigma^\mathrm{c}$, 
which is equivalent to enforcing $\text{Pr}(\Delta\mathbf{x}^{l}_{i,j}\geq\mathcal{A})\geq\frac{1+\sigma^\mathrm{c}}{2}$ and $\text{Pr}(\Delta\mathbf{x}^{l}_{i,j}\leq\mathcal{B})\geq\frac{1+\sigma^\mathrm{c}}{2}$.
By denoting $e^{l}_{i,j} = \Phi(\frac{1+\sigma^\mathrm{c}}{2})$ with $\Phi(\cdot)$ as the inverse cumulative distribution
function (CDF) of random variable $\Delta \mathbf{x}^l_{i,j}$ along each $l$th dimension of $\Delta \mathbf{x}_{i,j}$ , one can develop the following condition to ensure Eq.~\eqref{eq:proof_conn} holds true on each dimension:
{\footnotesize\begin{align}
    \forall l =1,...d:\quad 2e^{l}_{i,j}(G_{i,j}\mathbf{u}_{i,j})_l/\gamma&\leq -[(e_{i,j}^{l})^2 - R^2_\mathrm{c} +2e^{l}_{i,j}\Delta F^{l}_{i,j}/\gamma] 
    \notag\\
                        -2e^{l}_{i,j}(G_{i,j}\mathbf{u}_{i,j})_l/\gamma &\leq (e_{i,j}^{l})^2 + R^2_\mathrm{c} +2e^{l}_{i,j}\Delta F^{l}_{i,j}/\gamma 
\end{align}}
where \((G_{i,j}\mathbf{u}_{i,j})_l = (G_i\mathbf{u}_i - G_j\mathbf{u}_j)_l \in \mathbb{R}\) and \( \Delta F_{i,j}^l = F_i^l - F_j^l \in \mathbb{R}\) denote the $l$th element of $G_{i,j}\mathbf{u}_{i,j}\in\mathbb{R}^d$ and $\Delta F_{i,j}\in\mathbb{R}^d$ respectively.

Finally, one can combine the condition in all dimensions to get $B^{\sigma^\mathrm{c}}_{i,j}\in\mathbb{R}^{2d\times qN}$ and $f^{\sigma^\mathrm{c}}_{i,j}\in\mathbb{R}^{2d}$ for $B^{\sigma^\mathrm{c}}_{i,j}\mathbf{u}\leq f^{\sigma^\mathrm{c}}_{i,j}$ in $\mathcal{C}^{\sigma^\mathrm{c}}_\mathbf{u}(\hat{\mathbf{x}},\mathcal{G}^\mathrm{slos})$ as:

{\footnotesize\begin{align}
    B^{\sigma^\mathrm{c}}_{i,j} &= \underbrace{\begin{bmatrix}
    2e^{1}_{i,j}&\cdots&0 \\
    -2e^{1}_{i,j}&\cdots&0 \\
    \vdots & \ddots & \vdots \\
    0&\cdots&2e^{d}_{i,j} \\
    0&\cdots&-2e^{d}_{i,j} 
    \end{bmatrix}}_{\in\mathbb{R}^{2d\times d}} 
    \overbrace{
    [0,\ldots,G_i,\ldots,-G_j,\ldots,0]}^{\in\mathbb{R}^{d\times qN}} \notag\\
 \notag
    f^{\sigma^\mathrm{c}}_{i,j} &= \underbrace{
    \begin{bmatrix}
    -\gamma(e_{i,j}^{1})^2 + \gamma R^2_\mathrm{c} - 2e^{1}_{i,j}\Delta F^{1}_{i,j}\\
    \gamma(e_{i,j}^{1})^2 + \gamma R^2_\mathrm{c} +2e^{1}_{i,j}\Delta F^{1}_{i,j}\\
    \vdots\\
    -\gamma(e_{i,j}^{d})^2 + \gamma R^2_\mathrm{c} - 2e^{d}_{i,j}\Delta F^{d}_{i,j}\\
    \gamma(e_{i,j}^{d})^2 + \gamma R^2_\mathrm{c} +2e^{d}_{i,j}\Delta F^{d}_{i,j}
    \end{bmatrix}}_{\in\mathbb{R}^{2d}} \notag
\end{align}}

\subsection{Proof of Lemma~\ref{loscbcdefinition}}\label{app:sec:prooflemma2}
\begin{lemma}\label{lemma:valid}
(Summarized from \cite{capelli2020connectivity})
A function $h$ is a valid control barrier function, if the following properties are satisfied:
\begin{enumerate}
  \item The function $h$ is continuously differentiable.
  \item The first-order time derivative of $h$ depends explicitly on the control input $\mathbf{u}$ (i.e., $h$ is of relative degree one).
  \item It is possible to find an extended class-$\mathcal{K}$ function $\kappa(\cdot)$ such that $\sup_{\mathbf{u}\in\mathcal{U}}\{\dot{h}(\mathbf{x},\mathbf{u})+\kappa(h(\mathbf{x}))\}\geq 0$ for all $\mathbf{x}$. 
\end{enumerate}
\end{lemma}
\begin{customlemma}{3}
\textbf{Probabilistic Line-of-Sight Connectivity Barrier Certificates (PrLOS-CBC:)}
Given a LOS communication spanning graph $\mathcal{G}^\mathrm{slos}=(\mathcal{V},\mathcal{E}^\mathrm{slos})$, a desired set $\mathcal{H}^\mathrm{los}(\mathcal{G}^\mathrm{slos})$ in Eq.~(\ref{eq:hlosset}) with $h^\mathrm{los}_{i,j,o}$ from Eq.~(\ref{los1}), and a user-defined high probability ${\sigma^\mathrm{los}}\in(0,1)$, for any Lipschitz continuous controller $\mathbf{u}$, the PrLOS-CBC as admissible control space $ \mathcal{C}^{\sigma^\mathrm{los}}_\mathbf{u}(\hat{\mathbf{x}},\mathcal{C}_\mathrm{obs},\mathcal{G}^\mathrm{slos})$ defined below enforces system state to stay in 
$\{\bigcap_{\{v_i,v_j \in \mathcal{V}:(v_i,v_j)\in\mathcal{E}^\mathrm{slos}\}} \mathcal{H}^\mathrm{los}_{i,j}\}$ with high probability (by enforcing pairwise LOS separately with satisfying $\sigma^\mathrm{los}$): 
{\footnotesize
\begin{align}\tag{15}
&\mathcal{C}^{\sigma^\mathrm{los}}_\mathbf{u} (\hat{\mathbf{x}},\mathcal{C}_\mathrm{obs},\mathcal{G}^\mathrm{slos}) = \{\mathbf{u}\in\mathbb{R}^{qN}: \\
&\dot{h}^\mathrm{los}_{i,j,o}(\hat{\mathbf{x}}, \mathbf{x}^\mathrm{obs},\sigma^\mathrm{los},\mathbf{u})+\gamma h^\mathrm{los}_{i,j,o}(\hat{\mathbf{x}},\mathbf{x}^\mathrm{obs},\sigma^\mathrm{los})\geq 0,\forall (v_i,v_j)\in \mathcal{E}^\mathrm{slos},\forall o\} \notag
\end{align}
}
where $\dot{h}_{i,j,o}^\mathrm{los}(\hat{\mathbf{x}},\mathbf{x}^\mathrm{obs}, \mathbf{u}) = -(\mathbf{x}^\mathrm{obs}_{o}-\frac{\hat{\mathbf{x}}_i+\hat{\mathbf{x}}_j}{2})^{T}Q^{\sigma^\mathrm{los}}_{i,j}(F_{i,j}(\mathbf{x})+G_{i,j}(\mathbf{x})\mathbf{u}_{ij})$, $F_{i,j}(\mathbf{x}) = F_{i}(\mathbf{x}_i)+F_{j}(\mathbf{x}_j)$ and $G_{i,j}(\mathbf{x})\mathbf{u}_{ij} = G_{i}(\mathbf{x}_i)\mathbf{u}_{i}+ G_{j}(\mathbf{x}_j)\mathbf{u}_{j}$. 
\end{customlemma}
\begin{proof}
To prove the control constraints in $\mathcal{C}^{\sigma^\mathrm{los}}_\mathbf{u} (\hat{\mathbf{x}},\mathcal{C}_\mathrm{obs},\mathcal{G}^\mathrm{slos})$ for every pair of robots $i, j$ where $(v_i,v_j)\in\mathcal{E}^\mathrm{slos}$, ensuring they stay in $\mathcal{H}_{i,j}^\mathrm{los}$ with the prescribed probability $\sigma^\mathrm{los}\in (0,1)$, we first demonstrate that the function $h^\mathrm{los}_{i,j,o}$, as defined in Eq.~(\ref{los1}), is a valid control barrier function (CBF).

Considering our proposed candidate CBF $h_{i,j,o}^\mathrm{los}$ in Eq.~(\ref{los1}) for $\forall(v_i,v_j)\in\mathcal{E}^\mathrm{slos}$, $\forall o$, we can derive: (a) $\frac{\partial{h_{i,j,o}^\mathrm{los}(\hat{\mathbf{x}},\mathbf{x}^\mathrm{obs},\sigma^\mathrm{los})}}{\partial \mathbf{x}} = 2(\mathbf{x}^\mathrm{obs}_{o}-\frac{\hat{\mathbf{x}}_i+\hat{\mathbf{x}}_j}{2})^{T}Q^{\sigma^\mathrm{los}}_{i,j}$
, and (b) $\dot{h}_{i,j,o}^\mathrm{los}(\hat{\mathbf{x}},\mathbf{x}^\mathrm{obs}, \mathbf{u}) = -(\mathbf{x}^\mathrm{obs}_{o}-\frac{\hat{\mathbf{x}}_i+\hat{\mathbf{x}}_j}{2})^{T}Q^{\sigma^\mathrm{los}}_{i,j}(F_{i,j}(\mathbf{x})+G_{i,j}(\mathbf{x})\mathbf{u}_{ij})$, where $Q^{\sigma^\mathrm{los}}_{i,j}\in\mathbb{R}^{d\times d}$ is the positive-definite symmetric computed from the ellipsoid approximation for the occlusion-free condition between pairwise robots $i$ and $j$. Thus, we can conclude that $h_{i,j,o}^\mathrm{los}$ (a) is continuously differentiable, and (b) is of relative degree one, which satisfies the first two conditions in Lemma~\ref{lemma:valid}. 
Then, by substituting $\dot{h}_{i,j,o}^\mathrm{los}(\hat{\mathbf{x}},\mathbf{x}^\mathrm{obs},\sigma^\mathrm{los}, \mathbf{u}) = \frac{d}{dt}h_{i,j,o}^\mathrm{los}(\hat{\mathbf{x}},\mathbf{x}^\mathrm{obs},\sigma^\mathrm{los})$ in $\dot{h}^\mathrm{los}_{i,j,o}(\hat{\mathbf{x}}, \mathbf{x}^\mathrm{obs},\sigma^\mathrm{los},\mathbf{u})+\gamma h^\mathrm{los}_{i,j,o}(\hat{\mathbf{x}},\mathbf{x}^\mathrm{obs},\sigma^\mathrm{los})\geq 0$ from Eq.~(\ref{eq:prlos_cbc}), we can obtain: 
\begin{align}\begin{aligned}
\label{eq:provefeasibility_all}
&(\mathbf{x}^\mathrm{obs}_{o}-\frac{\hat{\mathbf{x}}_i+\hat{\mathbf{x}}_j}{2})^{T}Q^{\sigma^\mathrm{los}}_{i,j}(F_{i,j}(\mathbf{x})+G_{i,j}(\mathbf{x})\mathbf{u}_{ij})\\
&\leq \gamma[(\mathbf{x}^\mathrm{obs}_{o}-\frac{\hat{\mathbf{x}}_i+\hat{\mathbf{x}}_j}{2})^{T}Q^{\sigma^\mathrm{los}}_{i,j}(\mathbf{x}^\mathrm{obs}_{o}-\frac{\hat{\mathbf{x}}_i+\hat{\mathbf{x}}_j}{2})-1] 
\end{aligned}\end{align} 

Given the assumption that each pair-wise robot satisfies the occlusion-free condition initially, we can derive that $\gamma[(\mathbf{x}^\mathrm{obs}_{o}-\frac{\hat{\mathbf{x}}_i+\hat{\mathbf{x}}_j}{2})^{T}Q_{i,j}(\mathbf{x}^\mathrm{obs}_{o}-\frac{\hat{\mathbf{x}}_i+\hat{\mathbf{x}}_j}{2})-1] \geq 0$ on the right-hand side of Eq.~(\ref{eq:provefeasibility_all}) holds, i.e., there are no obstacles blocking the LOS between pair-wise robot $i$ and $j$. 
Besides, it is also straight to prove that $(\mathbf{x}^\mathrm{obs}_{o}-\frac{\hat{\mathbf{x}}_i+\hat{\mathbf{x}}_j}{2})Q^{\sigma^\mathrm{los}}_{i,j} \neq 0$ (matrix $Q^{\sigma^\mathrm{los}}_{i,j} \succ 0$ and vector $\mathbf{x}^\mathrm{obs}_{o}-\frac{\hat{\mathbf{x}}_i+\hat{\mathbf{x}}_j}{2} \neq 0$). 
Furthermore, it is possible to find a pair-wise control input $\mathbf{u}_i=\mathbf{u}^0_i$, $\mathbf{u}_j=\mathbf{u}^0_j$ such that $\dot{\mathbf{x}}_i=0$, $\dot{\mathbf{x}}_j=0$. In this case, $\dot{\mathbf{x}}_i+\dot{\mathbf{x}}_j = F_{i,j}(\mathbf{x})+G_{i,j}(\mathbf{x})\mathbf{u}^{0}_{ij} = 0$, then Eq.~(\ref{eq:provefeasibility_all}) holds.
Hence, we can conclude that, the proposed control barrier function $h^\mathrm{los}_{i,j,o}$ is a valid control barrier function. It is then straightforward to extend to all pairwise inter-robot in a LOS communication spanning graph $\mathcal{G}^\mathrm{slos}$ (i.e., $\forall (v_i,v_j)\in\mathcal{E}^\mathrm{slos}$) occlusion-free constraints in Eq.~(\ref{eq:prlos_cbc}). To that end, we can prove that the admissible control space constrained by Eq.~(\ref{eq:prlos_cbc}) is always non-empty.

Next, we will demonstrate that the control space defined by $(\mathbf{x}^\mathrm{obs}_{o}-\frac{\hat{\mathbf{x}}_i+\hat{\mathbf{x}}_j}{2})^{T}Q^{\sigma^\mathrm{los}}_{i,j}(F_{i,j}(\mathbf{x})+G_{i,j}(\mathbf{x})\mathbf{u}_{ij})\leq \gamma[(\mathbf{x}^\mathrm{obs}_{o}-\frac{\hat{\mathbf{x}}_i+\hat{\mathbf{x}}_j}{2})^{T}Q^{\sigma^\mathrm{los}}_{i,j}(\mathbf{x}^\mathrm{obs}_{o}-\frac{\hat{\mathbf{x}}_i+\hat{\mathbf{x}}_j}{2})-1],\forall o$ enforces the pairwise inter-robot stay occlusion-free with satisfying probability $\sigma^\mathrm{los}$ over time. 
Recall that each robot state is considered as a Gaussian distributed variable with ${\mathbf{x}}_i\sim \mathcal{N}({\hat{\mathbf{x}}}_i,{\sum}_i),\forall i\in\mathcal{I}$.
Hence, given the observed positions and uncertainty covariance of pairwise robots ${\mathbf{x}}_i\sim \mathcal{N}({\hat{\mathbf{x}}}_i,{\sum}_i)$, $ {\mathbf{x}}_j\sim \mathcal{N}({\hat{\mathbf{x}}}_j,{\sum}_j)$ that are initially LOS connected and the corresponding $\sqrt{\sigma^\mathrm{los}}$-confidence error ellipsoids $\mathcal{Q}^{\sqrt{\sigma^\mathrm{los}}}_i,\mathcal{Q}^{\sqrt{\sigma^\mathrm{los}}}_j$, we adopt MVCE $\mathcal{Q}^{\sigma^\mathrm{los}}_{i,j}(Q^{\sigma^\mathrm{los}}_{i,j}, \hat{\mathbf{p}}^0_{i,j})=\{\mathbf{p}\in\mathbb{R}^d:(\mathbf{p}-\hat{\mathbf{p}}^0_{i,j})^TQ^{\sigma^\mathrm{los}}_{i,j}(\mathbf{p}-\hat{\mathbf{p}}^0_{i,j})\leq 1\}$ over the set of $\sqrt{\sigma^\mathrm{los}}$-confidence error ellipsoids $\{\mathcal{Q}^{\sqrt{\sigma^\mathrm{los}}}_i,\mathcal{Q}^{\sqrt{\sigma^\mathrm{los}}}_j\}$ to approximate the occlusion-free condition for pairwise robots. Given that each robot has $\sqrt{\sigma^\mathrm{los}}$ probability to located within each confidence error ellipsoids and the two robot states are independent, then it is straight to prove that the probability of two robots are both located in the confidence error ellipsoids $\{\mathcal{Q}^{\sqrt{\sigma^\mathrm{los}}}_i,\mathcal{Q}^{\sqrt{\sigma^\mathrm{los}}}_j\}$ is $\sigma^\mathrm{los} =\sqrt{\sigma^\mathrm{los}}\times\sqrt{\sigma^\mathrm{los}} $. Hence, the MVCE approximation $\mathcal{Q}^{\sigma^\mathrm{los}}_{i,j}$ can guarantee that the two robots state are both within $\mathcal{Q}^{\sigma^\mathrm{los}}_{i,j}$ with $\sigma^\mathrm{los}$ probability. With this, we adopt this approximation to define the occlusion-free set $\mathcal{H}_{i,j}^\mathrm{los}$ in Eq.~(\ref{los1}) which is the sufficient condition to ensure pairwise robots stay occlusion-free with probability $\sigma^\mathrm{los}$, i.e. $\mathbf{x}_i,\mathbf{x}_j\in\mathcal{H}_{i,j}^\mathrm{los} \implies\text{Pr}(\mathbf{x}_i,\mathbf{x}_j \in \{\mathbf{x}_i, \mathbf{x}_j: \mathbf{x}_i(1-\Omega)+\mathbf{x}_j\Omega \notin \mathcal{C}_{\text{obs}},\;\forall \Omega\in [0,1]\})\geq\sigma^\mathrm{los}$. Since we assume that the pairwise robots are initially occlusion-free, considering the property of the Lemma~\ref{lem:cbf} and its probabilistic extension in \cite{luo2020multi}, we can derive $\mathbf{u}_i,\mathbf{u}_j \in \{ \mathbf{u}_i,\mathbf{u}_j: (\mathbf{x}^\mathrm{obs}_{o}-\frac{\hat{\mathbf{x}}_i+\hat{\mathbf{x}}_j}{2})^{T}Q^{\sigma^\mathrm{los}}_{i,j}(F_{i,j}(\mathbf{x})+G_{i,j}(\mathbf{x})\mathbf{u}_{ij})\leq \gamma[(\mathbf{x}^\mathrm{obs}_{o}-\frac{\hat{\mathbf{x}}_i+\hat{\mathbf{x}}_j}{2})^{T}Q^{\sigma^\mathrm{los}}_{i,j}(\mathbf{x}^\mathrm{obs}_{o}-\frac{\hat{\mathbf{x}}_i+\hat{\mathbf{x}}_j}{2})-1],\forall o\} \implies \mathbf{x}_i,\mathbf{x}_j\in\mathcal{H}_{i,j}^\mathrm{los} \implies\text{Pr}(\mathbf{x}_i,\mathbf{x}_j \in  \{\mathbf{x}_i,\mathbf{x}_j: \mathbf{x}_i(1-\Omega)+\mathbf{x}_j\Omega \notin \mathcal{C}_{\text{obs}},\;\forall \Omega\in [0,1]\})\geq\sigma^\mathrm{los}$. 

With this, we can guarantee that $(\mathbf{x}^\mathrm{obs}_{o}-\frac{\hat{\mathbf{x}}_i+\hat{\mathbf{x}}_j}{2})^{T}Q^{\sigma^\mathrm{los}}_{i,j}(F_{i,j}(\mathbf{x})+G_{i,j}(\mathbf{x})\mathbf{u}_{ij})\leq \gamma[(\mathbf{x}^\mathrm{obs}_{o}-\frac{\hat{\mathbf{x}}_i+\hat{\mathbf{x}}_j}{2})^{T}Q^{\sigma^\mathrm{los}}_{i,j}(\mathbf{x}^\mathrm{obs}_{o}-\frac{\hat{\mathbf{x}}_i+\hat{\mathbf{x}}_j}{2})-1],\forall o$ defines the admissible control space for pairwise robots $i,j$ that enforces the two robots to stay occlusion-free with satisfying probability $\sigma^\mathrm{los}$ over time.

Finally, consider the desired occlusion-free set for the robot team as $\{\bigcap_{\{v_i,v_j \in \mathcal{V}:(v_i,v_j)\in\mathcal{E}^\mathrm{slos}\}} \mathcal{H}^\mathrm{los}_{i,j}\}$, 
we seek to enforce all pairwise robots (i.e., $v_i,v_j \in \mathcal{V}:(v_i,v_j)\in\mathcal{E}^\mathrm{slos}$) to stay in this desired set so that the occlusion-free condition is satisfied with required probability. Since we proved that (a) the control space defined in Eq.~(\ref{eq:provefeasibility_all}) enforces the pairwise robots to stay occlusion-free with satisfying probability $\sigma^\mathrm{los}$, and (b) the control space for the set of pairwise robots determined by $\mathcal{E}^\mathrm{slos}$ in Eq.~(\ref{eq:prlos_cbc}) is always non-empty, we can guarantee that by enforcing the joint control input to stay in the admissible control space defined by our PrLOS-CBC, the joint system state $\mathbf{x}$ stays in the intersection set $\{\bigcap_{\{v_i,v_j \in \mathcal{V}:(v_i,v_j)\in\mathcal{E}^\mathrm{slos}\}} \mathcal{H}^\mathrm{los}_{i,j}\}$ with satisfying probability. Thus, we conclude the proof.
\end{proof}

\subsection{Feasibility Discussion of the Quadratic Programming Problem in Eq.~(\ref{eq:rawobj})}\label{app:sec:feasibility}

In this paper, the detailed control constraints are defined as:
{\footnotesize
\begin{align}\label{eq:our_constraints}
\mathcal{S}^{\sigma^\mathrm{s}}_\mathbf{u}(\mathbf{\hat{x}}) \textstyle \bigcap S^{\sigma^\mathrm{obs}}_\mathbf{u}(\mathbf{\hat{x}},\mathbf{x}^\mathrm{obs})
\bigcap \mathcal{C}^{\sigma^\mathrm{c}}_\mathbf{u}(\mathbf{\hat{x}},\mathcal{G}^\mathrm{slos})\bigcap \mathcal{C}^{\sigma^\mathrm{los}}_\mathbf{u}(\mathbf{\hat{x}},\mathcal{C}_\mathrm{obs},\mathcal{G}^\mathrm{slos})
\end{align}}
which consist of (a) the PrSBC $\mathcal{S}^{\sigma^\mathrm{s}}_\mathbf{u}(\mathbf{\hat{x}})$ and $S^{\sigma^\mathrm{obs}}_\mathbf{u}(\mathbf{\hat{x}},\mathbf{x}^\mathrm{obs})$ in Eq.~(\ref{Prsbc})- a set of safety prescribed control constraints, (b) $\mathcal{C}^{\sigma^\mathrm{c}}_\mathbf{u}(\mathbf{\hat{x}},\mathcal{G}^\mathrm{slos})$ in Eq.~(\ref{Prcbc})- a set of inter-robot communication distance prescribed control constraints, and (c) the Probabilistic Line-of-Sight Connectivity Barrier Certificates (PrLOS-CBC) in Eq.~(\ref{eq:prlos_cbc})- a set of inter-robot occlusion-free condition constraints. 

In \cite{luo2020multi}, the author has proven that the joint control $\mathbf{u} = \mathbf{u}^{0}$ leading the $\dot{\mathbf{x}} = [\dot{\mathbf{x}}_1,...,\dot{\mathbf{x}}_N] = \mathbf{0}\in\mathbb{R}^{dN}$ is always the feasible solution satisfying the constraints in  $\mathcal{S}^{\sigma^\mathrm{s}}_\mathbf{u}(\mathbf{\hat{x}})$ and $S^{\sigma^\mathrm{obs}}_\mathbf{u}(\mathbf{\hat{x}},\mathbf{x}^\mathrm{obs})$.
Due to the similar structure of pairwise inter-robot safety and communication distance constraints in Eq.~(\ref{eq:h_safe_per}) and Eq.~(\ref{eq:h_conn_per}), one can adopt the same idea to prove that $\mathbf{u} = \mathbf{u}^{0}$ is also always the feasible solution satisfying the constraints in $\mathcal{C}^{\sigma^\mathrm{c}}_\mathbf{u}(\hat{\mathbf{x}},\mathcal{G}^\mathrm{slos})$. Readers are referred to \cite{luo2020multi} for detailed discussion.

Besides, in Eq.~(\ref{eq:provefeasibility_all}) we have shown the detailed formulation for each single constraint on the pairwise robots in our PrLOS-CBC. Then it is straightforward to prove that one particular solution $\mathbf{u}=\mathbf{u}^0$ leading the $\dot{\mathbf{x}} = [\dot{\mathbf{x}}_1,...,\dot{\mathbf{x}}_N] = \mathbf{0}$, located at the intersection of the admissible control space constrained by PrLOS-CBC $\mathcal{C}^{\sigma^\mathrm{los}}_\mathbf{u}(\hat{\mathbf{x}},\mathcal{C}_\mathrm{obs},\mathcal{G}^\mathrm{slos})$, is one particular solution for the robot team to stay occlusion-free with satisfying probability. 
Hence, considering the unbounded control input, we can always guarantee that $\mathbf{u}=\mathbf{u}^0$ is always the feasible solution satisfying the constraints in Eq.~(\ref{eq:our_constraints}). In presence of bounded input constraints, i.e., $\mathcal{U}_i := \{\mathbf{u}_i \in \mathbb{R}^{q}: u_\mathrm{min}\leq||\mathbf{u}_{i}|| \leq u_\mathrm{max}\}$, the feasible set constrained by Eq.~(\ref{eq:rawconst}) could be empty. For example, due to the physical limitations of the robot's braking system, it is not possible for the robot to instantaneously decelerate to a complete stop. The authors in \cite{xiao2022sufficient} provided a novel method to find sufficient conditions, which
are captured by a single constraint and formulated from an additional CBF, to guarantee the feasibility of original CBF-based QPs.
Note that the additional CBF will always be compatible with the existing constraints, implying that it cannot make the previous feasible set of constraints infeasible. Readers are referred to \cite{xiao2022sufficient} for further details.

\subsection{Proof of Theorem~\ref{theorem:LCT}}\label{app:sec:proof_theorem4}
\begin{customthm}{5}
The Uncertainty-Aware Line-of-Sight Least Constraining Tree (ULOS-LCT) $\bar{\mathcal{T}}_w^\mathrm{los'}$ is (i) globally and subgroup LOS connected, and (ii) least violated if following nominal joint controller $\mathbf{u}$ compared to all other candidate spanning trees in $\{\mathcal{T}_w^\mathrm{los'}=(\mathcal{V},\mathcal{E}^T,\mathcal{W}^{T'})\}$.
\end{customthm}
\begin{proof}
Recall that in this paper, the optimal graph $\mathcal{G}^\mathrm{slos*}\subseteq\mathcal{G}^\mathrm{los}$ for robots to maintain at each time step should (a) be a spanning tree of current $\mathcal{G}^\mathrm{los}$ which has the least number ($N-1$) of edges to satisfy global and subgroup LOS connectivity, and (b) introduce the least violated constraints $\mathcal{C}^{\sigma^\mathrm{c}}_\mathbf{u}(\hat{\mathbf{x}},\mathcal{G}^\mathrm{slos*})\cap\mathcal{C}^{\sigma^\mathrm{los}}_\mathbf{u}(\hat{\mathbf{x}},\mathcal{C}_\mathrm{obs},\mathcal{G}^\mathrm{slos*})$ under nominal joint task-related controller $\tilde{\mathbf{u}}=[\tilde{\mathbf{u}}_1,\ldots,\tilde{\mathbf{u}}_N]$ 
prescribed by $\mathcal{G}^\mathrm{slos*}$. 

Given the requirement of the $\mathcal{G}^\mathrm{slos*}$ as described above, in Definition~\ref{def:uccst}, we introduce the connectivity condition weight $w_{i,j}^\mathrm{d}$ and the occlusion-free condition weight $w_{i,j}^\mathrm{los}$ for all edges in the entire LOS communication graph $\mathcal{G}^\mathrm{los} = (\mathcal{V},\mathcal{E}^\mathrm{los})$ 
between robots $i$ and $j$ under nominal controller $\tilde{\mathbf{u}}_i$ and $\tilde{\mathbf{u}}_j$ as follows.
{\footnotesize
\begin{align}
&w^\mathrm{los}_{i,j}=\frac{1}{L}\sum_{o=1}^{L}({{\dot{h}_{i,j,o}^\mathrm{los}(\hat{\mathbf{x}},\mathbf{x}^\mathrm{obs},\sigma^\mathrm{los},\tilde{\mathbf{u}})+\gamma h_{i,j,o}^\mathrm{los}(\hat{\mathbf{x}},\mathbf{x}^\mathrm{obs},\sigma^\mathrm{los})}}) \label{eq:revise_L} \notag\\
&w^\mathrm{d}_{i,j}=f^{\sigma^\mathrm{c}}_{i,j}-B^{\sigma^\mathrm{c}}_{i,j}\tilde{\mathbf{u}} \notag
\end{align}}
where $B^{\sigma^\mathrm{c}}_{i,j}$ and $f^{\sigma^\mathrm{c}}_{i,j}$, are from Eq.~(\ref{Prcbc}). Recall that the higher value of $w_{i,j}^\mathrm{d}$ and $w_{i,j}^\mathrm{los}$ indicates the less violation under nominal controller $\tilde{\mathbf{u}}_i$ and $\tilde{\mathbf{u}}_j$ . And then in Definition~\ref{def:uccst},  we define the sum of two weights as:
$
w_{i,j}^\mathrm{d+los} = w_{i,j}^\mathrm{d}+w_{i,j}^\mathrm{los} 
$

This weight is used to heuristically quantify the total violation of LOS connectivity on each $(v_i,v_j)\in\mathcal{E}^\mathrm{los}$ under nominal controller $\tilde{\mathbf{u}}_i$ and $\tilde{\mathbf{u}}_j$. With that, each spanning tree $\mathcal{T}^\mathrm{los}\subseteq \mathcal{G}^\mathrm{los}$ as candidate solution of $\mathcal{G}^\mathrm{slos*}$ can be redefined as a weighted spanning tree
$\mathcal{T}^\mathrm{los}_w=(\mathcal{V},\mathcal{E}^{T},\mathcal{W}^T)$
where $\mathcal{E}^T \subseteq \mathcal{E}^\mathrm{los}$ with weight $\mathcal{W}^T=\{-w^\mathrm{d+los}_{i,j}\}$. Hence, the optimal LOS communication graph $\mathcal{G}^\mathrm{slos*}$ satisfying constraints in Eq.~(\ref{eq:rawglobal}) and Eq.~(\ref{eq:rawconn}) can be defined as follows. 
\begin{equation}
\begin{split}
\mathcal{G}^\mathrm{slos*}&=\!\argmax_{\{\mathcal{T}_w^\mathrm{los}\}}\sum_{(v_i,v_j)\in \mathcal{E}^{T}}w^\mathrm{d+los}_{i,j}=
\!\argmin_{\{\mathcal{T}_w^\mathrm{los}\}}\!\sum_{(v_i,v_j)\in \mathcal{E}^{T}}-w^\mathrm{d+los}_{i,j}\\
\text{s.t.} &\quad  \mathcal{T}^\mathrm{los}_m=\mathcal{T}_w^\mathrm{los}[\mathcal{V}_m]\quad \text{is LOS connected},\forall m=1,\ldots,M \label{eq:constained_mst}  
\end{split}
\end{equation}

The optimal solution of problem in Eq.~(\ref{eq:constained_mst}) is a Constrained Minimum Spanning Tree (CMST) weighted by $\mathcal{W}^T=\{-w^\mathrm{d+los}_{i,j}\}$. 

Then we introduce the subgroup parameter $\beta >> 1$ in Definition~\ref{def:uccst} to transform this CMST to an unconstrained Minimum Spanning Tree (MST) problem with the same optimality guarantee, so that it can be solved efficiently by standard MST algorithm. To be more specific, the subgroup parameter $\beta$ ensures that for any pairwise robots, we have $w_{i, j}' < w_{i, j'}'$ if robot $i$ and $j$ are in the same subgroup while the robot $i$ and $j'$ are in the different subgroup. Then we denote the weight-modified graph as $\mathcal{G}'$ with its spanning trees $\{\mathcal{T}_w^\mathrm{los'}=(\mathcal{V},\mathcal{E}^T,\mathcal{W}^{T'})\}$ (ULOS-ST in Definition~\ref{def:uccst}), where $\mathcal{W}^{T'}=\{w'_{i,j}\}$ in Eq.~(\ref{eq:neww}). Next, we will prove that ULOS-LCT $\bar{\mathcal{T}}_w^\mathrm{los'}= \argmin_{\{\mathcal{T}_w^\mathrm{los'}\}} \sum_{(v_i,v_j)\in \mathcal{E}^{T}}\{w'_{i,j}\}$, which is the particular spanning tree $\mathcal{T}_w^\mathrm{los'}$ with the minimum total weight (i.e., MST), is the solution of the problem defined in Eq.~(\ref{eq:constained_mst}). Due to the property of MST, it is straightforward to prove that the edges across different subgroups are LOS connected after edges within each subgroup are LOS connected. With this, the MST of induced sub-graph $\mathcal{G}\mathrm{'}[\mathcal{V}_m]$ within subgroup $\mathcal{S}_m$ is LOS connected and optimal with minimum total weight, as shown below.
{\footnotesize\begin{equation}
\begin{split}
     \bar{\mathcal{T}}_w^\mathrm{los'}(m)&=\argmin_{\{\mathcal{T}_w^\mathrm{los'}(m)\}} \sum_{(v_i,v_j)\in \mathcal{E}^{T(m)}}w'_{i,j}\\
     &=\argmin_{\{\mathcal{T}_w^\mathrm{los'}(m)\}} \beta\cdot\sum_{(v_i,v_j)\in \mathcal{E}^{T(m)}}-w^\mathrm{d+los}_{i,j} \\
     &=\argmin_{\{\mathcal{T}_w^\mathrm{los'}(m)\}} \sum_{(v_i,v_j)\in \mathcal{E}^{T(m)}}-w^\mathrm{d+los}_{i,j}
\end{split}
\end{equation}}

The equality holds since $\beta > 0$. Then we consider $v_i$ and $v_j$ in different subgroups, i.e. $\mathcal{S}(v_i) \neq \mathcal{S}(v_j)$, while $(v_i, v_j)$ is the edge in spanning tree edges $\mathcal{E}^{T(m)}$ that is LOS connected. Connecting the minimum-weighted outgoing edge (MWOE) between different sub-groups yields:
{\footnotesize
\begin{equation} 
\begin{split}
    \bar{\mathcal{T}}_w^\mathrm{los'}
     &=\argmin_{\{\mathcal{T}_w^\mathrm{los'}\}} \sum_{(v_i,v_j)\in \mathcal{E}^{T(m)}}w'_{i,j}, \quad \mathcal{S}(v_i) \neq \mathcal{S}(v_j)\\
     &=\argmin_{\{\mathcal{T}_w^\mathrm{los'}\}} \sum_{(v_i,v_j)\in \mathcal{E}^{T}}-w^\mathrm{d+los}_{i,j}
\end{split} 
\end{equation}}
In this way, we prove that ULOS-LCT $\bar{\mathcal{T}}_w^\mathrm{los'}$ is globally and subgroup LOS connected and has the same optimal guarantee in Eq.~(\ref{eq:constained_mst}). Thus, we conclude the proof.
\end{proof}

\subsection{Proof of Proposition~\ref{proposition:graph}}\label{app:sec:proof_proposition5}
\begin{customprop}{6}
By Algorithm~\ref{alg:dis_LCT}, each robot agrees with the same LOS communication graph $\mathcal{G}^\mathrm{slos}$, which is the real-time ULOS-LCT $\bar{\mathcal{T}}_w^\mathrm{los'}$. 
\end{customprop}
\begin{proof}
The repeated connection of a fragment's MWOE to an adjacent node in another fragment is proven to contribute to a unique MST for graphs with unique edge weights \cite{gallager1983distributed,peleg2000distributed}.
At the beginning of our Algorithm~\ref{alg:dis_LCT}, each robot starts to create the fragment which contains only itself. With this, the iterations of Algorithm~\ref{alg:dis_LCT} starts with selecting the leader within the current fragment in Line~\ref{alg:update_leader}. Then the leader decides which adjacent node to connect based on the updated MWOE information in Line~\ref{algorithm:update_gaph}, yielding another fragment which is also part of the MST. In summary, during the construction of the MST in Line~\ref{alg:update_leader}-\ref{algorithm:update_ad} of Algorithm~\ref{alg:dis_LCT},  
the leader repeatedly guides the current fragment to connect the adjacent node in a different fragment with MWOE. When the algorithm converges, there is only one leader among the entire robot team. Since only the leader of each fragment updates the global adjacency matrix $Ad$ in Line~\ref{algorithm:update_ad} of Algorithm~\ref{alg:dis_LCT} within the fragment, the algorithm eventually terminates with only one fragment, i.e., the MST, with a single leader. This implies the convergence of the decentralized algorithm and guarantees consistency among all robots. Hence, the resulting graph $\bar{\mathcal{T}}_w^\mathrm{los'}$ generated by the decentralized construction is the same as the tree generated by the centralized construction.
\end{proof}
\subsection{Proof of Proposition~\ref{proposition:guarantee}}\label{app:sec:proof_proposotion6}
\begin{lemma} \label{lemma:equal} [summarized from \cite{boyd2011distributed}]
The centralized QP problem in the format of Eq.~(\ref{eq:obj_final}) can be decomposed for each robot in the format of Eq.~(\ref{eq:each_distributed_ustar}). Adopting the C-ADMM based algorithm by updating Eq.~(\ref{eq:local_update}) for each robot iteratively until convergence, it has been proven that the derived solution is the solution of centralized QP problem, if the object function and the constraints in Eq.~(\ref{eq:local_update}) are convex and the original centralized QP problem has the feasible solution. 
\end{lemma}
\begin{customprop}{7}
By following Algorithm~\ref{alg:dis_LCT} at each time step, robots reach consensus regarding the admissible control space prescribed by 
$\mathcal{C}^{\sigma^\mathrm{c}}_\mathbf{u}(\hat{\mathbf{x}},\mathcal{G}^\mathrm{slos})\bigcap \mathcal{C}^{\sigma^\mathrm{los}}_\mathbf{u}(\hat{\mathbf{x}},\mathcal{C}_\mathrm{obs},\mathcal{G}^\mathrm{slos})$ where $\mathcal{G}^\mathrm{slos}=\bar{\mathcal{T}}_w^\mathrm{los'}$, and the derived $\mathbf{u}_i^*,\forall i=[1,N]$ is the solution of Eq.~(\ref{eq:obj_final}), rendering the resulting $\mathcal{G}^\mathrm{los} \supseteq\bar{\mathcal{T}}_w^\mathrm{los'}$ globally and subgroup LOS connected with satisfying probability at all times.
\end{customprop}

\begin{proof}
Since the process of the Algorithm~\ref{alg:dis_LCT} is updated over time, to prove Proposition~\ref{proposition:guarantee}, we give the explicit form of the LOS communication graph with respect to time $t$ (e.g., $\mathcal{G}^\mathrm{los} = \mathcal{G}^\mathrm{los}(t)$). Recall that in the manuscript, for ease of the notation, we omit the notion of dependence on time.
At each time step, by following the Algorithm~\ref{alg:dis_LCT}, it is guaranteed that by updating the safety constraints $\mathcal{S}_{\mathbf{u}^{i}}^{\sigma^{\mathrm{s}}}(\hat{\mathbf{x}}^{i}(t))$ and $\mathcal{S}_{\mathbf{u}^{i}}^{\sigma^{\mathrm{obs}}}(\hat{\mathbf{x}}^{i}(t),\mathbf{x}^\mathrm{obs})$ on Line~\ref{algorithm:update_safe} (through the message acquired from neighbors containing $\hat{\mathbf{x}}^i$ and $\tilde{\mathbf{u}}^i$) and LOS connectivity constraints $\mathcal{C}^{\sigma^\mathrm{c}}_{\mathbf{u}^{i}}(\hat{\mathbf{x}}^{i}(t),\mathcal{G}^\mathrm{slos}_i(t))\cap\mathcal{C}^{\sigma^\mathrm{los}}_{\mathbf{u}^{i}}(\hat{\mathbf{x}}^{i}(t),\mathcal{C}_\mathrm{obs},\mathcal{G}^\mathrm{slos}_i(t))$ on Line~\ref{algorithm:update_control}, the admissible control space for the team of robots is the equivalent transformation as the admissible control space in Eq.~(\ref{eq:const_final}). Since our objective function and constraints are convex in Eq.~(\ref{eq:local_update}) and the QP problem in Eq.~(\ref{eq:obj_final}) has the feasible solution (discussed in Section~\ref{app:sec:feasibility}), following Lemma~\ref{lemma:equal}, the derived $\mathbf{u}_i^*(\hat{\mathbf{x}}_i(t)),\forall i=[1,N]$ is guaranteed to be the solution of Eq.~(\ref{eq:obj_final}).

Since we assume the team of robots is globally and subgroup LOS connected initially, then with Proposition~\ref{proposition:graph}, it is guaranteed that each robot will agree with the same ULOS-LCT $\bar{\mathcal{T}}_w^\mathrm{los'}(t=t_0)\subseteq \mathcal{G}^\mathrm{los}(t=t_0)$ that is both globally and subgroup LOS connected. Then, considering the state-dependent Lipschitz continuous controller $\mathbf{u}(\hat{\mathbf{x}}(t)) \in \mathcal{C}^{\sigma^\mathrm{c}}_\mathbf{u}(\hat{\mathbf{x}}(t),\bar{\mathcal{T}}_w^\mathrm{los'}(t_0))\bigcap \mathcal{C}^{\sigma^\mathrm{los}}_\mathbf{u}(\hat{\mathbf{x}}(t),\mathcal{C}_\mathrm{obs},\bar{\mathcal{T}}_w^\mathrm{los'}(t_0))$ for all $t\in [t_0,t_0 +\tau]$, it guarantees that each pairwise robots in the LOS communication graph $\bar{\mathcal{T}}_w^\mathrm{los'}(t=t_0)$ remains LOS connected with satisfying probability. With that, we can guarantee that $\mathcal{G}^\mathrm{los}(t)\supseteq\bar{\mathcal{T}}_w^\mathrm{los'}(t=t_0)$ remains globally and subgroup LOS connected with satisfying probability within $t\in [t_0,t_0 +\tau]$. Hence, at the next time step $t_1 = t_0 + \tau$, it guarantees that $\bar{\mathcal{T}}_w^\mathrm{los'}(t=t_0) \subseteq \mathcal{G}^\mathrm{los}(t =t_1)$ as well. In other words, it guarantees that $\mathcal{G}^\mathrm{los}(t=t_1)$ is globally and subgroup LOS connected with satisfying probability. In summary, the Dec-LOS-LCT algorithm can guarantee that the preserved optimal spanning tree $\bar{\mathcal{T}}_w^\mathrm{los'}(t) = \mathcal{G}^\mathrm{slos*}(t)$ is always the subgraph of the resulting LOS communication graph $\mathcal{G}^\mathrm{los}(t)$ that are thereby globally and subgroup LOS connected with satisfying probability. 
\end{proof}

\subsection{Proof of Proposition~\ref{proposition:probability}}\label{app:sec:proof_proposotion8}
\begin{customprop}{8}
By choosing $\sigma^\mathrm{los}=1-\frac{1-\sigma^\mathrm{graph}}{N-1}$ as the pair-wise robots occlusion-free confidence level to guarantee the occlusion-free condition for ULOS-LCT $\bar{\mathcal{T}}_w^\mathrm{los'}$, the resultant LOS communication graph $\mathcal{G}^\mathrm{los} \supseteq\bar{\mathcal{T}}_w^\mathrm{los'}$, satisfies the occlusion-free condition, ensuring at least one occlusion-free path between every pair of vertices with a probability greater than $\sigma^\mathrm{graph}$. 
\end{customprop}
\begin{proof}
Given the occlusion-free condition for each pairwise robots in Eq.~(\ref{los1}), the set of pair-wise robots that doesn't satisfy the occlusion condition is the complement of set $\mathcal{H}^\mathrm{los}_{i,j}$, denoted as $(\mathcal{H}^\mathrm{los}_{i,j})^\complement$. Then given the  ULOS-LCT $\bar{\mathcal{T}}_w^\mathrm{los'}=\mathcal{G}^\mathrm{slos*} = (\mathcal{V},\mathcal{E}^\mathrm{slos*})$, the set for the multi-robot team that doesn't satisfy the occlusion-free condition through a specific LOS communication graph $\mathcal{G}^\mathrm{slos*}$ can be defined as the complement of set $D= \{\mathbf{x}\in\mathbb{R}^{dN}|\bigcap_{\{v_i,v_j \in \mathcal{V}:(v_i,v_j)\in\mathcal{E}^\mathrm{slos*}\}} \mathcal{H}^\mathrm{los}_{i,j}\}$, which can be denoted as $D^\complement$. Because the optimal LOS communication graph $\mathcal{G}^\mathrm{slos*}$ is the subgraph of the LOS communication graph $\mathcal{G}^\mathrm{los}$ ($\mathcal{G}^\mathrm{slos*}\subseteq\mathcal{G}^\mathrm{los}$), the set for the robot team doesn't satisfy the occlusion-free condition (i.e., there is no occlusion-free path in current $\mathcal{G}^\mathrm{los}$), denoted as $Z$, is the subset of $D^\complement$, i.e, $Z\subseteq D^\complement$. Recall that the LOS communication graph $\mathcal{G}^\mathrm{slos*}$, which we aim to maintain, is the particular MST of the graph $\mathcal{G}^\mathrm{los}$ with exactly $N-1$ numbers of edges to be LOS connected. With this, the probability of set $Z$ is bounded by:
{\footnotesize\begin{align}
    \text{Pr}(\mathbf{x} \in Z) &\leq\text{Pr}(\mathbf{x} \in D^\complement) \notag \\
    &= \text{Pr}(\bigcup \mathbf{x}_{i,j} \in (\mathcal{H}^\mathrm{los}_{i,j})^\complement),\forall (v_i,v_j)\in\mathcal{E}^\mathrm{slos*} \notag \\
    &\leq \sum_{(v_i,v_j)\in\mathcal{E}^\mathrm{slos*}}\text{Pr}(\mathbf{x} \in (\mathcal{H}^\mathrm{los}_{i,j})^\complement) 
    =(N-1)(1-\sigma^\mathrm{los})
\end{align}}
Given that, with our Lemma~\ref{loscbcdefinition}, if one desires to ensure the optimal graph $\mathcal{G}^\mathrm{slos*}$ satisfies the occlusion-free condition with at least $\sigma^\mathrm{graph}\in(0,1)$ probability, then one can choose $\sigma^\mathrm{los}=1-\frac{1-\sigma^\mathrm{graph}}{N-1}$ to ensure the pair-wise robot satisfies the occlusion-free condition. This also indicates that, by maintaining a specific graph $\mathcal{G}^\mathrm{slos*}$, the probability of resultant LOS communication graph $\mathcal{G}^\mathrm{los}$ that satisfies the occlusion-free condition through at least one occlusion-free path between every pair of vertices on this graph is greater than $\sigma^\mathrm{graph}$. Thus, we conclude the proof.\end{proof}
\end{document}